\documentclass{article}

    \PassOptionsToPackage{numbers, compress}{natbib}

    \usepackage[final]{neurips_2022}

\usepackage{extarrows}
\usepackage[utf8]{inputenc} 
\usepackage[T1]{fontenc}    
\usepackage{hyperref}       
\usepackage{url}            
\usepackage{booktabs}       
\usepackage{amsfonts,amssymb}       
\usepackage{nicefrac}       
\usepackage{microtype}      
\usepackage[dvipsnames]{xcolor}
\usepackage{wrapfig}
\usepackage{multirow,mathtools } 
\usepackage{algorithm,algpseudocode}
\usepackage[color=gray!20,textsize=scriptsize]{todonotes}

\algnewcommand{\algorithmicforeach}{\textbf{for each}}
\algdef{SE}[FOR]{ForEach}{EndForEach}[1]
  {\algorithmicforeach\ #1\ \algorithmicdo}
  {\algorithmicend\ \algorithmicforeach}

\newtheorem{myprop}{\bf{Proposition}}

\DeclareMathOperator*{\minimize}{\text{minimize}}

\DeclareMathOperator*{\st}{\text{subject to}}
\DeclareMathAlphabet\mathbfcal{OMS}{cmsy}{b}{n}
\newcommand{\Def}[0]{\mathrel{\mathop:}=}

\newcommand{\mask}{\mathbf{m}}
\newcommand{\gradz}{\nabla_{\mathbf z} \ell (\mathbf z)\left |_{\mathbf z = {\mask} \odot {\btheta}^*} \right.}
\newcommand{\gradzsimple}{\nabla_{\mathbf z} \ell (\mathbf z)}

\newcommand{\btheta}{\boldsymbol{\theta}}

\usepackage{adjustbox}

\newcommand{\ltr}{\ell_{\mathrm{tr}}}
\newcommand{\YH}[1]{\textcolor{orange}{YH: #1}}
\newcommand{\YY}[1]{\textcolor{OliveGreen}{YY: #1}}
\newcommand{\RV}[1]{\textcolor{black}{#1}}
\newcommand{\biprune}{\textsc{BiP}}
\newcommand{\hydra}{\textsc{Hydra}}
\newcommand{\grasp}{\textsc{Grasp}}

\newcommand{\snip}{\textsc{Snip}}

\newcommand{\synflow}{\textsc{SynFlow}}

\newcommand{\xmark}{\textcolor{red}{\ding{55}}}
\definecolor{ompcolor}{rgb}{0.53, 0.66, 0.42}
\definecolor{Gray}{gray}{0.9}
\usepackage{pifont}

\usepackage{amsmath,amsfonts,bm}

\def\eqref#1{(\ref{#1})}

\def\1{\bm{1}}

\DeclareMathAlphabet{\mathsfit}{\encodingdefault}{\sfdefault}{m}{sl}
\SetMathAlphabet{\mathsfit}{bold}{\encodingdefault}{\sfdefault}{bx}{n}

\DeclareMathOperator*{\argmin}{arg\,min}

\definecolor{Sijia_color}{rgb}{0.858, 0.188, 0.478}

\newcommand{\CR}[1]{\textcolor{black}{#1}}

\usepackage[most]{tcolorbox}

\title{
Advancing Model Pruning via Bi-level Optimization
}

\author{%
  Yihua Zhang\textsuperscript{1,*}
  \\
  \And
  Yuguang Yao\textsuperscript{1,*} \\
  \And
  Parikshit Ram\textsuperscript{2} \\
  \And
  Pu Zhao\textsuperscript{3} \\
  \And
  Tianlong Chen\textsuperscript{4} \\
  \And 
  Mingyi Hong\textsuperscript{5} \\
  \And 
  Yanzhi Wang\textsuperscript{3} \\
  \And 
  Sijia Liu\textsuperscript{1,2}\\
  \AND \vspace*{-5mm}
  \\
  \textsuperscript{1}Michigan State University,
  \textsuperscript{2} IBM Research, 
  \textsuperscript{3}Northeastern University, \\
  \textsuperscript{4}University of Texas at Austin, 
  \textsuperscript{5}University of Minnesota, Twin Cities\\
  \textsuperscript{*}Equal contribution
}

\begin{document}

\maketitle

\begin{abstract}
   The deployment constraints in practical applications necessitate the pruning of large-scale deep learning models, \textit{i.e.}, promoting their weight sparsity. As illustrated by the Lottery Ticket Hypothesis (LTH), pruning 
    also has the potential of improving their generalization ability. At the core of LTH, iterative magnitude pruning (IMP) is the predominant pruning method to successfully find `winning tickets'.
    Yet, the computation cost of IMP  grows prohibitively as the targeted pruning ratio increases. 
   To reduce the computation overhead, various efficient `one-shot' pruning methods have been developed but these schemes are usually unable to find winning tickets as good as IMP. This raises the question of \textit{how to close the gap between pruning accuracy and pruning efficiency?}
   To tackle it,  we pursue the algorithmic advancement of model pruning. Specifically, we formulate the pruning problem  from a fresh and novel viewpoint, bi-level optimization  (BLO). We show that the BLO interpretation  provides a technically-grounded optimization base for an efficient implementation of the pruning-retraining learning paradigm used in IMP.
   We also show that the proposed \underline{bi}-level optimization-oriented \underline{p}runing method (termed \textit{{\biprune}}) is a special class of BLO problems with a bi-linear problem structure.
    By leveraging such bi-linearity, we theoretically show that {\biprune} can be solved as easily as first-order optimization, thus inheriting
    the computation efficiency. 
    Through extensive experiments on {\em both structured and unstructured pruning} with 5 model architectures and 4 data sets, we demonstrate that {\biprune} can  find   better winning tickets
    than IMP in most cases, and  is computationally as efficient as the one-shot pruning schemes, demonstrating 2-7$\times$ speedup over IMP for the same level of model accuracy and sparsity. Codes are available at \url{https://github.com/OPTML-Group/BiP}.
\end{abstract}

\section{Introduction }
%
While over-parameterized structures are key to the improved generalization of deep neural networks (DNNs)~\citep{krizhevsky2012imagenet, brown2020language,bartlett2021deep}, they create new problems -- the millions or even billions of parameters not only increase computational costs during inference, but also pose serious deployment challenges on resource-limited devices~\citep{han2015deep}. As a result, model pruning has seen a lot of research interest in recent years, focusing on reducing model sizes by removing (or pruning) redundant parameters~\citep{han2015deep, blalock2020state, han2015learning, he2018amc, mao2017exploring}. \RV{Model sparsity (achieved by pruning)  also benefits   adversarial robustness \cite{sehwag2020hydra}, out-of-distribution generalization \cite{diffenderfer2021winning}, and transfer learning \cite{chen2021lottery}. Some  pruning methods (towards structured sparsity) facilitate model deployment on hardware   \cite{ma2020pconv,chen2022coarsening}.}

Among various proposed model pruning algorithms\,\cite{blalock2020state,sehwag2020hydra,chen2021lottery,lecun1989optimal, ren2018admmnn, ramanujan2020sHiddenSubnetwork, frankle2018lottery, Renda2020Comparing, frankle2020linear, chen2020lottery,lee2018snip, ma2021sanity, wang2020pick,tanaka2020pruning, alizadeh2022prospect, lee2020layer,csordas2020neural},
the heuristics-based Iterative Magnitude Pruning (IMP) is the current dominant approach to achieving model sparsity without suffering performance loss, as suggested and empirically justified by the Lottery Ticket Hypothesis (LTH) \cite{frankle2018lottery}.
The LTH hypothesizes the existence of a subnetwork (the so-called `winning ticket') when trained in isolation (\textit{e.g.}, from either a random initialization \cite{frankle2018lottery} or an early-rewinding point of dense model training \cite{frankle2020linear}),
can match the performance of the original dense model \cite{blalock2020state,chen2021lottery,frankle2018lottery,Renda2020Comparing,frankle2020linear,chen2020lottery}. The core idea of IMP is to iteratively prune and retrain the model while progressively pruning a small ratio of the remaining weights in each iteration, and continuing till the desired pruning ratio has been reached. While IMP often finds the winning ticket, it incurs the cost of repeated model retraining from scratch, making it prohibitively expensive for large datasets or large model architectures \cite{you2019drawing, zhang2021efficient}.
To improve the efficiency of model pruning, numerous heuristics-based one-shot pruning methods\,\cite{frankle2018lottery, lee2018snip, ma2021sanity, wang2020pick, tanaka2020pruning, alizadeh2022prospect}, \emph{e.g.}, one-shot magnitude pruning (OMP), have been proposed. These schemes directly prune the model to the target sparsity and are significantly more efficient than IMP. 
{Yet,  the promise of current one-shot pruning methods is dataset/model-specific~\cite{ma2021sanity, su2020sanity_random_win} and mostly lies in the low pruning ratio regime~\cite{alizadeh2022prospect}.
As systematically studied in~\cite{ma2021sanity, frankle2020pruning}, there exists a clear performance gap between one-shot pruning and IMP.}
%
As an alternative to heuristics-based schemes, optimization-based pruning methods~\cite{sehwag2020hydra, ren2018admmnn, ramanujan2020sHiddenSubnetwork, lee2020layer, csordas2020neural} still follow the pruning-retraining paradigm and adopt sparsity regularization\,\cite{liu2019admm, wang2020neural}  or parameterized masking\,\cite{sehwag2020hydra, ramanujan2020sHiddenSubnetwork, csordas2020neural} to prune models efficiently. However, these 
methods do not always
match the accuracy of IMP and  thus have not been widely used to find winning tickets\,
\cite{frankle2018lottery, lee2018snip, wang2020pick,tanaka2020pruning, alizadeh2022prospect, zhang2021efficient}.
{The empirical results showing that optimization underperforms heuristics} 
motivate us to revisit the algorithmic fundamentals of pruning.
%

\begin{wrapfigure}{r}{60mm}
\vspace*{-5mm}
\centerline{
\includegraphics[width=58mm]{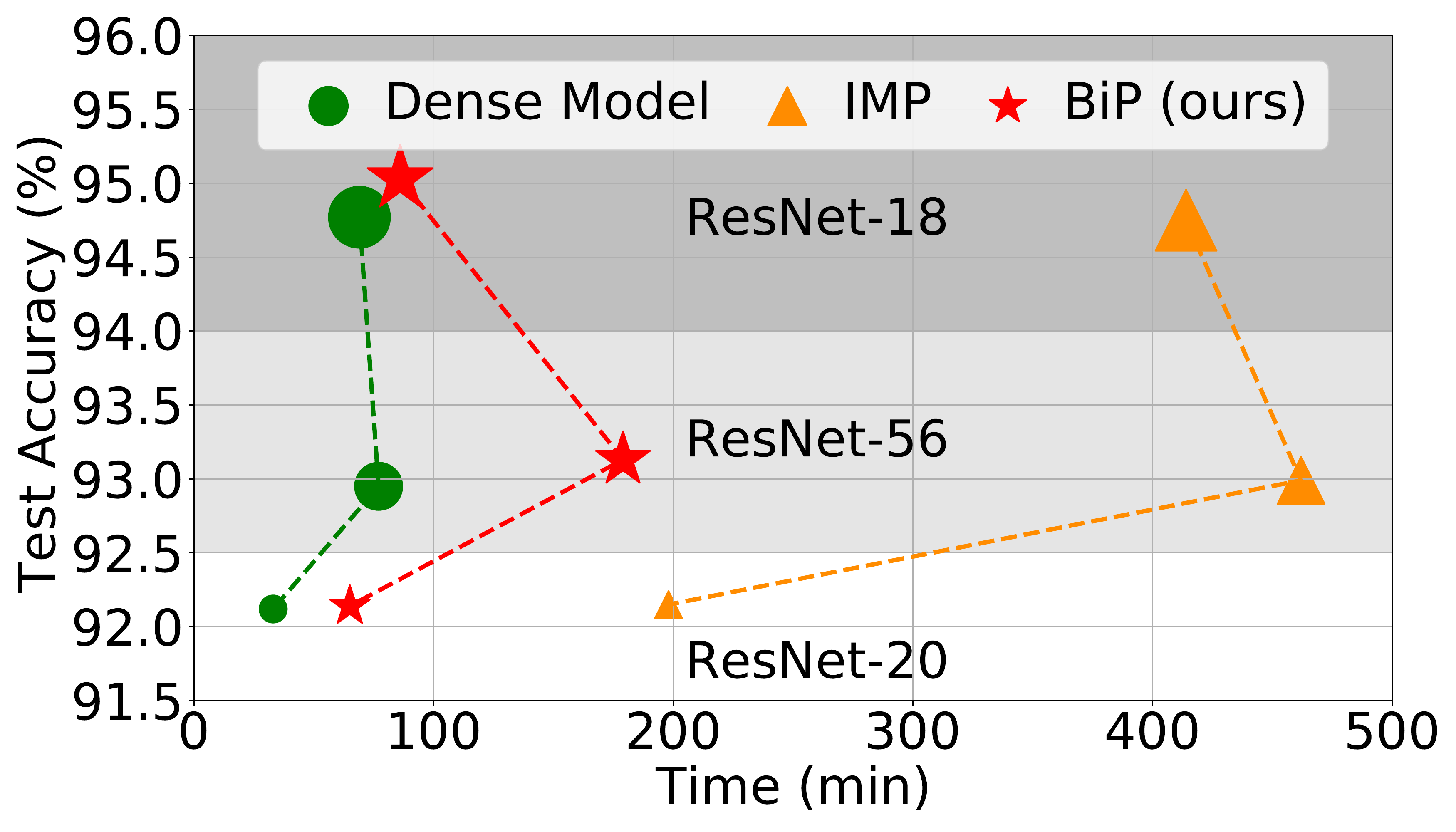}
}
\vspace*{-4mm}
\caption{\footnotesize{
A performance snapshot of the proposed {\biprune} method vs. the IMP baseline and the original dense model across three pruned ResNet models (ResNet-20, ResNet-56, and ResNet-18) with 74\% sparsity on CIFAR-10. The marker size indicates the relative model size. The uni-color region corresponds to the same model type used by different pruning methods.
}}
\label{fig: intro-c2}
\vspace*{-5mm}
\end{wrapfigure}
To this end, we put forth a novel perspective of model pruning as a bi-level optimization (BLO) problem. In this new formulation, we show that BLO provides a technically-grounded optimization basis for an efficient implementation of the pruning-retraining paradigm, the key algorithmic component used in IMP. 
To the best of our knowledge, we make the first rigorous connection between model pruning and BLO. Technically, we propose a novel \textbf{bi}-level optimization-enabled \textbf{p}runing method (termed {\biprune}).  We further show how {\biprune} takes advantage of the bi-linearity of the pruning problem to avoid the computational challenges of common BLO methods, and is as  efficient as any first-order alternating optimization scheme. Practically, we demonstrate the superiority of the proposed {\biprune} in terms of accuracy, sparsity, and  computation efficiency through extensive experiments. {\biprune} finds the best winning ticket nearly in all settings while taking time comparable to the one-shot OMP. In Fig.\,\ref{fig: intro-c2}, we present a snapshot of our empirical results for CIFAR-10 with 3 ResNet architectures at 74\% pruning ratio. In all cases, {\biprune}
(\textcolor{red}
{$\star$}) finds winning tickets, improving accuracy over the dense model (\textcolor{ForestGreen}{ $\bullet$}) and matching IMP (\textcolor{orange}{$\blacktriangle$}), while being upto 5$\times$ faster than IMP.

Our \textbf{contributions} can be summarized as follows:

\noindent $\bullet$ (Formulation) We rethink the algorithmic foundation of model pruning through the lens of BLO (bi-level optimization). The new BLO-oriented formulation 
disentangles
 pruning and retraining variables, providing the flexibility to design the interface between pruning and retraining.

\noindent $\bullet$ (Algorithm) We propose the new bi-level pruning ({\biprune}) algorithm, which is built upon the aforementioned BLO formulation and the implicit gradient-based optimization theory. Unlike computationally intensive standard BLO solvers, we theoretically show that {\biprune} is as efficient as any first-order optimization by taking advantage of the bi-linear nature of the pruning variables. 

\noindent $\bullet$ (Experiments) We conduct extensive experiments across 4 datasets (CIFAR-10, CIFAR-100, Tiny-ImageNet and ImageNet), 5 model architectures, and 3 pruning settings (unstructured pruning, filter-wise structured pruning, and channel-wise structured pruning).
We show that (i)~{\biprune} achieves higher test accuracy than IMP and finds the best winning tickets nearly in all settings, (ii)~{\biprune} is highly efficient (comparable to one-shot pruning schemes), that is able to achieve 2-7$\times$ speedup over IMP for the same level of model accuracy and sparsity, and (iii)~{\biprune} is able to find subnetworks that achieve better performance than the dense model regardless of initialization rewinding.  





\section{Related Work and Open Question}
\label{sec: preliminary}
%
\paragraph{Neural network pruning.}
As neural networks become deeper and more sophisticated,
model pruning technology has gained increasing attention over the last decade since pruned models are necessary for the deployment of deep networks in practical applications~\cite{han2015deep, denil2013overparamterized_and_redundancy, liu2020primer}. 
With the goal of finding highly-sparse {\em and} highly-accurate subnetworks from original dense models, 
a variety of pruning methods have been developed such as    heuristics-based pruning \cite{frankle2018lottery,lee2018snip, wang2020pick,tanaka2020pruning, alizadeh2022prospect, zhang2021efficient, zhu2017prune}  and  optimization-based pruning \cite{sehwag2020hydra, ramanujan2020sHiddenSubnetwork, lee2020layer, csordas2020neural}.
The former identifies redundant model weights by leveraging heuristics-based metrics such as weight magnitudes~\cite{han2015learning,frankle2018lottery,frankle2020linear,chen2021lottery, ma2021sanity, janowsky1989pruning,  frankle2020pruning, zhu2017prune,peste2021ac}, gradient magnitudes~\cite{lee2018snip, wang2020pick, tanaka2020pruning, mozer1989skeletonization, evci2020rigging}, and Hessian statistics~\cite{molchanov2019importance, yao2020pyhessian, lecun1990optimal, hassibi1993second, molchanov2016pruning, singh2020woodfisher}.
The latter is typically built on: 1)~sparsity-promoting optimization \cite{ren2018admmnn, wang2020neural, liu2017learning,he2017channel,zhou2016less,louizos2017learning},
 where model weights are trained by penalizing their sparsity-inducing norms, such as $\ell_0$ and $\ell_1$ norms for irregular weight pruning, and $\ell_2$ norm for structured pruning; 2)~parameterized masking \cite{ramanujan2020sHiddenSubnetwork, sehwag2020hydra, guo2021gdp, liberis2021differentiable, kusupati2020soft, xue2021rethinking, zhou2021effective}, where model weight scores are optimized to filter the most important weights and achieve better performance. 
 
\noindent \textbf{Iterative vs. one-shot pruning, and motivation.}
Existing schemes can be further categorized into one-shot or iterative pruning based on the pruning schedule employed for achieving the targeted model sparsity.
Among the iterative schemes, the IMP (Iterative Magnitude Pruning scheme) \cite{frankle2018lottery, chen2020lottery, gale2019state, pmlr-v139-zhang21c, chen2020lottery2, yu2019playing, chen2021gans, ma2021good, gan2021playing, chen2021unified, kalibhat2021winning, chen2021ultra,zhu2017prune} 
has played a significant role in identifying high-quality `winning tickets', as postulated by 
LTH  (Lottery Ticket Hypothesis)  \cite{Renda2020Comparing,frankle2020linear}. 
\RV{To enable consistent comparisons among different methods, we extend the original definition of winning tickets in \cite{frankle2018lottery} to `matching subnetworks'  \cite{chen2020lottery} so as to cover different implementations of winning tickets, \textit{e.g.}, the use of
 early-epoch rewinding   for model re-initialization \cite{Renda2020Comparing} and the no-rewinding (\textit{i.e.}, fine-tuning) variant \cite{chen2020long}. Briefly, the matching subnetworks should  match or surpass the performance of the original dense model\,\cite{chen2020lottery}.} In this work, if a matching subnetwork is found better than the winning ticket obtained by the same method that follows the original LTH setup \cite{Renda2020Comparing,frankle2020linear},
we will also call such a matching subnetwork   a winning ticket throughout the paper.

\begin{wrapfigure}{r}{80mm}
\vspace*{-6mm}
\centerline{
\begin{tabular}{cc}
\hspace*{0mm}\includegraphics[width=.3\textwidth,height=!]{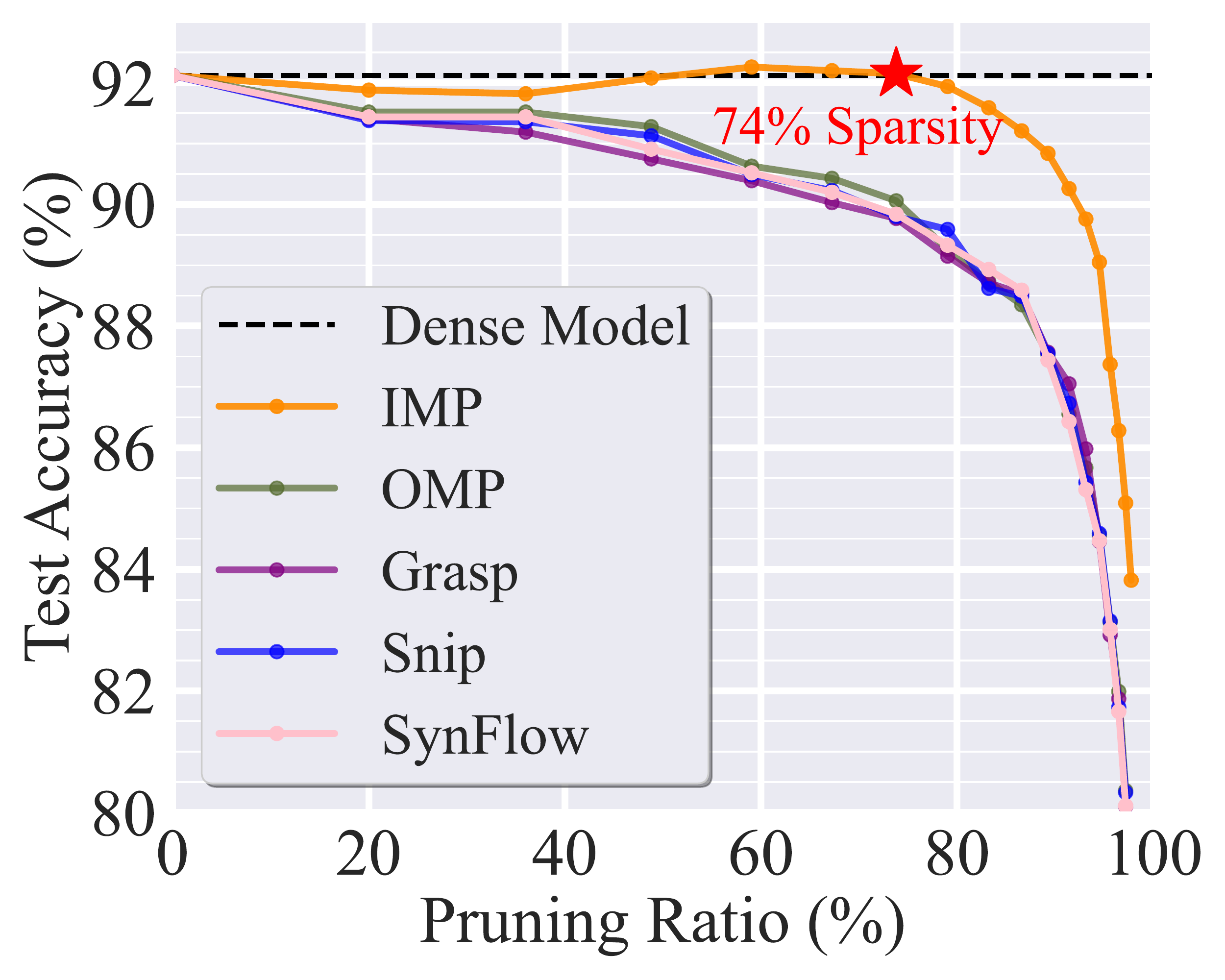}  
&\hspace*{-4mm}\includegraphics[width=.31\textwidth,height=!]{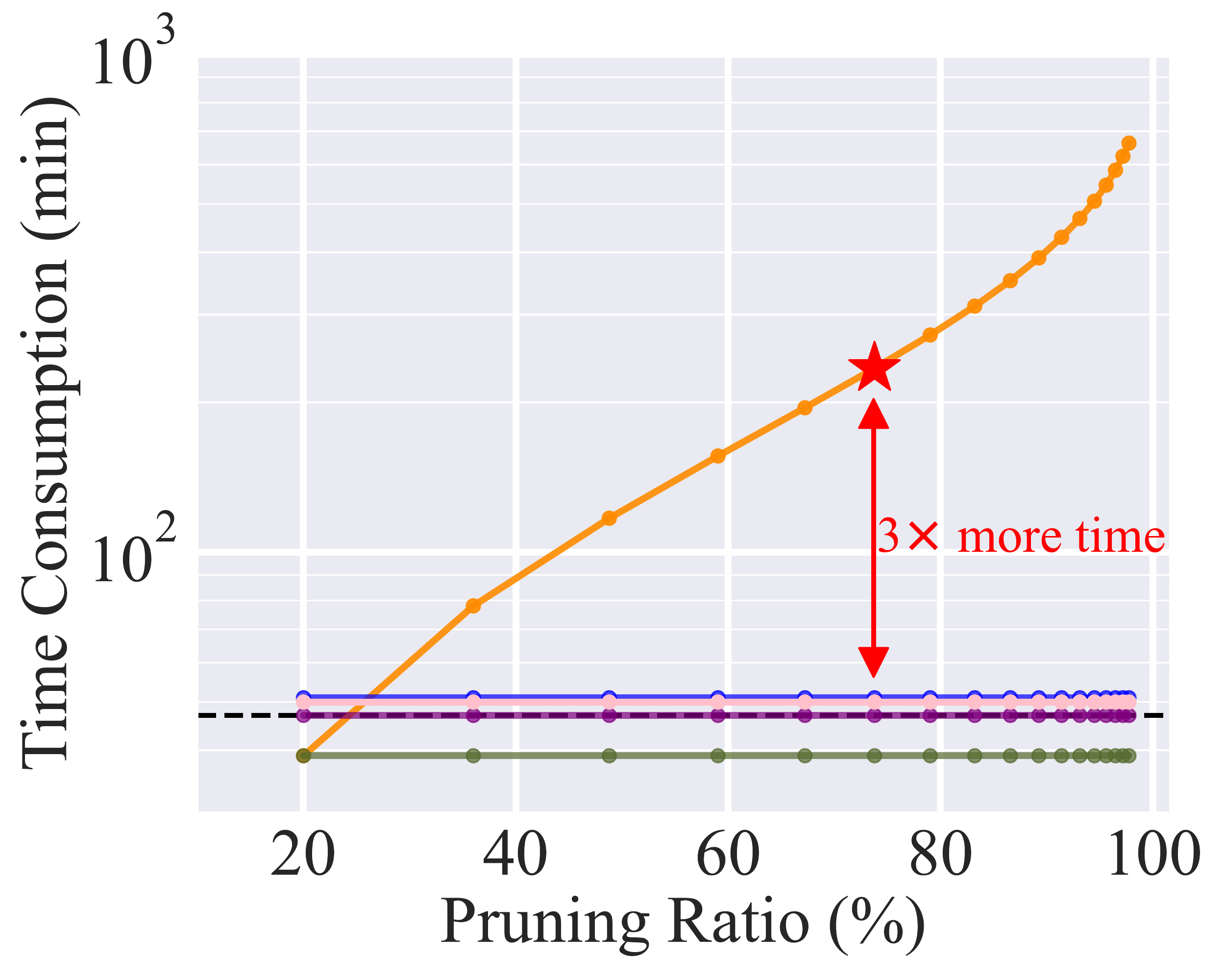}
\end{tabular}}
\vspace*{-3mm}
\caption{\footnotesize{
Illustration of the pros and cons of different pruning methods executed over (ResNet-20, CIFAR-10).
{\bf Left}: test accuracy vs. different pruning ratios of IMP and one-shot pruning methods (\textsc{OMP}\,\cite{frankle2018lottery}, \textsc{Grasp}\,\cite{wang2020pick}, \textsc{Snip}\,\cite{lee2018snip}, \textsc{SynFlow}\,\cite{tanaka2020pruning}). {\bf Right}: comparison of the efficiency with different sparsity. 
}}
  \label{fig: one_shot_results_merged}
  \vspace*{-3.8mm}
\end{wrapfigure}
For example, the current  state-of-the-art (SOTA) implementation of IMP in \cite{ma2021sanity} can lead to a pruned ResNet-20 on
CIFAR-10 with 74\% sparsity and $92.12\%$ test accuracy, matching the performance of the original dense model (see \textcolor{red}{red $\star$} in Fig.\,\ref{fig: one_shot_results_merged}-{\bf Left}). 
The IMP algorithm typically contains two key ingredients: \textbf{(i)}~a \textit{temporally-evolving pruning schedule} to progressively increase model sparsity over pruning iterations, and \textbf{(ii)}~the \textit{pruning-retraining learning mechanism} applied at each pruning iteration. 
With a target pruning ratio of $p\%$ with $T$ pruning
iterations, an example pruning schedule in (i) could be as follows -- each iteration prunes $(p\%)^{\nicefrac{1}{T}}$ of the currently unpruned model
weights, progressively pruning fewer weights in each iteration.
For (ii), 
the unpruned weights in each pruning iteration are re-set to the  weights at initialization or at an early-training epoch \cite{Renda2020Comparing}, and   re-trained till convergence.
In brief, IMP repeatedly prunes, resets, and trains the network over multiple iterations.
%

%
However, winning tickets found by IMP incur significant computational costs. The sparsest winning ticket found by IMP in Fig.~\ref{fig: one_shot_results_merged}-{\bf Left} (\textcolor{red}{red $\star$}) utilizes $T=6$  
pruning iterations. As shown in Fig.~\ref{fig: one_shot_results_merged}-{\bf Right}, this takes 3$\times$ more time than the original training of the dense model.
%
%
%
To avoid the computational cost of IMP, different kinds of `accelerated' pruning methods were developed  \cite{frankle2018lottery,lee2018snip, wang2020pick,tanaka2020pruning, alizadeh2022prospect, zhang2021efficient}, and many fall into the \textit{one-shot pruning} category: The network is directly pruned to the target sparsity  and retrained once. 
In particular, \textsc{OMP}  (one-shot magnitude pruning) is an important baseline that simplifies IMP~\cite{frankle2018lottery}.
It follows the pruning-retraining paradigm, but  prunes the model to the target ratio with a single pruning iteration. 
Although one-shot pruning schemes are computationally cheap (Fig.\,\ref{fig: one_shot_results_merged}-{\bf Right}), they incur a significant accuracy drop compared to IMP (Fig.\,\ref{fig: one_shot_results_merged}-{\bf Left}).
{Even if IMP is customized with a reduced number of training epochs  per pruning round, the pruning accuracy also drops largely (see Fig.\,\ref{fig: compare_IMP_OMP_GMP}).}
Hence, there is a need for advanced model pruning techniques to find winning tickets like IMP, while being efficient like one-shot pruning. 

Different from unstructured weight pruning described above,  structured pruning takes into consideration the sparse patterns of the model, such as filter and channel-wise pruning~\cite{chen2022coarsening, he2017channel, guo2021gdp, li2016pruning, niu2020patdnn, ma2020image, wang2022paca, van2020single}.
Structured pruning is desirable for deep model deployment in the presence of hardware constraints~\cite{han2015deep, denil2013overparamterized_and_redundancy}.
However, compared to unstructured pruning, it is usually more challenging to maintain the performance and find structure-aware winning tickets~\cite{you2019drawing, zhang2021efficient}. 

\paragraph{Open question.}
As discussed above, 
{one-shot pruning methods are unable to match the predictive performance of IMP, and structure-aware winning tickets are hard to find. 
Clearly, the best optimization foundation of model pruning is underdeveloped. Thus, we ask:} 
\begin{tcolorbox}
\vspace*{-2mm}
Is there an \textbf{optimization basis} for a successful pruning algorithm that can attain high pruned model accuracy (like IMP) {\em and} high computational efficiency (like one-shot pruning)?
\vspace*{-2mm}
\end{tcolorbox}
The model pruning problem has a natural hierarchical structure -- we need to find the best mask to prune model parameters, {\em and then}, given the mask, find the best model weights for the unpruned model parameters. Given this hierarchical structure, we believe that the {\em bi-level optimization (BLO)  framework} is one promising optimization basis for a successful pruning algorithm.
\paragraph{Bi-level optimization (BLO).} BLO is a general hierarchical optimization framework, where the upper-level problem depends on the solution to the lower-level problem. Such a nested structure makes the BLO in its most generic form very difficult to solve, since it is hard to characterize the influence of the lower-level optimization on the upper-level problem. Various existing literature focuses on the design and analysis of algorithms for various special cases of BLO. Applications range from the classical approximate descent methods\,\cite{falk1995bilevel, vicente1994descent, chen2022gradient}, penalty-based methods\,\cite{white1993penalty, gould2016differentiating}, to recent BigSAM\,\cite{sabach2017first} and its extensions\,\cite{liu2020generic, li2020improved}. It is also studied in the area of stochastic algorithms\,\cite{ghadimi2018approximation, ji2020bilevel, hong2020two} and back-propagation based algorithms\,\cite{franceschi2017forward, grazzi2020iteration, shaban2019truncated}. BLO has also advanced adversarial training\,\cite{zhang2021bat}, meta-learning\,\cite{rajeswaran2019meta}, data poisoning attack generation\,\cite{huang2020metapoison}, neural architecture search\,\cite{liu2018darts} as well as reinforcement learning\,\cite{chen2019research}. 
\RV{Although BLO was referred in \cite{louizos2017learning} for model pruning, it actually called an ordinary alternating optimization  procedure, without taking the hierarchical learning structure of BLO  into consideration.
To the best of our knowledge, the BLO framework  has not been  considered for model pruning in-depth and systematically}. 
{We will show that model pruning yields a special class of BLO problems with bi-linear optimization variables. We will also theoretically  show that  this specialized BLO problem for model pruning can be  solved as efficiently as first-order optimization. 
This is in a sharp contrast to existing BLO applications  that rely on heuristics-based BLO solvers (\textit{e.g.}, gradient unrolling in meta learning \cite{rajeswaran2019meta} and neural architecture search \cite{liu2018darts,ning2020dsa}).
}
\section{{\biprune}: Model Pruning via Bi-level Optimization}
\label{bi-level prune methodology}
In this section, we re-investigate the problem of model pruning through the lens of  BLO and develop the bi-level pruning (\biprune) algorithm.
We  can theoretically    show that {\biprune} 
can be solved as easily as the first-order alternating optimization by taking advantage of the bi-linearity of pruning variables.

\paragraph{A BLO viewpoint on model pruning}
As described in the previous section, the pruning-retraining learning paradigm covers two kinds of tasks: \ding{182} pruning  that determines the sparse pattern of model weights, and \ding{183} training remaining non-zero weights  to recover the model accuracy. 
In existing optimization-based pruning methods \cite{wen2016learning,zhang2018systematic,hoefler2021sparsity,wang2020proximal}, the tasks \ding{182}-\ding{183} are  typically achieved by optimizing model weights,   together with penalizing their sparsity-inducing norms, \textit{e.g.}, the $\ell_1$  and $\ell_2$ norms \cite{bach2012optimization}. Different from the above formulation, we propose to  separate optimization variables involved in the pruning tasks  \ding{182} and \ding{183}. This leads to the (binary) pruning mask variable $\mask \in \{ 0, 1\}^n$
and the model weight variable $\btheta \in \mathbb R^n$. Here $n$ denotes the total number of model parameters. 
Accordingly, the pruned model is given by $(\mathbf m \odot  \btheta)$, where $\odot$ denotes the element-wise multiplication. As will be evident later, this form of \textit{variable disentanglement} enables us to 
explicitly depict how the pruning and retraining process co-evolve, and helps customize  the pruning task with high flexibility.


We next elaborate on how BLO can be established to co-optimize the pruning mask $\mathbf m$  and the retrained sparse model weights $\boldsymbol \theta$. 
Given the pruning ratio $p\%$, the sparsity constraint   is given by $\mathbf m \in  \mathcal{S} $, where $\mathcal S = \{ \mathbf m\,|\, \mathbf m \in \{ 0, 1\}^n, \mathbf  1^T \mathbf m \leq k \}$ and $k = (1-p\%) n$.
Our goal is to  prune the original  model \textit{directly} to the targeted pruning ratio $p\%$ (\textit{i.e.}, without calling for the IMP-like sparsity schedule as described in Sec.\,\ref{sec: preliminary}) and obtain the optimized sparse model $(\mathbf m \odot \btheta)$. To this end, we interpret the pruning task (\textit{i.e.}, \ding{182}) and the model retraining task (\textit{i.e.}, \ding{183}) as \textit{two optimization levels}, where the former is formulated as an \textit{upper-level} optimization problem, and it relies  on the optimization of the \textit{lower-level}  retraining task. We thus cast the model pruning problem as the following BLO problem (with \ding{183} being nested inside \ding{182}):

{\small{
\vspace*{-6mm}
\begin{align}
\displaystyle
    \begin{array}{llll}
\displaystyle \minimize_{\mathbf \mask \in \mathcal{S}}         & \underbrace{\ell(\mask \odot {\btheta}^*(\mask))}_\text{ \ding{182}: Pruning task}; 
& \quad 
  \st &  \underbrace{\displaystyle {\btheta}^*(\mask) = \argmin_{\btheta \in \mathbb{R}^n} \overbrace{ \ell(\mask\odot \btheta) + \frac{\gamma}{2} \|\btheta\|_2^2}^\text{$\Def g(\mask, \btheta)$}}_\text{\ding{183}: Sparsity-fixed model retraining},
    \end{array}
    \label{eq: prob_bi_level}
    \vspace*{-6mm}
\end{align}
}}%
where $\ell$ denotes the training loss (\textit{e.g.}, cross-entropy), 
$\mathbf m$ and $\btheta$ are the upper-level and lower-level optimization variables respectively, 
${\btheta}^*(\mask)$ signifies the lower-level solution obtained by minimizing the objective function $g$ given the pruning mask $\mask$, 
and $\gamma > 0$ is a regularization parameter introduced to convexify the lower-level optimization so as to stabilize the gradient flow from   $\btheta^*(\mathbf m)$ to $\mathbf m$ and thus the convergence of BLO \cite{hong2020two, shaham2015understanding}.
In a sharp contrast to existing single-level optimization-based model pruning methods \cite{wen2016learning,zhang2018systematic,hoefler2021sparsity,wang2020proximal}, the BLO formulation \eqref{eq: prob_bi_level} brings in  \underline{two advantages}. 

\underline{First}, BLO has the flexibility to use   {mismatched} pruning and retraining objectives at the upper and lower optimization levels, respectively. 
{This flexibility allows us to regularize the lower-level training objective function  in \eqref{eq: prob_bi_level}  and customize the implemented optimization methods at both levels.}
To be more specific, one can  update the upper-level pruning mask $\mask$ using a data batch (called $\mathcal B_2$) distinct from the one (called $\mathcal B_1$) used for  obtaining the lower-level solution ${\btheta}^*(\mask)$. The resulting BLO procedure can then mimic the idea of  meta learning to improve model generalization \cite{deng2021meta}, where the lower-level problem fine-tunes $\btheta$ using $\mathcal B_1$, and the upper-level problem validates the generalization of the sparsity-aware finetuned model $(\mathbf m \odot \boldsymbol \theta^* (\mathbf m))$ using $\mathcal B_2$.

\underline{Second}, BLO enables us to explicitly model and optimize the coupling  between the retrained model weights $\btheta^*(\mathbf m)$ and the pruning mask $\mathbf m$ through the implicit gradient (IG)-based optimization routine. Here IG refers to  the gradient of the lower-level solution $\btheta^*(\mathbf m)$ with respect to (w.r.t.) the upper-level variable $\mathbf m$, and its derivation  calls the implicit function theory \cite{gould2016differentiating}.
The use of     IG makes our proposed BLO-oriented pruning \eqref{eq: prob_bi_level} significantly different from the greedy alternating minimization  \cite{csiszar1984information} that learns the upper-level and lower-level variables independently (\textit{i.e.}, minimizes one variable by fixing the other). We refer readers to the following section for the detailed IG theory.
 We will also show in Sec.\,\ref{sec: exp} that
 the pruning strategy from \eqref{eq: prob_bi_level} can outperform IMP in many pruning scenarios but is much more efficient as it does not call for the scheduler of iterative pruning ratios. 

\paragraph{Optimization foundation of {\biprune}.}
The key optimization challenge of solving the {\biprune} problem \eqref{eq: prob_bi_level} lies in the computation of IG (implicit gradient). Prior to developing an effective solution, we first elaborate on the \textit{IG  challenge}, the unique characteristic of BLO. In the context of gradient descent, the gradient of the   objective function in  \eqref{eq: prob_bi_level} yields

\vspace*{-5mm}
{
\small{
\begin{align}
\displaystyle
    \underbrace{\frac{d\ell(\mask \odot \btheta^*(\mask))}{d\mask}}_\text{Gradient of objective} = \nabla_{\mask} \ell(\mask \odot {\btheta}^*(\mask)) + \underbrace{
    \frac{{d ( {\btheta}^*(\mask)^{\top} )
    }
    }{d \mask}}_{\text{{IG}}} \nabla_{\btheta}\ell (\mask \odot \btheta^*(\mask)),
    \label{eq: GD_bi_level}
\end{align}}}%
where $\nabla_{\mask}$ and $\nabla_{\btheta}$  denote the \textit{partial derivatives} of the bi-variate function $\ell(\mask \odot \btheta)$  w.r.t. the variable $\mask$ and $\btheta$ respectively, 
$d \btheta^\top / d \mask  \in \mathbb R^{n \times n} $ denotes the vector-wise \textit{full derivative},  
and for ease of notation, we will omit
the transpose $\top$ when the context is clear.
In \eqref{eq: GD_bi_level}, the IG challenge refers to the demand for   computing the full gradient of  the implicit function $\btheta^*(\mask) = \argmin_{\btheta} g(\mask, \btheta)$ w.r.t. $\mask$, where recall from \eqref{eq: prob_bi_level}  that $ g(\mask, \btheta) \Def  \ell(\mask\odot \btheta) + \frac{\gamma}{2} \|\btheta\|_2^2 $.

Next,
we derive the IG formula following the 
rigorous 
implicit function theory \cite{gould2016differentiating,hong2020two,rajeswaran2019meta}.
Based on the fact that $\btheta^*(\mask)$ satisfies the stationarity condition for the lower-level objective function in \eqref{eq: GD_bi_level},  it is not difficult to obtain that  (see derivation in  Appendix\,\ref{eq: app_IG})

\vspace*{-5mm}
{\small
\begin{align}
\displaystyle  \frac{d\btheta^*(\mask)}{d\mask} 
 =  -   \nabla^2_{\mask \btheta} \ell (\mask \odot \btheta^* ) [ \nabla^2_{\btheta } \ell(\mask \odot \btheta^*) + \gamma \mathbf I ]^{-1},
 \label{eq: exact_IG}
\end{align}
}%
where   $\nabla_{\mask \btheta}^2 \ell $ and $\nabla_{\btheta}^2 \ell $ denote the second-order partial derivatives of $\ell$ respectively, and $(\cdot)^{-1}$ denotes the matrix inversion operation. 

Yet, the exact   IG formula \eqref{eq: exact_IG} remains difficult to calculate due to the presence of matrix inversion and second-order partial derivatives.
To  simplify it, 
we impose the Hessian-free assumption, 
$ \nabla^2_{\btheta} \ell = \mathbf 0$, which  is mild in general; For example,   the decision boundaries of neural networks with ReLU activations are piece-wise linear in  a tropical hyper-surface \cite{alfarra2020decision}, and this assumption 
has been widely used in BLO-involved applications such as meta learning \cite{finn2017model} and adversarial learning \cite{zhang2021bat}. Given $ \nabla^2_{\btheta } \ell = \mathbf 0$,  the matrix inversion in \eqref{eq: exact_IG} can   be then mitigated, leading to the IG formula

\vspace*{-5mm}
{\small
\begin{align}
    \frac{d\btheta^*(\mask)}{d\mask} 
    =  -  \frac{1}{\gamma} \nabla^2_{\mask \btheta} \ell (\mask \odot \btheta^* ).
    \label{eq: IG_simplify}
\end{align}
\vspace*{-3mm}
}%

At the first glance, the computation of the simplified IG   \eqref{eq: IG_simplify} still requires the mixed (second-order) partial derivative $\nabla^2_{\mask \btheta} \ell$. However, {\biprune} is a special class of BLO problems with bi-linear variables 
$(\mask \odot \btheta)$. Based on this bi-linearity, we can prove that  IG  in \eqref{eq: IG_simplify} can be \textit{analytically} expressed using only \textit{first-order} derivatives; see the following theorem. 

\begin{myprop}
\label{thr: ig}
Assuming $\nabla^2_{ \btheta} \ell = 0$ and defining {$\gradzsimple \Def \gradz$}, the implicit gradient \eqref{eq: IG_simplify} is then given by

\vspace*{-5mm}
{\small
\begin{align}
\displaystyle
\frac{d\btheta^*(\mask)}{d\mask} 
= -  \frac{1}{\gamma} \mathrm{diag}( \nabla_{\mathbf z} \ell(\mathbf z)
);
\label{eq: IG_final}
\end{align}
}%
Further, the   gradient of the  objective function given by \eqref{eq: GD_bi_level} becomes

{\vspace*{-5mm}
{
\small
\begin{align}
\displaystyle
    \frac{d \ell(\mask \odot \btheta^*)}{d \mask} = ( \btheta^* 
    - 
    \frac{1}{\gamma} \mask \odot \gradzsimple) \odot \gradzsimple,
    \label{eq: upper_update_final}
\end{align}}}%
where
$\odot$ denotes the element-wise multiplication.
\end{myprop}
\textbf{Proof}: Using chain-rule, we can obtain that

\vspace*{-6mm}
{\small
\begin{align}
&    \nabla_{\btheta}\ell(\mask \odot \btheta^*) = 
\mathrm{diag}(\mathbf m)
\gradzsimple = \mask \odot \gradzsimple;
    \label{eq: chainrule_theta} \\
\text{similarly,   } & \nabla_\mask \ell(\mask \odot \btheta^*) 
    = \mathrm{diag}(\btheta^*) \gradzsimple 
    = \btheta^* \odot \gradzsimple \label{eq: chainrule_mask}
\end{align}
}
where $\mathrm{diag}(\cdot)$ represents a diagonal matrix with $\cdot$ being the main diagonal vector.
Further, we can convert \eqref{eq: IG_simplify} to

\vspace*{-6mm}
{
\small
\begin{align}
\displaystyle
    \nabla^2_{{\mask} {\btheta}} \ell({\mask}  \odot {\btheta^*}) 
    & \overset{\eqref{eq: chainrule_theta}}{=} \nabla_{{\mask}} \left [ {\mask} \odot  \nabla_{\mathbf z} \ell(\mathbf z) \right ]
    \nonumber \overset{\text{chain rule}}{=} 
    \mathrm{diag}( {\gradzsimple}) 
    + 
    \mathrm{diag}({\mask}) [ \nabla_{{\mask}} ( \nabla_{\mathbf z} \ell(\mathbf z) ) ]
     \nonumber \\
    & \overset{\eqref{eq: chainrule_mask}}{=}  \mathrm{diag}( {\gradzsimple}) 
    + 
    \mathrm{diag}({\mask}) [ \mathrm{diag}(\btheta^*) \nabla^2_{\mathbf z } \ell(\mathbf z) ] = \mathrm{diag}(\gradzsimple) ,\label{eq: ezBiP}
 \end{align}}%
where the last equality holds due to the Hessian-free assumption. With \eqref{eq: ezBiP} and \eqref{eq: IG_simplify} we can prove \eqref{eq: IG_final}.

Next, substituting the IG \eqref{eq: IG_final}  to the upper-level gradient \eqref{eq: GD_bi_level}, we  obtain that

\vspace*{-3.9mm}
{
\small
\begin{align}
\displaystyle
     \frac{d \ell(\mask \odot \btheta^*)}{d \mask} 
    & \,\,\,\,=\nabla_{\mask}\ell({\mask} \odot {\btheta}^*) - 
    {\frac{1}{\gamma}
    \gradzsimple
    \odot \nabla_{{\btheta}} \ell({\mask} \odot {\btheta^*})
    } \nonumber \\
    &
    \underset{ }{\overset{\eqref{eq: chainrule_theta}, \eqref{eq: chainrule_mask}}{=}}  \btheta^* \odot \gradzsimple
    - \frac{1}{\gamma} \gradzsimple
    \odot
    ( \mask \odot \gradzsimple
    ) 
    =  (\btheta^* - \frac{1}{\gamma} \mask \odot \gradzsimple
    ) \odot \gradzsimple, \nonumber
\end{align}
}%
which leads to \eqref{eq: upper_update_final}. 
The proof is now complete. \hfill $\square$ 

The key insight drawn from Prop.\,\ref{thr: ig} is that the bi-linearity of pruning variables (\textit{i.e.}, $\mask \odot \boldsymbol \theta^*$) makes the  IG-involved gradient \eqref{eq: GD_bi_level} easily solvable, and the   computational complexity is almost the same as that of computing the first-order gradient $\nabla_{\mathbf z} \ell (\mathbf z)$ just once, as supported by   \eqref{eq: upper_update_final} 


\paragraph{{\biprune} algorithm and implementation.}
We next 
formalize the {\biprune} algorithm 
based on Prop.\,\ref{thr: ig} and the alternating gradient descent  based BLO solver~\cite{hong2020two}. At iteration $t$, there are two main steps.

\ding{72} \textit{Lower-level SGD for model retraining}: 
Given $\mask^{(t-1)}$, $\btheta^{(t-1)}$, and 
$
\mathbf z^{(t-1)} 
\Def \mask^{(t-1)} \odot \btheta^{(t-1)}$, we update $\btheta^{(t)}$ by randomly selecting a data batch with the learning rate $\alpha$ and applying SGD (stochastic gradient descent) to the lower-level problem of \eqref{eq: prob_bi_level},%

\vspace*{-5mm}
{\small 
\begin{align}
    \btheta^{(t)} & = \btheta^{(t-1)} - \alpha \nabla_{\btheta} g(\mask^{(t-1)}, \btheta^{(t-1)})  
   \overset{
   \eqref{eq: chainrule_theta}}{=}
    \btheta^{(t-1)} - \alpha [ \mask^{(t-1)} \odot 
    \nabla_{\mathbf z} \ell(\mathbf z )\left | \right._{\mathbf z = \mathbf z^{(t-1)}} + \gamma \btheta^{(t-1)} ],
    \tag{$\btheta$-step}
    \label{eq: theta_step}
\end{align}}%

\ding{72} \textit{Upper-level {SPGD} for pruning}: Given $\mask^{(t-1)}$, $ \btheta^{(t)}$, and $\mathbf z^{(t+1/2)} \Def \mask^{(t-1)} \odot \btheta^{(t)}$, we update   $\mask$ using SPGD (stochastic projected gradient descent) along the IG-enhanced descent direction \eqref{eq: GD_bi_level},%

\vspace*{-5mm}
{\small
\begin{align}
    \mask^{(t)} = & \mathcal P_{\mathcal S} \left [ 
    \mask^{(t-1)}  - \beta  \frac{d \ell(\mask \odot \btheta^{(t)})}{d \mask} \left | \right._{\mask =     \mask^{(t-1)} }
    \right ] \nonumber \\
\overset{\eqref{eq: upper_update_final}}{=}    
    &\mathcal P_{\mathcal S} \left [ 
    \mask^{(t-1)}  - \beta  \left ( \btheta^{(t)} - 
    \frac{1}{\gamma} \mask^{(t-1)} \odot \gradzsimple \left | \right._{\mathbf z = \mathbf z^{(t+1/2)}} \right ) \odot \gradzsimple  \left | \right._{\mathbf z = \mathbf z^{(t+1/2)}}
    \right ],
    \tag{$\mask$-step}
    \label{eq: m_step}
    \vspace{-7mm}
\end{align}}%
where $\beta > 0$ is the upper-level learning rate, and $\mathcal P_{\mathcal S}(\cdot)$ denotes the Euclidean projection   onto the constraint set $\mathcal S$ given by
$\mathcal S = \{ \mathbf m\,|\, \mathbf m \in \{ 0, 1\}^n, \mathbf  1^T \mathbf m \leq k \}$ in \eqref{eq: prob_bi_level} and is achieved by the top-$k$ hard-thresholding operation as will be detailed later.

\begin{figure}[t]
\centerline{
\includegraphics[width=1.0\textwidth]{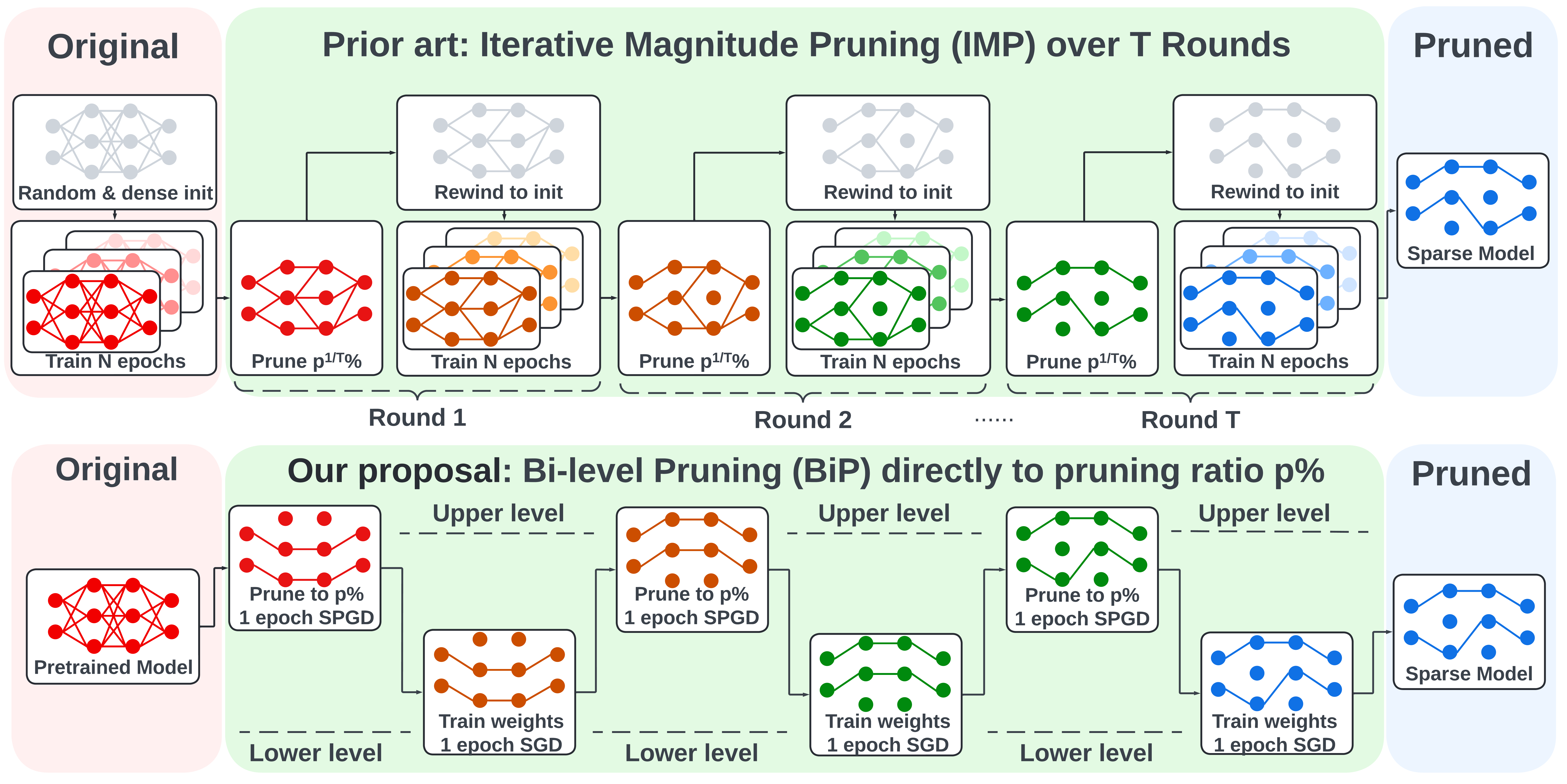}
}
\caption{\footnotesize{
Visualization of pruning pipeline comparison   between IMP and {\biprune}.   
Edge refers to the mask update and   color refers to the weight update.
}}
\label{fig: alg_compare}
\vspace*{-3mm}
\end{figure}

In {\biprune}, the \eqref{eq: theta_step} and \eqref{eq: m_step} steps execute iteratively. 
For clarity,   Fig.\,\ref{fig: alg_compare} shows the difference between the pruning pipelines of {\biprune} and IMP. In contrast to IMP that progressively prunes and retrains a model with a growing pruning ratio, {\biprune} directly prunes the model to the targeted sparsity level without involving costly re-training process.
In practice, we find that both the upper- and lower-level optimization routines of  {\biprune} converge very well (see Fig.\,\ref{fig: convergence} and Fig.\,\ref{fig: IoU_Trajectory_rebuttal}).
It is also worth noting that both \eqref{eq: theta_step} and \eqref{eq: m_step} only require the first-order information $\nabla_{\mathbf z} \ell (\mathbf z)$, 
demonstrating that {\biprune} can be conducted as efficiently as first-order optimization.
\CR{In Fig.\,\ref{fig: algorithm overview}, we highlight the algorithmic details on the {\biprune} pipeline.}
We present more implementation details of {\biprune} below and {refer readers to  Appendix\,\ref{app: algorithm}  for a detailed algorithm description}.

\ding{70} \textit{Discrete optimization over $\mask$:} We follow the  {`convex relaxation + hard thresholding'} mechanism used in \cite{sehwag2020hydra,ramanujan2020sHiddenSubnetwork}.
Specifically, we relax the binary masking variables   to continuous masking   scores $\mathbf m \in [\mathbf 0, \mathbf 1]$. We then acquire  loss gradients at  the backward pass based on the relaxed  $\mathbf m$. At the forward pass, we project it  onto the discrete constraint set $\mathcal S$ using the hard thresholding operator, where   the top $k$ elements  are set to $1$s and the others to $0$s. See Appendix\,\ref{app: algorithm} for more discussion.

\ding{70} \textit{Data batch selection for lower-level and upper-level optimization:} We adopt different data batches (with the same batch size) when implementing \eqref{eq: theta_step} and \eqref{eq: m_step}. This is one of the advantages of the BLO formulation, which enables the flexibility to customize the lower-level and upper-level problems. The use of diverse data batches  is beneficial to generalization as shown in \cite{deng2021meta}. 
 
\ding{70} \textit{Hyperparameter tuning:} As described in \eqref{eq: theta_step}-\eqref{eq: m_step},  {\biprune} needs to set two learning rates $\alpha$ and $\beta$ for lower-level and upper-level optimization, respectively. We choose $\alpha = 0.01$ and $\beta = 0.1$ in all experiments, where we adopt the mask learning rate $\beta$ from Hydra\,\cite{sehwag2020hydra} and set a smaller lower-level learning rate $\alpha$, as $\btheta$ is initialized by a pre-trained dense model. We show ablation study on $\alpha$ in Fig.\,\ref{fig: ablation_study}(c).
BLO also brings in the low-level convexification parameter $\gamma $. We set $\gamma = 1.0 $ in   experiments and refer readers to {Fig.\,\ref{fig: ablation_study}(b)} for a sanity check. 

\ding{70} \textit{One-step vs. multi-step SGD:}
In  \eqref{eq: theta_step}, the   one-step SGD is used and helps reduce the computation overhead. In practice, we also find that the one-step SGD is sufficient:  The use of multi-step SGD in {\biprune} does not yield much significant improvement over the one-step version; see {Fig.\,\ref{fig: ablation_study}(a)}.

\ding{70} \textit{Extension to structured pruning:}
We formulate and solve the {\biprune} problem in the context of unstructured (element-wise)  weight pruning. However, if define the pruning mask $\mask$ \emph{w.r.t.} model's structural units  (\textit{e.g.}, filters), {\biprune} is easily applied to structured pruning (see Fig.\,\ref{fig: structured_performance_cifar} and Fig.\,\ref{fig: structured_performance_cifar_appendix}).

\section{Experiments}
\label{sec: exp}
In this section, we present extensive experimental results to show the effectiveness of {\biprune}  
across multiple model architectures,  various datasets, and different pruning setups. Compared to IMP, one-shot pruning, and optimization-based pruning baselines, 
we find that {\biprune} can find better winning tickets in most cases and is computationally   efficient. 

\begin{figure}[t]
\vspace*{-3mm}
\centerline{
\begin{tabular}{cccc}
    \hspace*{-2mm} \includegraphics[width=.25\textwidth,height=!]{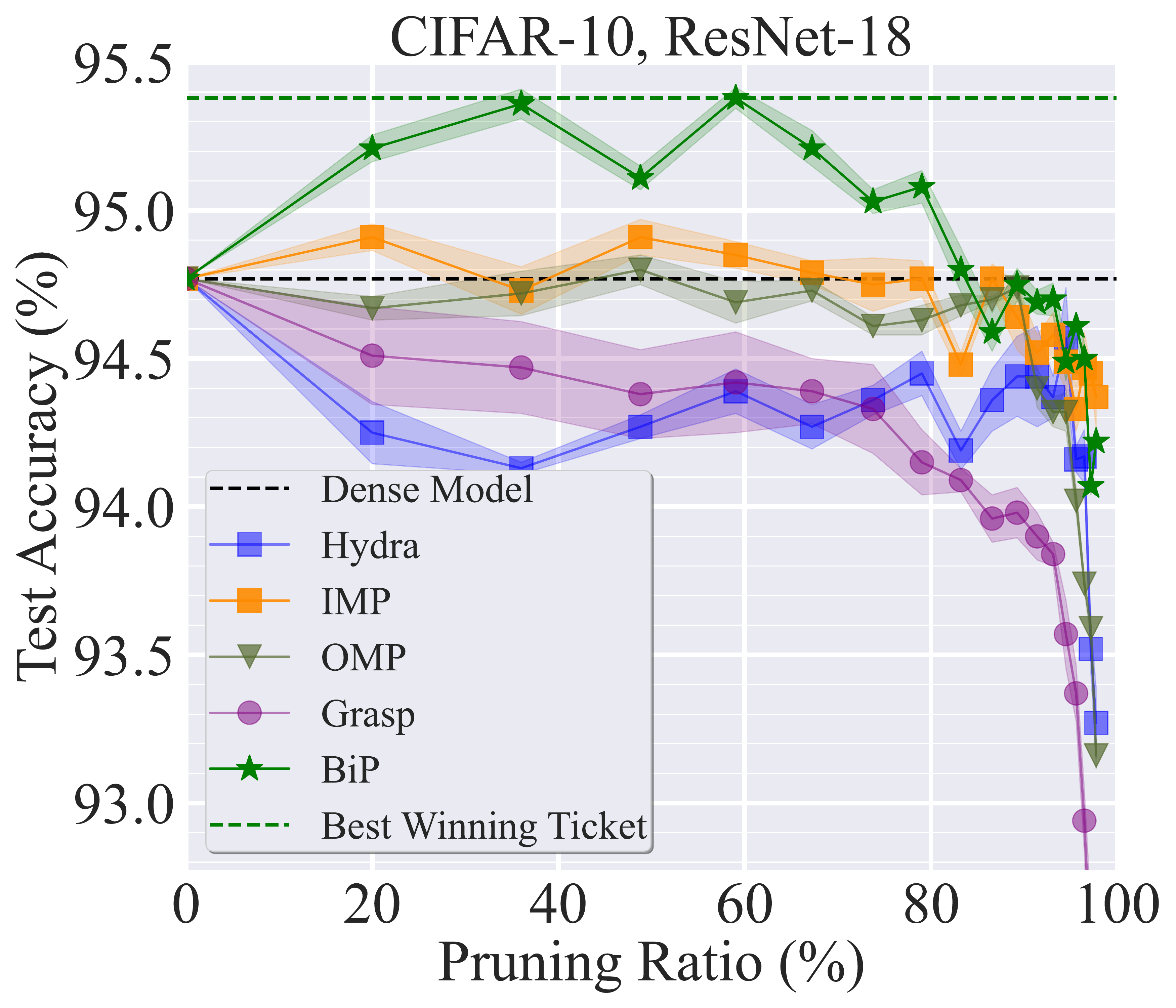} &
    \hspace*{-5mm}  \includegraphics[width=.25\textwidth,height=!]{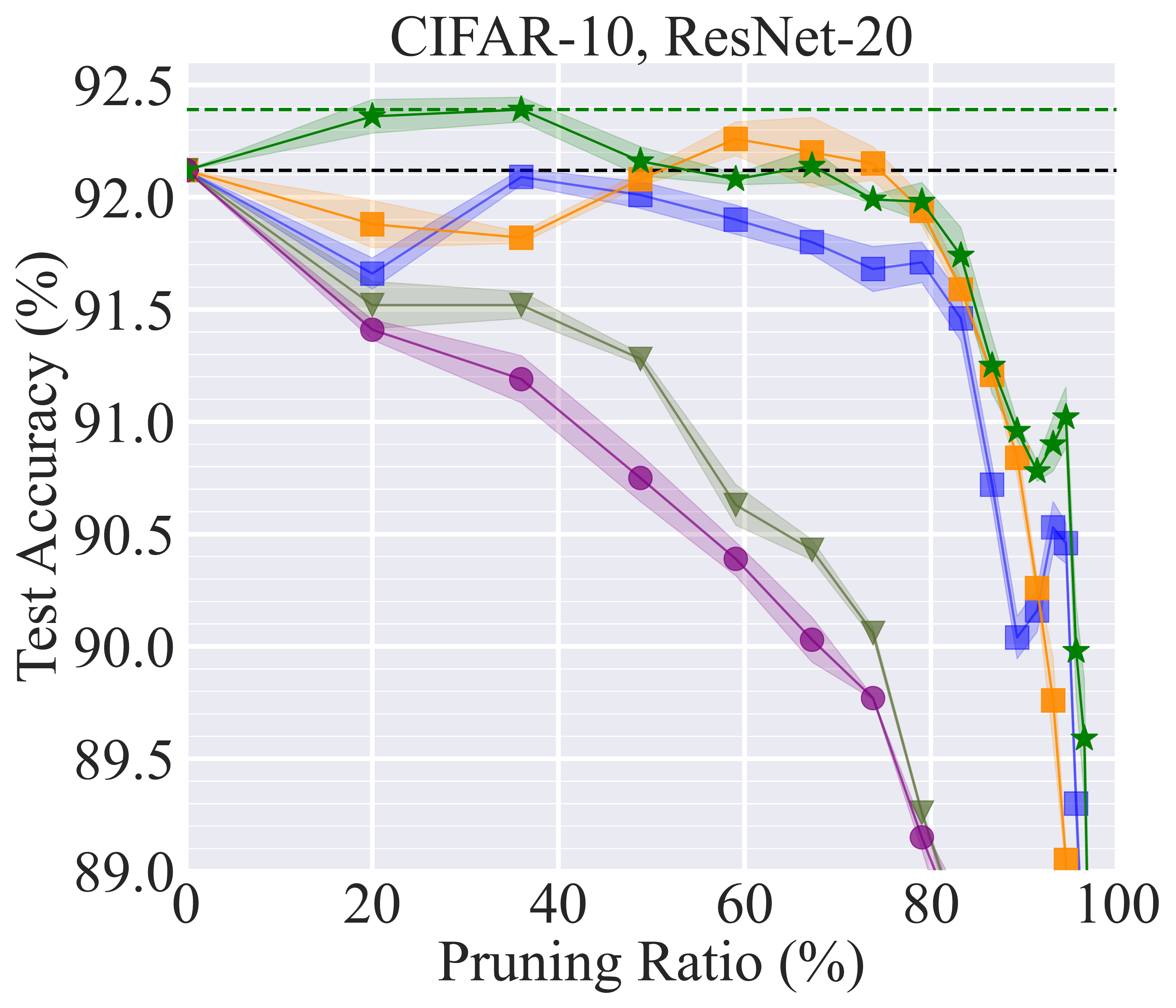} &
    \hspace*{-5mm}  \includegraphics[width=.25\textwidth,height=!]{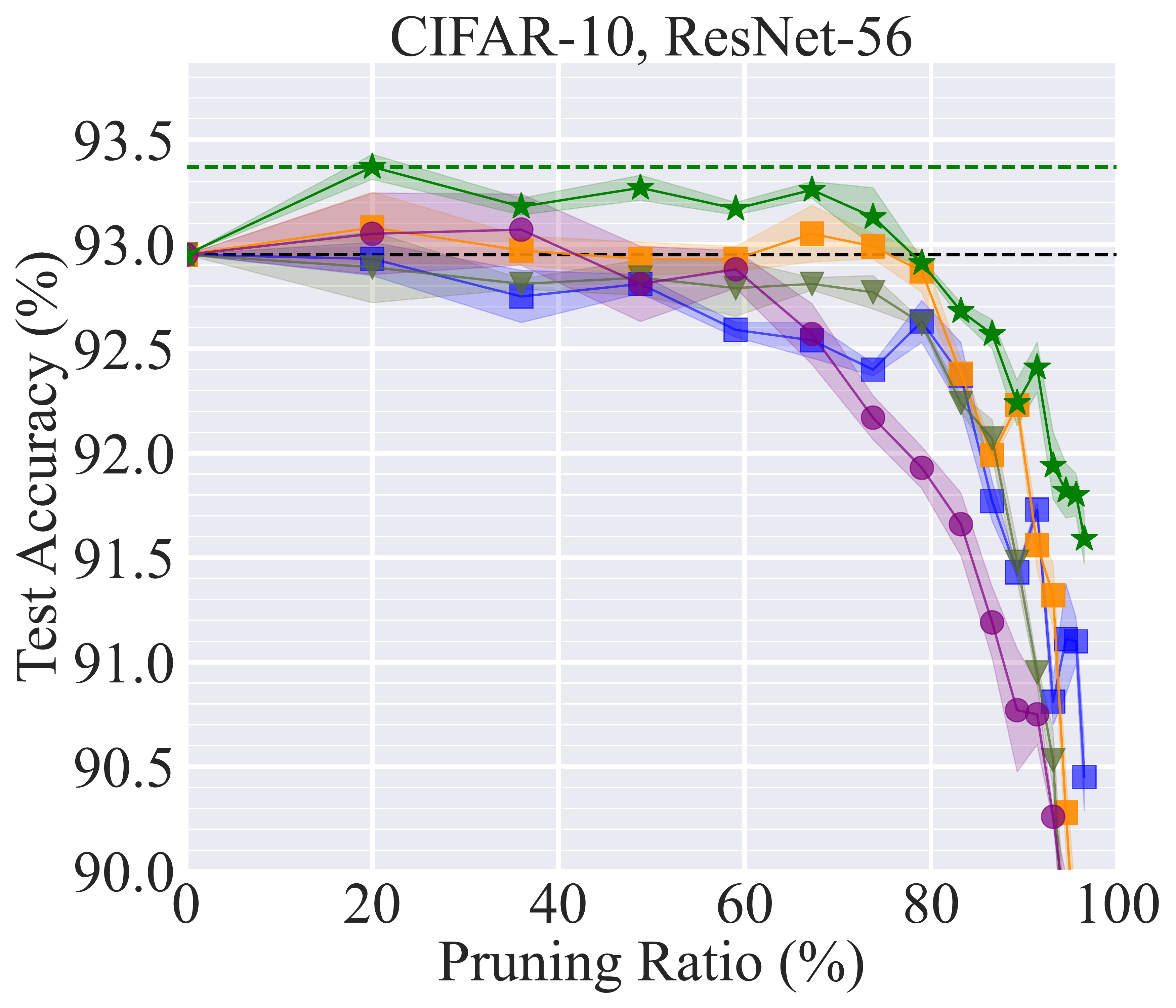} &
    \hspace*{-5mm} \includegraphics[width=.25\textwidth,height=!]{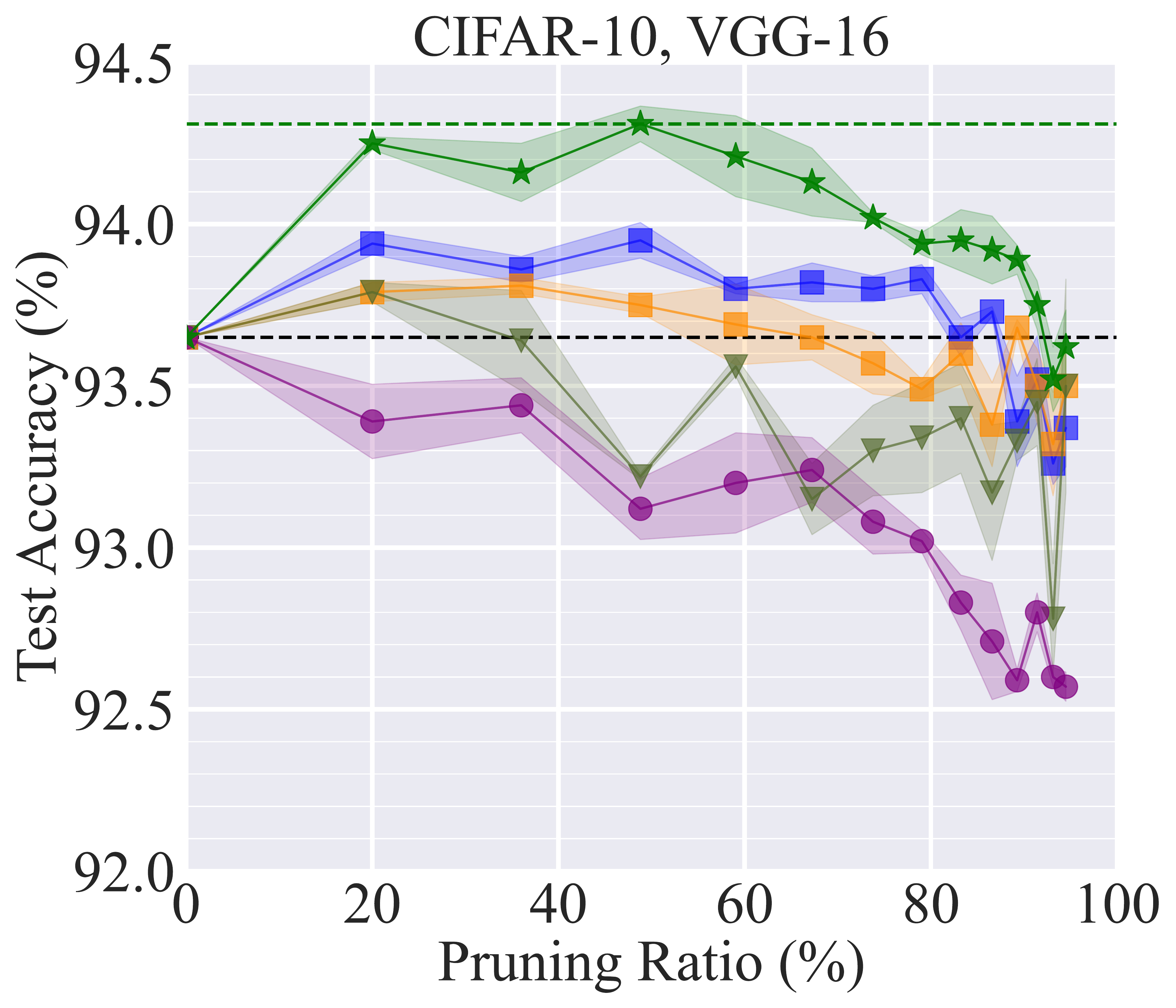} \\
    
    \hspace*{-2mm} \includegraphics[width=.25\textwidth,height=!]{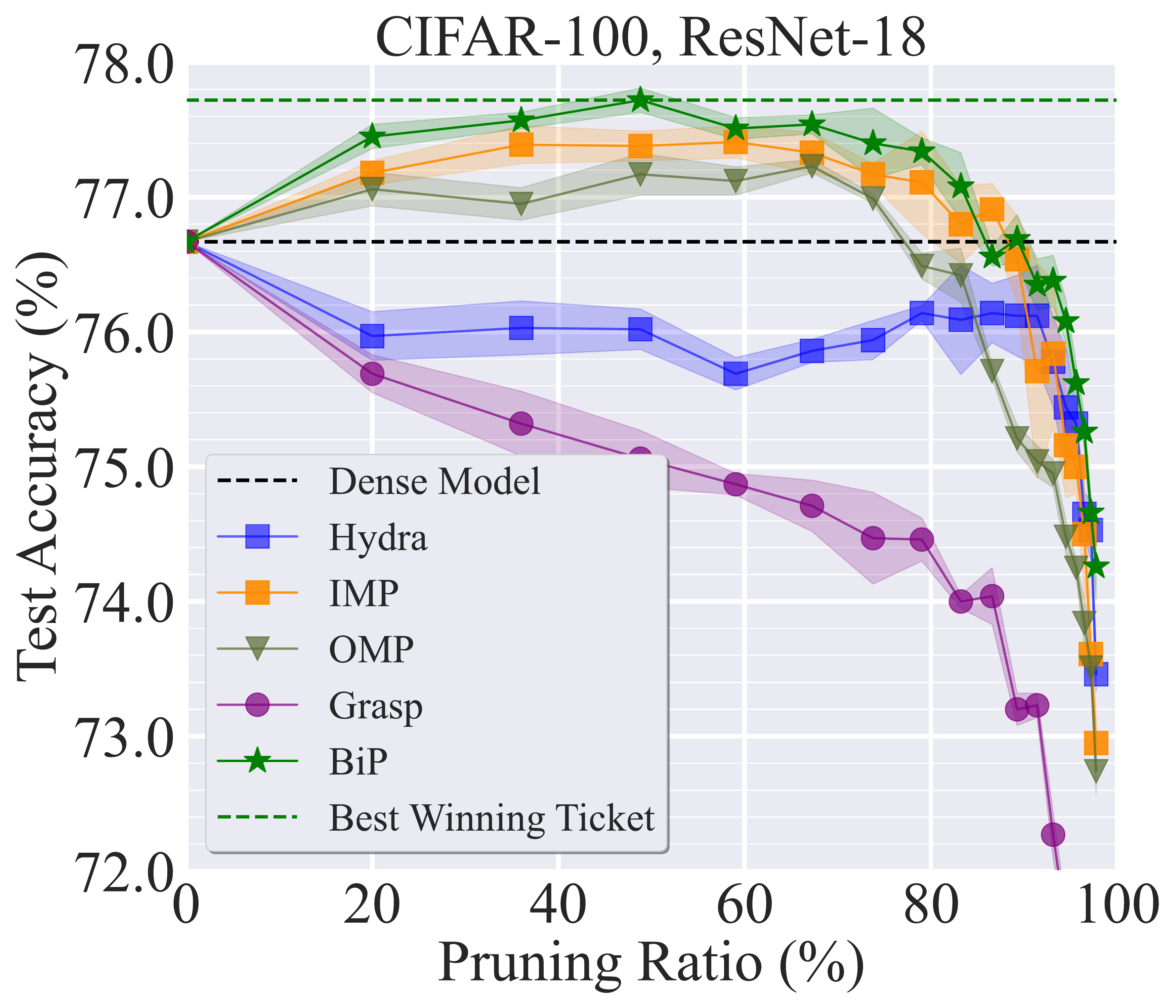} &
    \hspace*{-5mm}  \includegraphics[width=.25\textwidth,height=!]{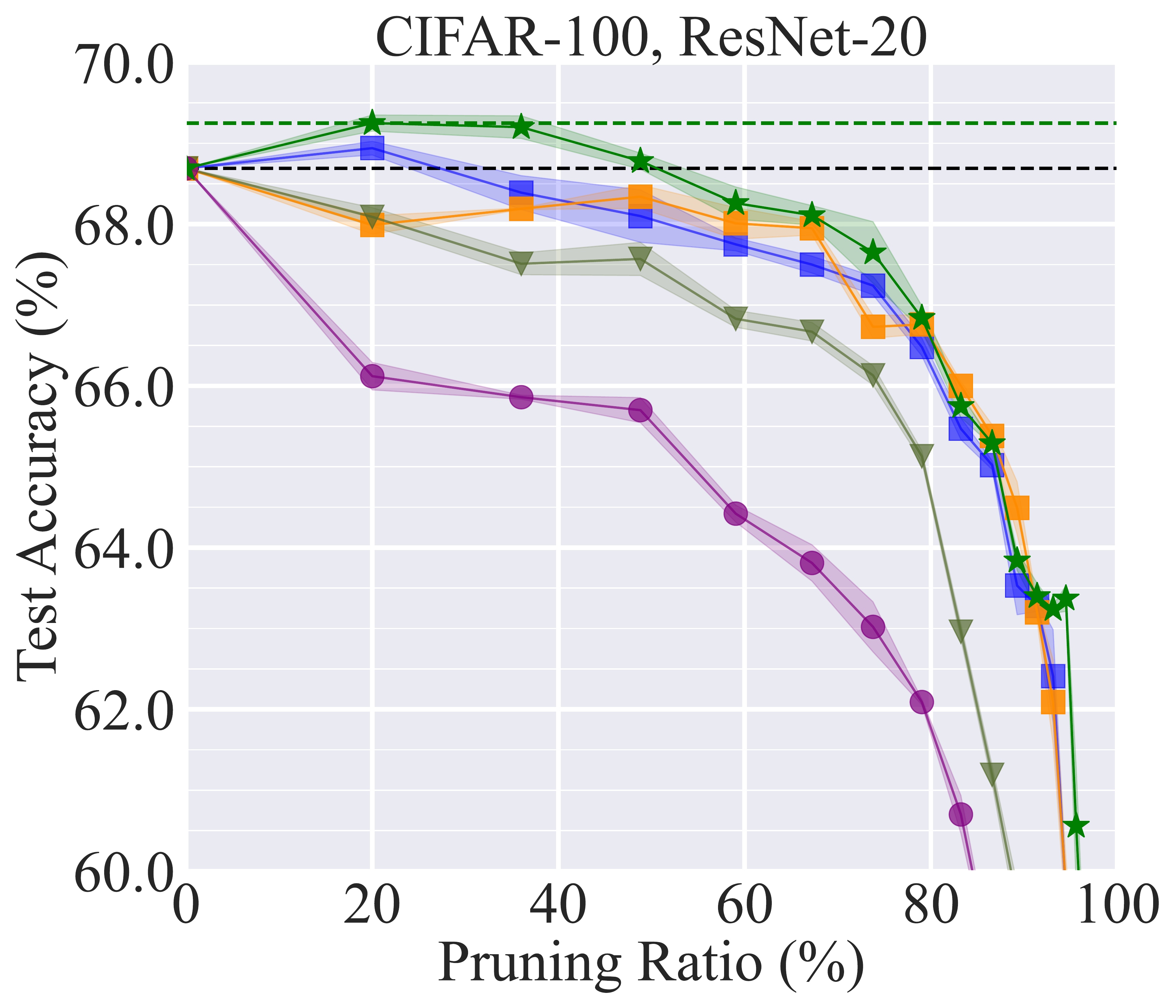} &
    \hspace*{-5mm}  \includegraphics[width=.25\textwidth,height=!]{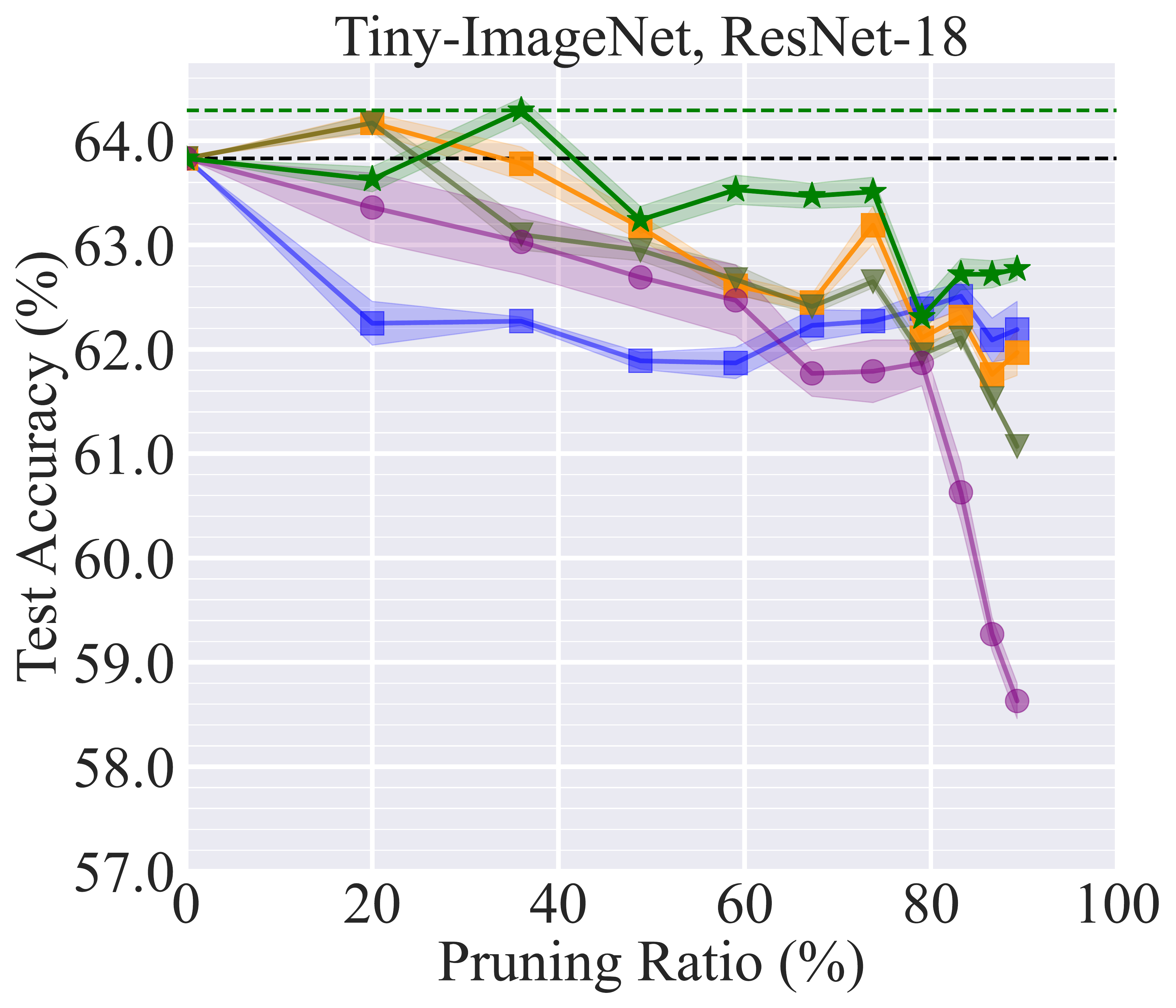} &
    \hspace*{-5mm} \includegraphics[width=.25\textwidth,height=!]{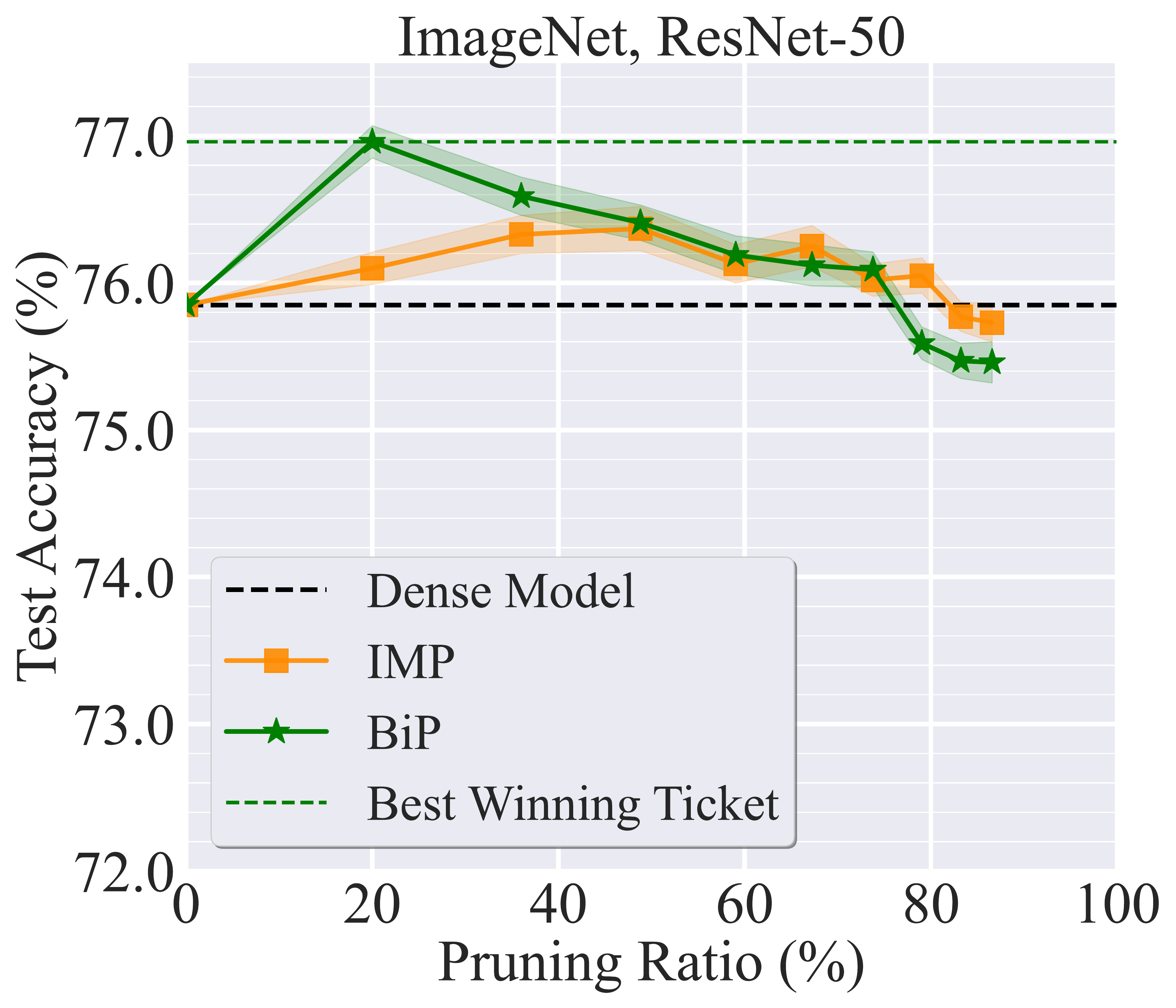}
\end{tabular}}
\caption{\footnotesize{Unstructured pruning trajectory given by test accuracy (\%) vs. sparsity (\%) on various dataset-model pairs.
The proposed \textcolor{Green}{\biprune} is compared with \textcolor{blue}{\hydra}\,\cite{sehwag2020hydra}, \textcolor{orange}{IMP}\,\cite{ma2021sanity}, \textcolor{ompcolor}{OMP}\,\cite{ma2021sanity}, \textcolor{Purple}{\grasp}\,\cite{wang2020pick}. And the performance of   dense model and that of the best winning ticket  are marked using   dashed lines in each plot. 
The solid line and shaded area of each pruning method represent  the mean and variance of test accuracies over 3 independent trials. We observe that
{\biprune}  consistently outperforms the other baselines.
Note in the (ImageNet, ResNet-50) setting, we only compare {\biprune} with our strongest baseline IMP due to computational resource constraints.
}}
\label{fig: unstructured_performance_cifar}
\end{figure}

\subsection{Experiment Setup}
\paragraph{Datasets and models.}
Following the pruning benchmark in \cite{ma2021sanity}, 
we consider 4 datasets including CIFAR-10\,\cite{krizhevsky2009learning}, CIFAR-100\,\cite{krizhevsky2009learning}, Tiny-ImageNet\,\cite{le2015tiny}, ImageNet\,\cite{deng2009imagenet}, and 5
architecture types  including  ResNet-20/56/18/50
and VGG-16 \cite{he2016deep,simonyan2014very}. 
Tab.\,\ref{table: impl_details} summarizes these datasets and model configurations and setups.  


\paragraph{Baselines,  training, and evaluation.}
As baselines, we mainly focus on 4 SOTA pruning methods, \ding{172} IMP \cite{frankle2018lottery}, \ding{173} OMP \cite{frankle2018lottery},
 \ding{174} {\grasp} \cite{wang2020pick} (a one-shot pruning method by analyzing gradient flow at initialization), and \ding{175} {\hydra} \cite{sehwag2020hydra} (an optimization-based pruning method that optimizes masking scores). It is worth noting that there exist various implementations of IMP, \textit{e.g.}, specified by different learning rates and model initialization or `rewinding' strategies \cite{Renda2020Comparing}. To make a fair comparison, we follow the recent IMP benchmark in \cite{ma2021sanity}, which can find the best winning tickets over current heuristics-based pruning baselines.
We also remark that {\hydra} is originally proposed for improving the adversarial robustness of a pruned model, but it can be easily customized for standard pruning when setting the adversary's strength as $0$ \cite{sehwag2020hydra}. 
We choose {\hydra} as a baseline because it can be regarded as a single-level variant of {\biprune} with
post-optimization weight retraining.
When implementing {\biprune}, unless specified otherwise, we use the $1$-step SGD in \eqref{eq: theta_step}, and set the learning rates $(\alpha, \beta)$ and the lower-level regularization parameter $\gamma$ as described in the previous section. 
When implementing baselines, we follow their official repository setups.
We evaluate the performance of all methods mainly from two perspectives: 
(1) the test accuracy of the sub-network, and (2) the runtime of pruning to reach the desired sparsity. 
We refer readers to Tab.\,\ref{tab: training_details} and Appendix\,\ref{app: training} for more training and evaluation details, such as training epochs and learning rate schedules.
\subsection{Experiment Results}

\begin{figure}[t]
\vspace{-0.5mm}
\centerline{
\begin{tabular}{cccc}
    \hspace*{-2mm} \includegraphics[width=.25\textwidth,height=!]{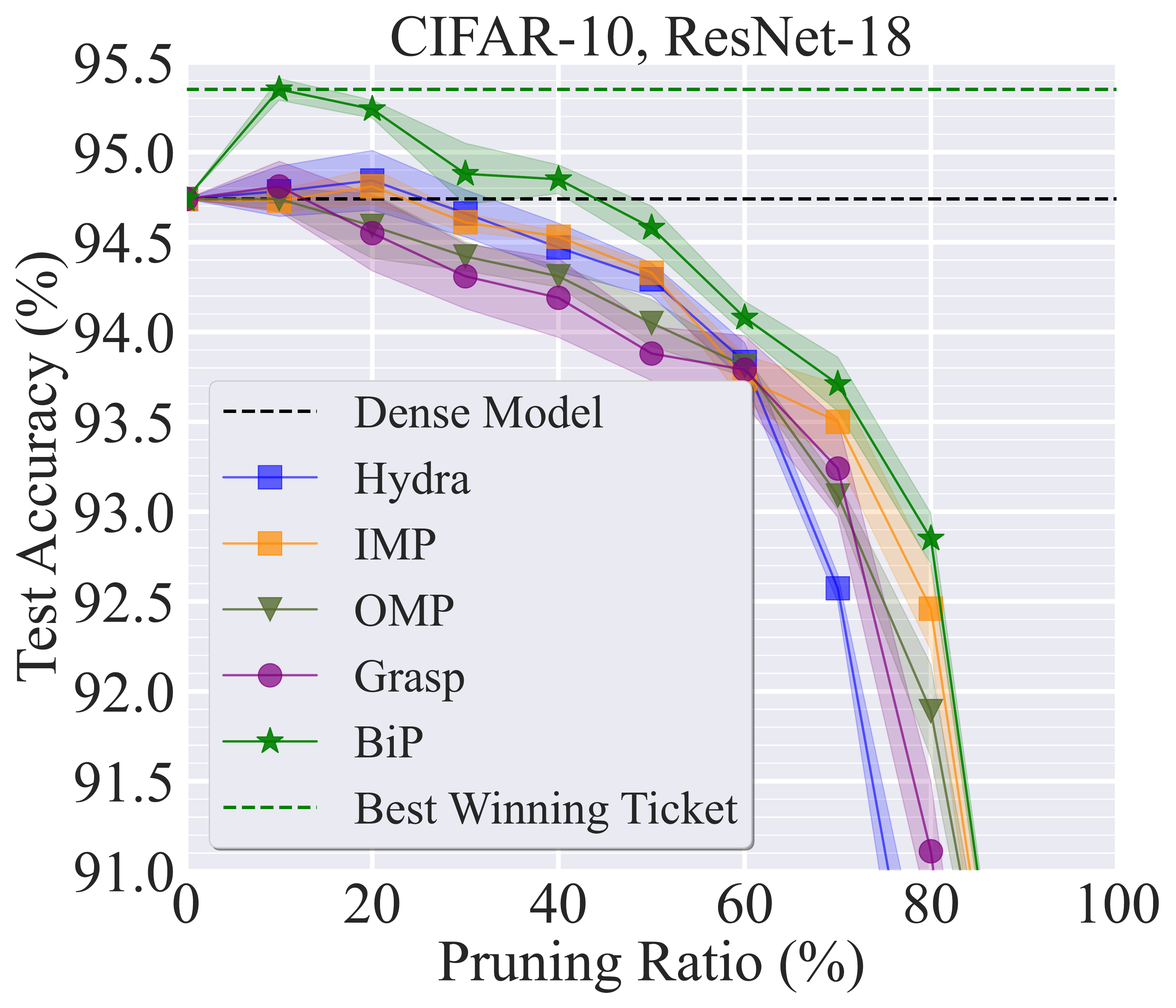} &
    \hspace*{-5mm}  \includegraphics[width=.25\textwidth,height=!]{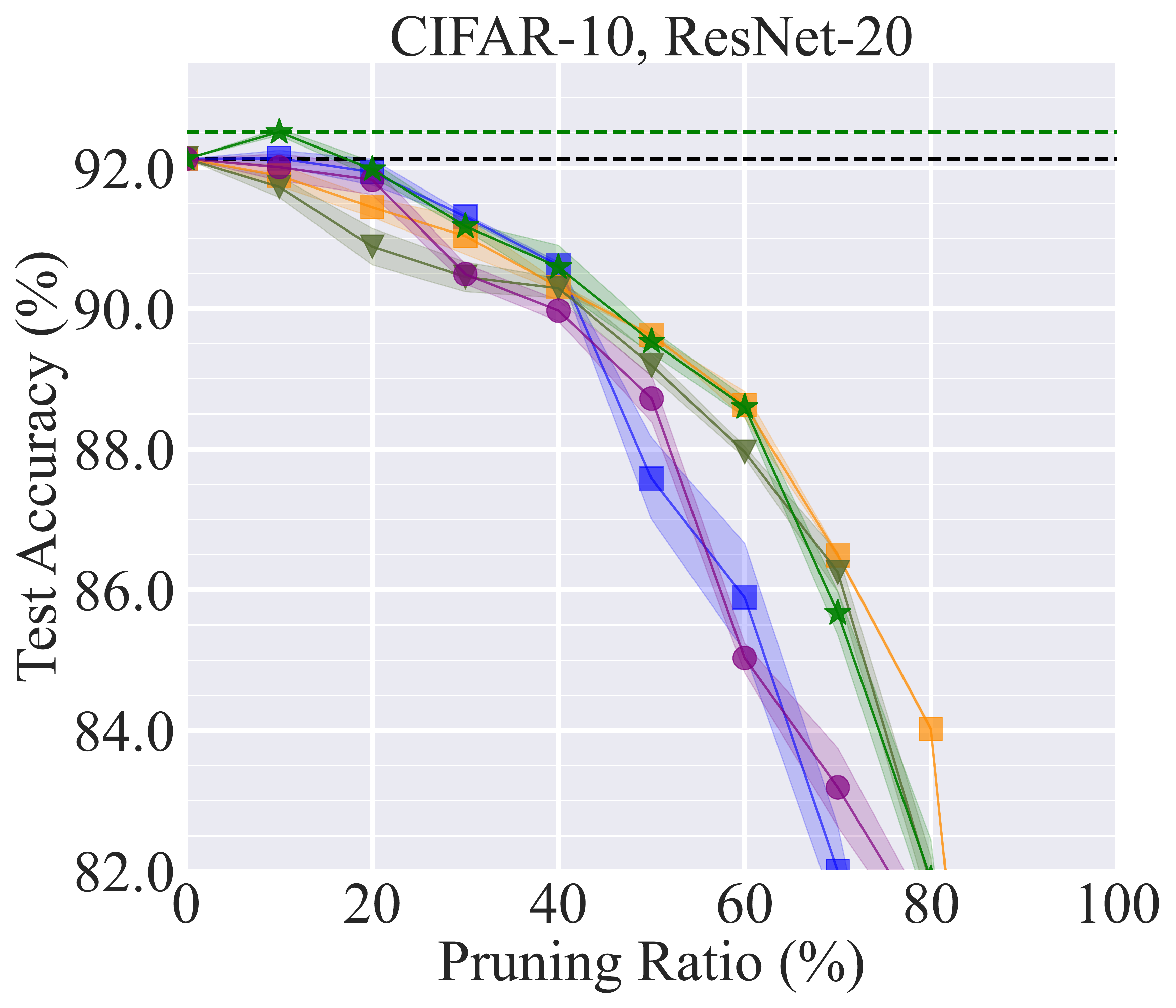} &
    \hspace*{-5mm}  \includegraphics[width=.25\textwidth,height=!]{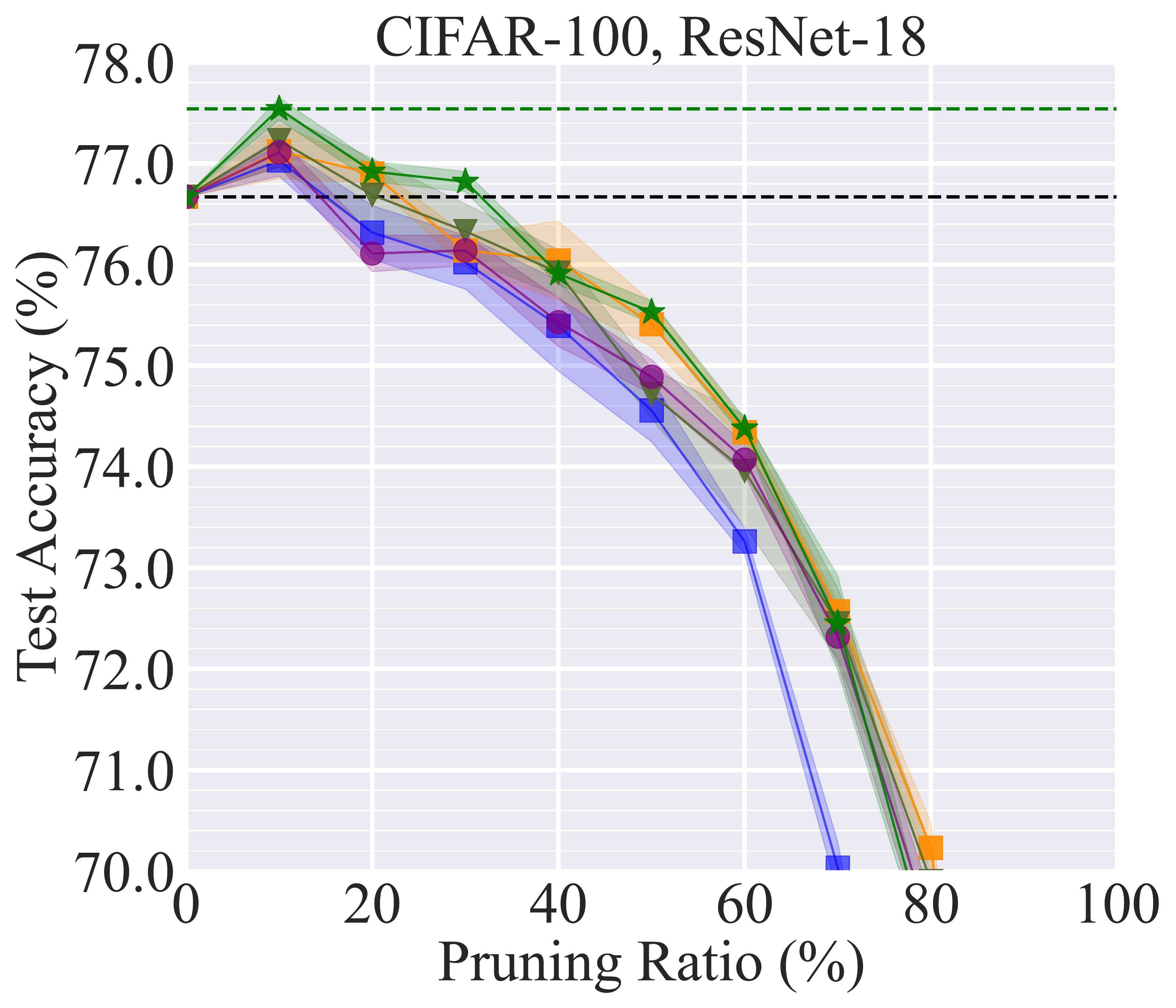} &
    \hspace*{-5mm}  \includegraphics[width=.25\textwidth,height=!]{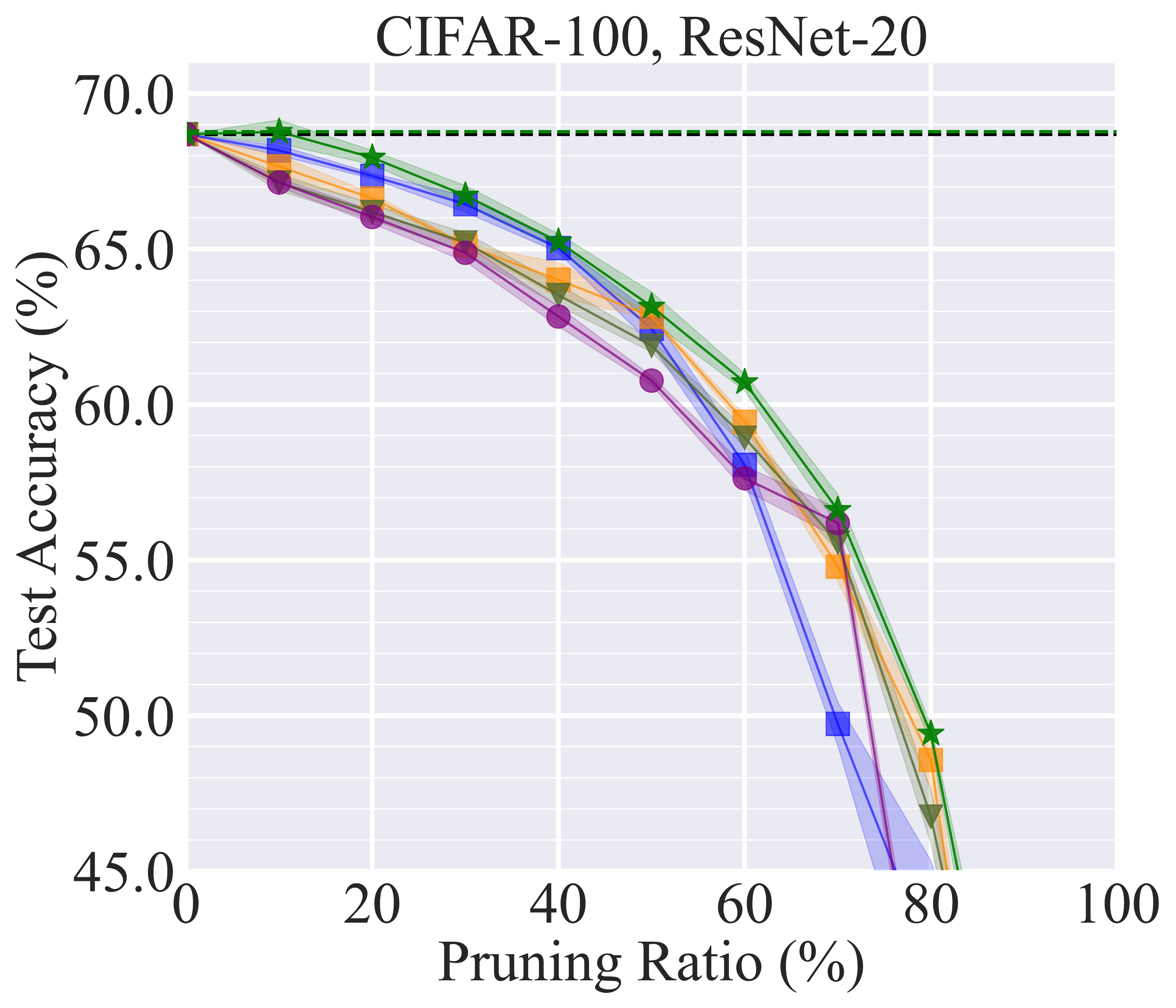} \\
    
\end{tabular}}
\caption{\footnotesize{Filter pruning given by test accuracy  (\%) vs. pruning ratio (\%). 
   The visual presentation setting is  the same as Fig.\,\ref{fig: unstructured_performance_cifar}. We observe that {\biprune}   identifies winning tickets of structured pruning in certain sparsity regimes.
    }}
\label{fig: structured_performance_cifar}
\end{figure}

\paragraph{\CR{{\biprune}   identifies high-accuracy subnetworks.}}
In what follows, we look at the quality of winning tickets identified by {\biprune}.
\textit{Two  key observations} can be drawn from our results: (1)~{\biprune}   finds winning tickets of higher accuracy and/or higher  sparsity than  the baselines in most cases (as shown in Fig.\,\ref{fig: unstructured_performance_cifar} and Tab.\,\ref{tab: sparsest_wt}); (2)~The superiority of {\biprune} holds for both unstructured pruning and structured pruning (as shown in Fig.\,\ref{fig: structured_performance_cifar} and Fig.\,\ref{fig: structured_performance_cifar_appendix}). We refer to more experiment results in Appendix\,\ref{app: add_results}.

\underline{Fig.\,\ref{fig: unstructured_performance_cifar}} shows the \textit{unstructured pruning trajectory} (given by test accuracy vs. pruning ratio)
of {\biprune} and baseline methods in $8$ model-dataset setups.  For comparison, we also present the performance of the original dense model. 
As we can see, the proposed   {\biprune} approach finds the best winning tickets  (in terms of the highest accuracy) compared to the baselines across all the pruning setups. 
Among the baseline methods, IMP is the most competitive method to ours. However, the improvement brought by {\biprune} is significant with respect to the variance of IMP, except for the $60\% $-$80\%$ sparsity regime in (CIFAR-10, ResNet-20). In the case of (CIFAR-100, ResNet-20), where IMP can not find any winning tickets (as confirmed by\,\cite{ma2021sanity}),
{\biprune} still manages to find winning tickets with around $0.6\%$ improvement over the dense model.
In \underline{Tab.\,\ref{tab: sparsest_wt}},
we summarize the sparsest  winning tickets along the pruning trajectory identified by different pruning methods. {\biprune} can identify the winning tickets with higher sparsity levels than the other methods, except in the case of (CIFAR-10, ResNet-20). 

\begin{table}[t]
\vspace*{-2mm}
\centering
\caption{\footnotesize{The sparsest winning tickets found by different methods at different data-model setups. Winning tickets refer to the sparse models with an average test accuracy no less than the dense model\,\cite{chen2020lottery}. In each cell, $p\%$ (acc$\pm$std\%) represents the sparsity as well as the test accuracy. The test accuracy of dense models can be found in the header. {\xmark} signifies that no winning ticket is found by a pruning method. Given the data-model setup (\textit{i.e.}, per column),
the sparsest winning ticket is highlighted in \textbf{bold}.
}}
\label{tab: sparsest_wt}
\vspace*{-2mm}
\resizebox{0.9\columnwidth}{!}{%
\begin{tabular}{c|c|c|c|c|c|c}
\toprule[1pt]
\midrule
\multirow{3}{*}{Method} & \multicolumn{4}{c}{CIFAR-10}                               & \multicolumn{2}{|c}{CIFAR-100} \\
                        & ResNet-18                 & ResNet-20                         & ResNet-56                         & VGG-16                        & ResNet-18                 & ResNet-20      \\ 
                        & (94.77\%)                 & (92.12\%)                         & (92.95\%)                         & (93.65\%)                     & (76.67\%)                 &   (68.69\%)      \\\midrule
IMP                     &  87\% (94.77$\pm$0.10\%)  &\textbf{74}\% (92.15$\pm$0.15\%)   & \textbf{74}\% (92.99$\pm$0.12\%)  & 89\% (93.68$\pm$0.05\%)       & 87\% (76.91$\pm$0.19\%)   & \xmark       \\ 
OMP                     & 49\% (94.80$\pm$0.10\%)    & \xmark                            & \xmark                            & 20\% (93.79$\pm$0.06\%)       & 74\% (76.99$\pm$0.07\%)   & \xmark         \\
\grasp                  & \xmark                    & \xmark                            & 36\% (93.07$\pm$0.34\%)           & \xmark                        & \xmark                    & \xmark         \\
{\hydra}                & \xmark                    & \xmark                            & \xmark                            & 87\% (93.73$\pm$0.03\%)       & \xmark                    & 20\% (68.94$\pm$0.17\%)  \\
{\biprune}              & \textbf{89}\% (94.79$\pm$0.15\%)   & 67\% (92.14$\pm$0.15\%)  & \textbf{74}\% (93.13$\pm$0.04\%)  & \textbf{93}\% (93.75$\pm$0.15\%)&  \textbf{89}\% (76.69$\pm$0.18\%)  & \textbf{49}\% (68.78$\pm$0.10\%)\\ \midrule
\bottomrule[1pt]
\end{tabular}%
}
\end{table}

\begin{wrapfigure}{r}{.4\textwidth}
\vspace*{-4mm}
\centerline{
\includegraphics[width=.4\textwidth,height=!]{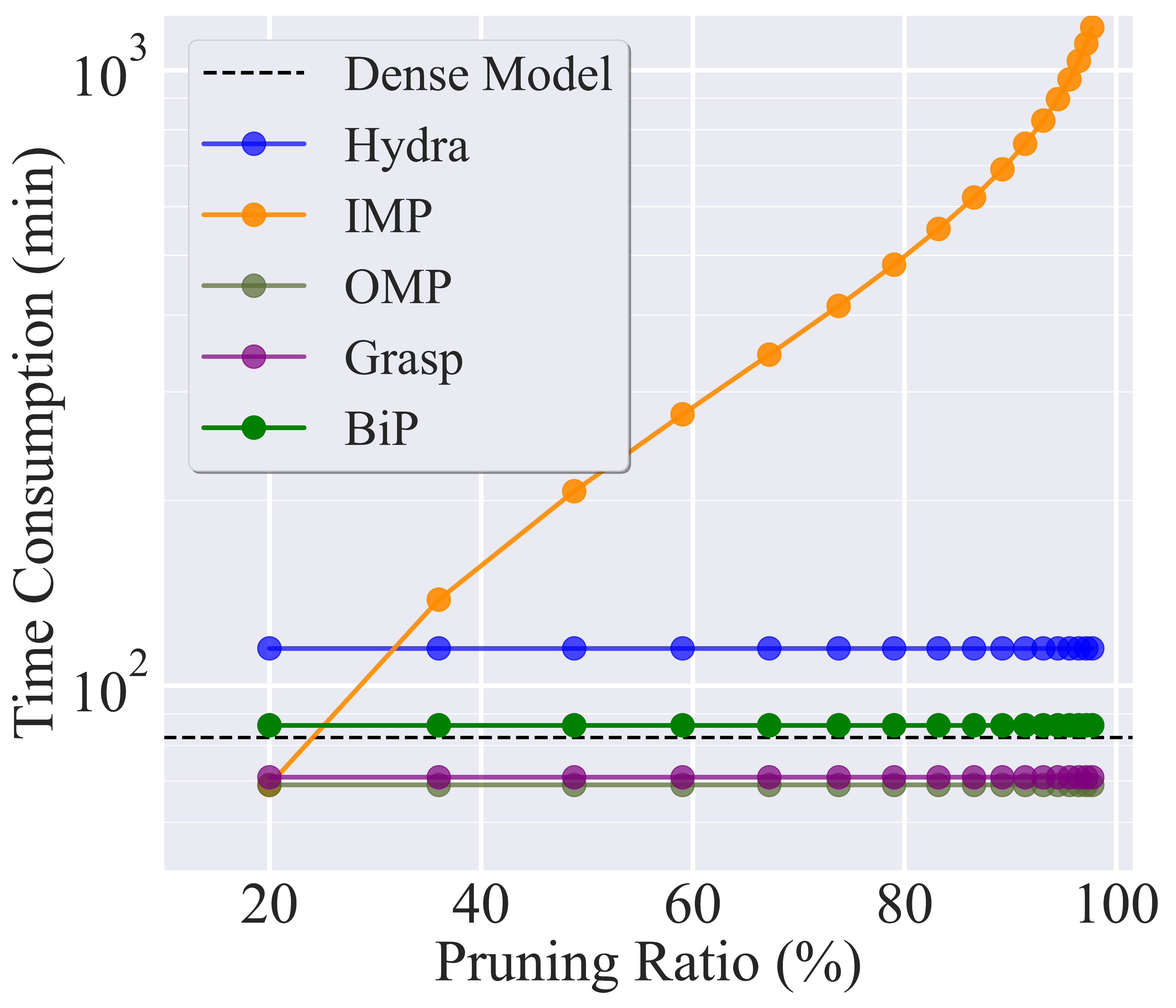}
}
\vspace*{-4mm}
\caption{\footnotesize{Time consumption comparison on (CIFAR-10, ResNet-18) with different pruning ratio $p$. 
}}
\label{fig: time_compare_res18c10}
\vspace*{-5mm}
\end{wrapfigure}

\underline{Fig.\,\ref{fig: structured_performance_cifar}} demonstrates the \textit{structured pruning trajectory}  on the CIFAR-10/100 datasets.
Here we focus on filter pruning, where the filter is regarded as a masking unit in \eqref{eq: prob_bi_level}. We refer readers to Fig.\,\ref{fig: structured_performance_cifar_appendix} for channel-wise pruning results.
Due to the page limit, we only report the results of the filter-wise pruning in the main paper and please refer to Appendix\,\ref{app: add_results} for channel-wise pruning. 
Compared to Fig.\,\ref{fig: unstructured_performance_cifar}, Fig.\,\ref{fig: structured_performance_cifar} shows that it becomes more difficult to find winning tickets of high accuracy and sparsity in the structured pruning, and the gap among different methods decreases. This is not surprising, since
filter pruning imposes much stricter pruning structure constraints than irregular pruning. However, {\biprune} still outperforms all the baselines. Most importantly, it identifies clear winning tickets in the low sparse regime even when IMP fails. 

\paragraph{{\biprune} is computationally efficient.}
In our experiments, another \textbf{key observation} is that 
{\biprune} yields sparsity-agnostic runtime complexity while IMP leads to runtime exponential to the target sparsity.
\underline{Fig.\,\ref{fig: time_compare_res18c10}} shows the computation cost of different methods versus pruning ratios on (CIFAR-10, ResNet-18). 
For example, {\biprune} takes 86 mins to find the sparsest winning ticket (with $89\%$ sparsity in Tab.\,\ref{tab: sparsest_wt}). 
This yields
$7\times$ less runtime than IMP, which consumes $620$ mins to find a comparable winning ticket with $87\%$ sparsity. 
Compared to the optimization-based baseline {\hydra}, {\biprune} is more efficient as it  does not rely on the extra post-optimization retraining; see Tab.\,\ref{tab: efficiency} for a detailed summary of runtime and number of training epochs required by different pruning methods. 
Further, {\biprune} takes about {1.25 $\times$} more computation time than {\grasp} and OMP. However, the latter methods lead to worse pruned model accuracy, as demonstrated by their failure to find
winning tickets  in  Tab.\,\ref{tab: sparsest_wt}, Fig.\,\ref{fig: unstructured_performance_cifar}, and Fig.\,\ref{fig: structured_performance_cifar}.

\begin{figure}[t]
\centerline{
\begin{tabular}{ccc}
    \includegraphics[width=.32\textwidth,height=!]{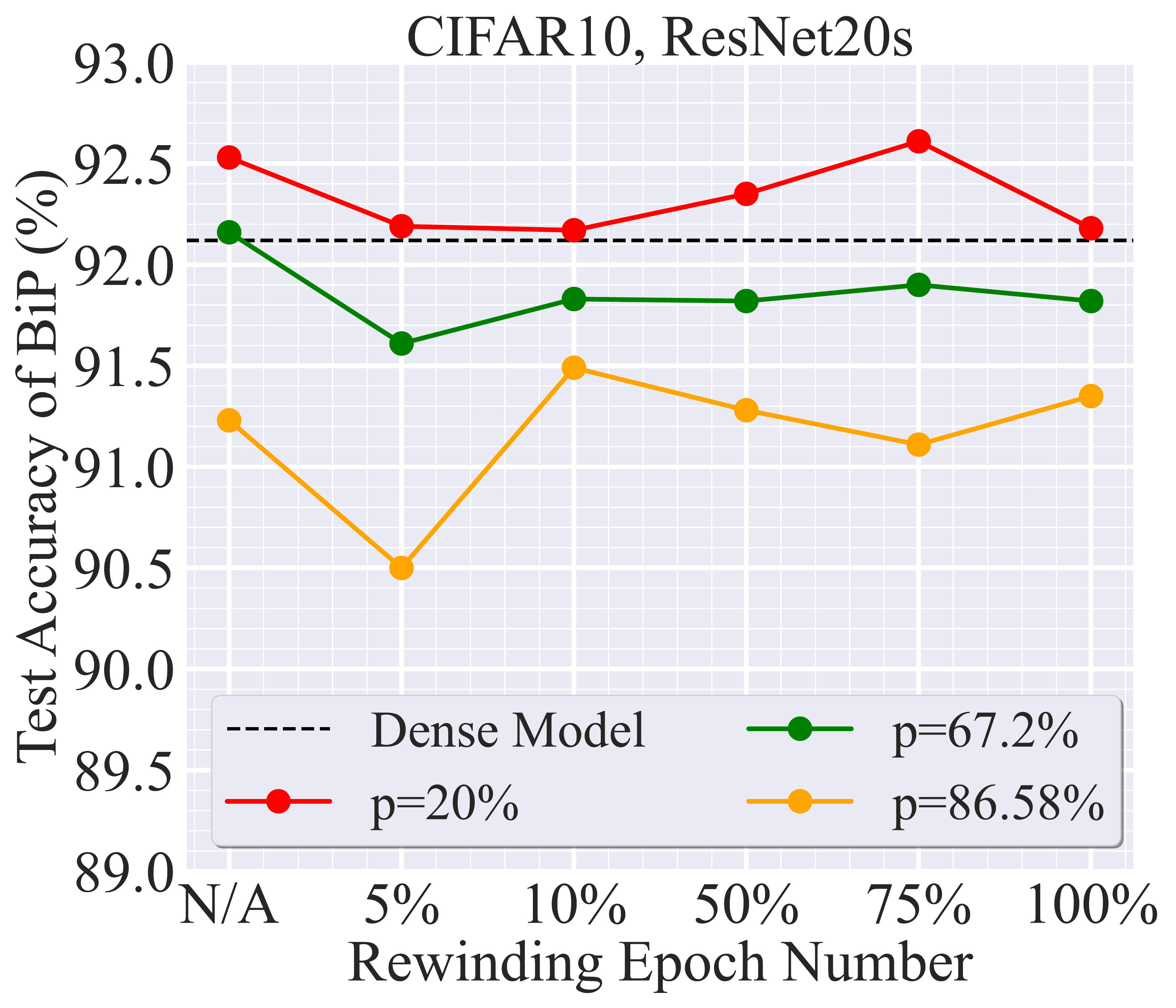} &
    \includegraphics[width=.32\textwidth,height=!]{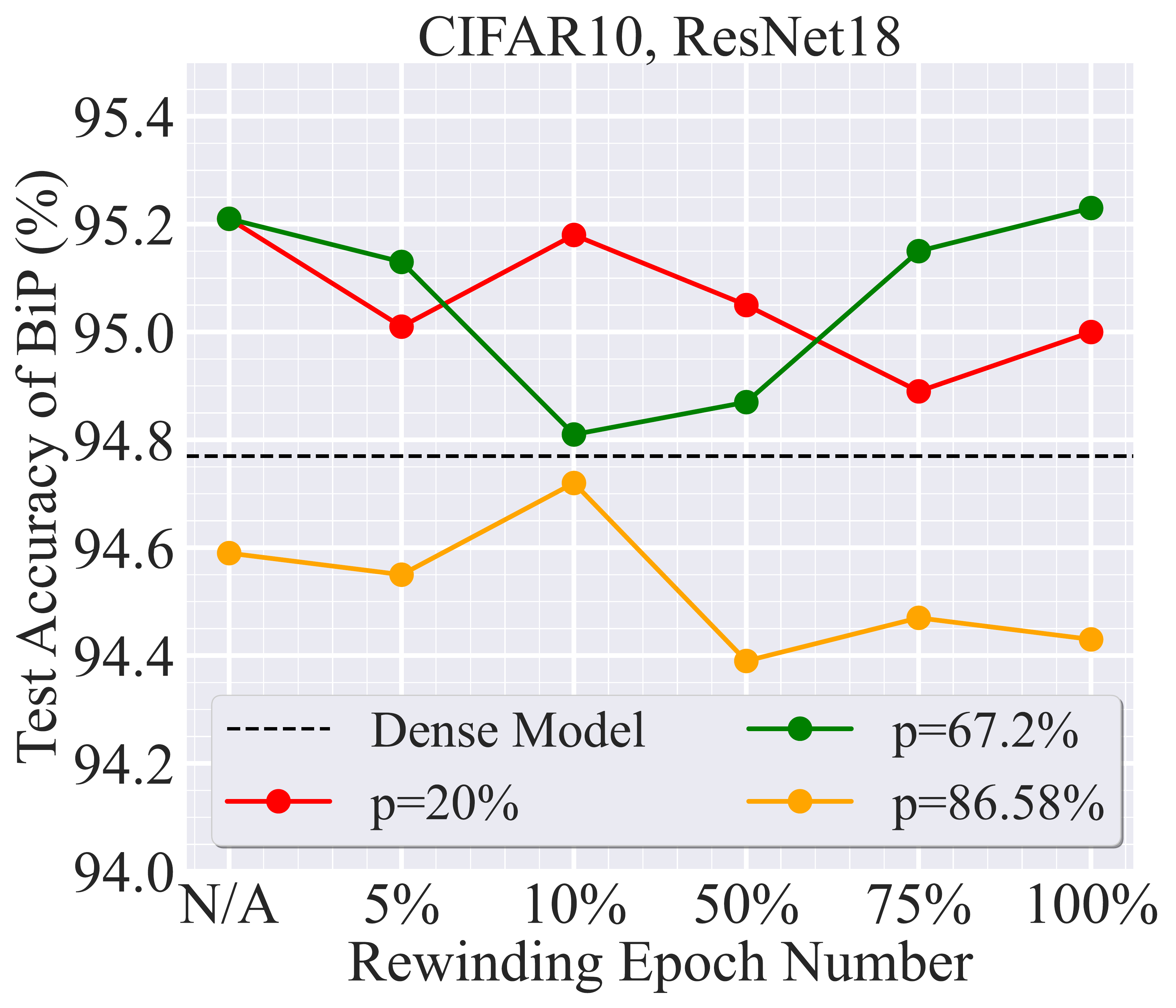} \vspace*{-0.5mm}&
    \includegraphics[width=.32\textwidth,height=!]{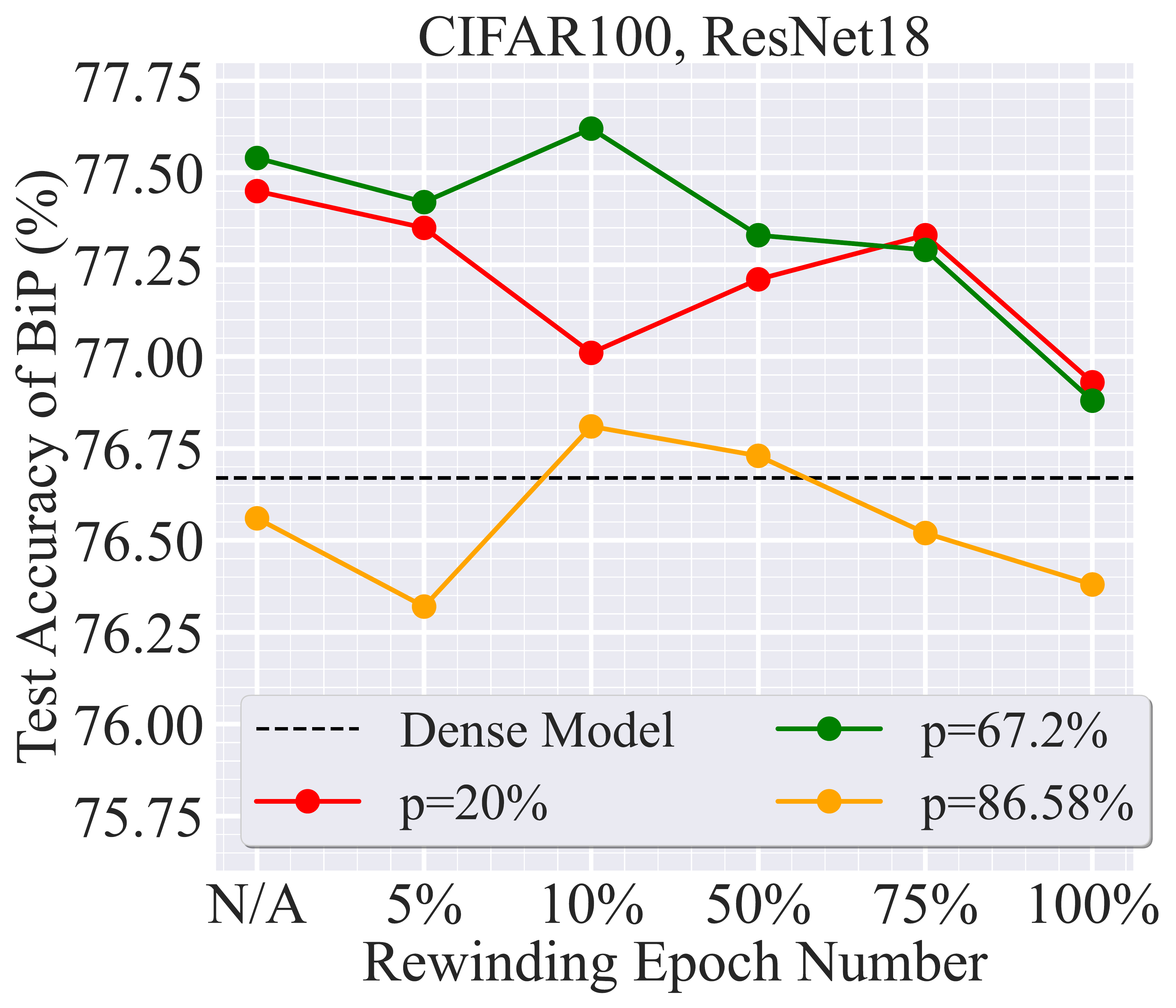} 
\end{tabular}}
\caption{\footnotesize{{The sensitivity of {\biprune} to rewinding epoch numbers on different datasets and model architectures. "N/A" in the x-axis indicates {\biprune} without retraining.} 
}}
\label{fig: rewind_effect}
\end{figure}

\paragraph{\CR{{\biprune} requires no rewinding.}}
Another advantage of {\biprune} is that it insensitive to   model rewinding to find matching subnetworks. 
Recall that rewinding is a strategy used in LTH \cite{frankle2020linear} to determine what model initialization should be used for retraining a pruned   model. As shown in \cite{ma2021sanity}, an IMP-identified winning ticket could be sensitive to the choice of a rewinding point.
\underline{Fig.\,\ref{fig: rewind_effect}} shows the test accuracy of the {\biprune}-pruned model when it is retrained at different rewinding epochs under various datasets and model architectures, where `N/A' in the x-axis represents the case of no retraining (and thus no rewinding). 
As we can see, a carefully-tuned rewinding scheme does not lead to a significant improvement over {\biprune} without retraining. This suggests that the subnetworks found by {\biprune} are already of high quality and does not require any rewinding operation. 

\paragraph{{Additional results.}}
We include more experiment results in Appendix\,\ref{app: add_results}. In particular, we show more results in both unstructured and structured pruning settings in Fig.\,\ref{fig: compare_init}, Fig.\,\ref{fig: unstructured_rebuttal}, Fig.\,\ref{fig: structured_rebuttal} and Fig.\,\ref{fig: resnet18_imagenet_rebuttal}, where we compare {\biprune} with more baselines and cover more model architectures. We also study the sensitivity of {\biprune} to the lower-level step number, lower-level regularization coefficient, the significance of the implicit gradient term  \eqref{eq: GD_bi_level}, learning rate, and batch size, as shown in Fig.\,\ref{fig: ablation_study} and Fig.\,\ref{fig: ablation_study_full}.
To demonstrate the convergence of the upper-level and lower-level optimization in {\biprune}, we show the training trajectory of {\biprune} for  accuracy (Fig.\ref{fig: convergence}) and mask score (Fig.\ref{fig: IoU_Trajectory_rebuttal}), and show how the lower-level step number affects the convergence speed (Fig.\,\ref{fig: step_convergence_rebuttal})).
Further,  we show the performance of {\biprune} vs. the growth of training epochs (Fig.\,\ref{fig: epoch_extreme_sparsity_rebuttal}), and its performance vs. different data batch schedulers (see Fig.\,\ref{fig: batch_diverse_rebuttal}).

\section{Conclusion}
We proposed the {\biprune}  method to find sparse networks through the lens of BLO.  
Our work advanced the algorithmic foundation of model pruning by characterizing its pruning-retraining hierarchy using   BLO.
We theoretically showed that {\biprune} can be solved as easily as first-order optimization by exploiting the bi-linearity of pruning variables. We also empirically showed that {\biprune} can  find high-quality winning tickets very efficiently compared to the predominant iterative pruning method.
In the future, we will \RV{seek the optimal curriculum of training data at different optimization levels of {\biprune}, and will investigate the performance of our proposal for   actual hardware acceleration.}

\vspace{-1mm}
\section*{\CR{Acknowledgement}}

The work of Y. Zhang, Y. Yao, and S. Liu was partially supported by National Science Foundation (NSF) Grant IIS-2207052 and Cisco Research Award. The work of M. Hong was supported by NSF  grants CIF-1910385 and CMMI-1727757. The work of Y. Wang was supported NSF grant CCF-1919117.
The computing resources used in this work were also supported by the MIT-IBM Watson AI Lab, IBM Research and the Institute for Cyber-Enabled Research (ICER) at Michigan State University.

\newpage
{{
\bibliographystyle{IEEEbib}
\bibliography{reference,refs_report, refs_prune, refs, ref_related}

\begin{thebibliography}{100}

\bibitem{krizhevsky2012imagenet}
Alex Krizhevsky, Ilya Sutskever, and Geoffrey~E Hinton,
\newblock ``Imagenet classification with deep convolutional neural networks,''
\newblock in {\em Advances in neural information processing systems}, 2012, pp.
  1097--1105.

\bibitem{brown2020language}
Tom Brown, Benjamin Mann, Nick Ryder, Melanie Subbiah, Jared~D Kaplan, Prafulla
  Dhariwal, Arvind Neelakantan, Pranav Shyam, Girish Sastry, Amanda Askell,
  et~al.,
\newblock ``Language models are few-shot learners,''
\newblock {\em Advances in neural information processing systems}, vol. 33, pp.
  1877--1901, 2020.

\bibitem{bartlett2021deep}
Peter~L Bartlett, Andrea Montanari, and Alexander Rakhlin,
\newblock ``Deep learning: a statistical viewpoint,''
\newblock {\em Acta numerica}, vol. 30, pp. 87--201, 2021.

\bibitem{han2015deep}
Song Han, Huizi Mao, and William~J Dally,
\newblock ``Deep compression: Compressing deep neural networks with pruning,
  trained quantization and huffman coding,''
\newblock {\em arXiv preprint arXiv:1510.00149}, 2015.

\bibitem{blalock2020state}
Davis Blalock, Jose Javier~Gonzalez Ortiz, Jonathan Frankle, and John Guttag,
\newblock ``What is the state of neural network pruning?,''
\newblock {\em arXiv preprint arXiv:2003.03033}, 2020.

\bibitem{han2015learning}
Song Han, Jeff Pool, John Tran, and William Dally,
\newblock ``Learning both weights and connections for efficient neural
  network,''
\newblock {\em Advances in neural information processing systems}, vol. 28,
  2015.

\bibitem{he2018amc}
Yihui He, Ji~Lin, Zhijian Liu, Hanrui Wang, Li-Jia Li, and Song Han,
\newblock ``Amc: Automl for model compression and acceleration on mobile
  devices,''
\newblock in {\em Proceedings of the European conference on computer vision
  (ECCV)}, 2018, pp. 784--800.

\bibitem{mao2017exploring}
Huizi Mao, Song Han, Jeff Pool, Wenshuo Li, Xingyu Liu, Yu~Wang, and William~J
  Dally,
\newblock ``Exploring the granularity of sparsity in convolutional neural
  networks,''
\newblock in {\em Proceedings of the IEEE Conference on Computer Vision and
  Pattern Recognition Workshops}, 2017, pp. 13--20.

\bibitem{sehwag2020hydra}
Vikash Sehwag, Shiqi Wang, Prateek Mittal, and Suman Jana,
\newblock ``Hydra: Pruning adversarially robust neural networks,''
\newblock {\em Advances in Neural Information Processing Systems}, vol. 33, pp.
  19655--19666, 2020.

\bibitem{diffenderfer2021winning}
James Diffenderfer, Brian Bartoldson, Shreya Chaganti, Jize Zhang, and Bhavya
  Kailkhura,
\newblock ``A winning hand: Compressing deep networks can improve
  out-of-distribution robustness,''
\newblock {\em Advances in Neural Information Processing Systems}, vol. 34, pp.
  664--676, 2021.

\bibitem{chen2021lottery}
Tianlong Chen, Jonathan Frankle, Shiyu Chang, Sijia Liu, Yang Zhang, Michael
  Carbin, and Zhangyang Wang,
\newblock ``The lottery tickets hypothesis for supervised and self-supervised
  pre-training in computer vision models,''
\newblock in {\em Proceedings of the IEEE/CVF Conference on Computer Vision and
  Pattern Recognition}, 2021, pp. 16306--16316.

\bibitem{ma2020pconv}
Xiaolong Ma, Fu-Ming Guo, Wei Niu, Xue Lin, Jian Tang, Kaisheng Ma, Bin Ren,
  and Yanzhi Wang,
\newblock ``Pconv: The missing but desirable sparsity in dnn weight pruning for
  real-time execution on mobile devices,''
\newblock in {\em Proceedings of the AAAI Conference on Artificial
  Intelligence}, 2020, vol.~34, pp. 5117--5124.

\bibitem{chen2022coarsening}
Tianlong Chen, Xuxi Chen, Xiaolong Ma, Yanzhi Wang, and Zhangyang Wang,
\newblock ``Coarsening the granularity: Towards structurally sparse lottery
  tickets,''
\newblock {\em arXiv preprint arXiv:2202.04736}, 2022.

\bibitem{lecun1989optimal}
Yann LeCun, John Denker, and Sara Solla,
\newblock ``Optimal brain damage,''
\newblock {\em Advances in neural information processing systems}, vol. 2,
  1989.

\bibitem{ren2018admmnn}
Ao~Ren, Tianyun Zhang, Shaokai Ye, Jiayu Li, Wenyao Xu, Xuehai Qian, Xue Lin,
  and Yanzhi Wang,
\newblock ``Admm-nn: An algorithm-hardware co-design framework of dnns using
  alternating direction method of multipliers,'' 2018.

\bibitem{ramanujan2020sHiddenSubnetwork}
Vivek Ramanujan, Mitchell Wortsman, Aniruddha Kembhavi, Ali Farhadi, and
  Mohammad Rastegari,
\newblock ``What's hidden in a randomly weighted neural network?,''
\newblock in {\em Proceedings of the IEEE/CVF Conference on Computer Vision and
  Pattern Recognition}, 2020, pp. 11893--11902.

\bibitem{frankle2018lottery}
Jonathan Frankle and Michael Carbin,
\newblock ``The lottery ticket hypothesis: Finding sparse, trainable neural
  networks,''
\newblock {\em arXiv preprint arXiv:1803.03635}, 2018.

\bibitem{Renda2020Comparing}
Alex Renda, Jonathan Frankle, and Michael Carbin,
\newblock ``Comparing rewinding and fine-tuning in neural network pruning,''
\newblock in {\em 8th International Conference on Learning Representations},
  2020.

\bibitem{frankle2020linear}
Jonathan Frankle, Gintare~Karolina Dziugaite, Daniel Roy, and Michael Carbin,
\newblock ``Linear mode connectivity and the lottery ticket hypothesis,''
\newblock in {\em International Conference on Machine Learning}. PMLR, 2020,
  pp. 3259--3269.

\bibitem{chen2020lottery}
Tianlong Chen, Jonathan Frankle, Shiyu Chang, Sijia Liu, Yang Zhang, Zhangyang
  Wang, and Michael Carbin,
\newblock ``The lottery ticket hypothesis for pre-trained bert networks,''
\newblock {\em Advances in neural information processing systems}, vol. 33, pp.
  15834--15846, 2020.

\bibitem{lee2018snip}
Namhoon Lee, Thalaiyasingam Ajanthan, and Philip~HS Torr,
\newblock ``Snip: Single-shot network pruning based on connection
  sensitivity,''
\newblock {\em arXiv preprint arXiv:1810.02340}, 2018.

\bibitem{ma2021sanity}
Xiaolong Ma, Geng Yuan, Xuan Shen, Tianlong Chen, Xuxi Chen, Xiaohan Chen, Ning
  Liu, Minghai Qin, Sijia Liu, Zhangyang Wang, et~al.,
\newblock ``Sanity checks for lottery tickets: Does your winning ticket really
  win the jackpot?,''
\newblock {\em arXiv preprint arXiv:2107.00166}, 2021.

\bibitem{wang2020pick}
Chaoqi Wang, Guodong Zhang, and Roger Grosse,
\newblock ``Picking winning tickets before training by preserving gradient
  flow,''
\newblock {\em arXiv preprint arXiv:2002.07376}, 2020.

\bibitem{tanaka2020pruning}
Hidenori Tanaka, Daniel Kunin, Daniel~L Yamins, and Surya Ganguli,
\newblock ``Pruning neural networks without any data by iteratively conserving
  synaptic flow,''
\newblock {\em Advances in Neural Information Processing Systems}, vol. 33, pp.
  6377--6389, 2020.

\bibitem{alizadeh2022prospect}
Milad Alizadeh, Shyam~A. Tailor, Luisa~M Zintgraf, Joost van Amersfoort,
  Sebastian Farquhar, Nicholas~Donald Lane, and Yarin Gal,
\newblock ``Prospect pruning: Finding trainable weights at initialization using
  meta-gradients,''
\newblock in {\em International Conference on Learning Representations}, 2022.

\bibitem{lee2020layer}
Jaeho Lee, Sejun Park, Sangwoo Mo, Sungsoo Ahn, and Jinwoo Shin,
\newblock ``Layer-adaptive sparsity for the magnitude-based pruning,''
\newblock {\em arXiv preprint arXiv:2010.07611}, 2020.

\bibitem{csordas2020neural}
R{\'o}bert Csord{\'a}s, Sjoerd van Steenkiste, and J{\"u}rgen Schmidhuber,
\newblock ``Are neural nets modular? inspecting functional modularity through
  differentiable weight masks,''
\newblock {\em arXiv preprint arXiv:2010.02066}, 2020.

\bibitem{you2019drawing}
Haoran You, Chaojian Li, Pengfei Xu, Yonggan Fu, Yue Wang, Xiaohan Chen,
  Richard~G Baraniuk, Zhangyang Wang, and Yingyan Lin,
\newblock ``Drawing early-bird tickets: Towards more efficient training of deep
  networks,''
\newblock {\em arXiv preprint arXiv:1909.11957}, 2019.

\bibitem{zhang2021efficient}
Zhenyu Zhang, Xuxi Chen, Tianlong Chen, and Zhangyang Wang,
\newblock ``Efficient lottery ticket finding: Less data is more,''
\newblock in {\em International Conference on Machine Learning}. PMLR, 2021,
  pp. 12380--12390.

\bibitem{su2020sanity_random_win}
Jingtong Su, Yihang Chen, Tianle Cai, Tianhao Wu, Ruiqi Gao, Liwei Wang, and
  Jason~D Lee,
\newblock ``Sanity-checking pruning methods: Random tickets can win the
  jackpot,''
\newblock {\em Advances in Neural Information Processing Systems}, vol. 33, pp.
  20390--20401, 2020.

\bibitem{frankle2020pruning}
Jonathan Frankle, Gintare~Karolina Dziugaite, Daniel~M Roy, and Michael Carbin,
\newblock ``Pruning neural networks at initialization: Why are we missing the
  mark?,''
\newblock {\em arXiv preprint arXiv:2009.08576}, 2020.

\bibitem{liu2019admm}
Sijia Liu, Parikshit Ram, Deepak Vijaykeerthy, Djallel Bouneffouf, Gregory
  Bramble, Horst Samulowitz, Dakuo Wang, Andrew Conn, and Alexander Gray,
\newblock ``An {ADMM} based framework for automl pipeline configuration,''
  2019.

\bibitem{wang2020neural}
Huan Wang, Can Qin, Yulun Zhang, and Yun Fu,
\newblock ``Neural pruning via growing regularization,''
\newblock {\em arXiv preprint arXiv:2012.09243}, 2020.

\bibitem{denil2013overparamterized_and_redundancy}
Misha Denil, Babak Shakibi, Laurent Dinh, Marc'Aurelio Ranzato, and Nando
  De~Freitas,
\newblock ``Predicting parameters in deep learning,''
\newblock {\em Advances in neural information processing systems}, vol. 26,
  2013.

\bibitem{liu2020primer}
Sijia Liu, Pin-Yu Chen, Bhavya Kailkhura, Gaoyuan Zhang, Alfred~O Hero~III, and
  Pramod~K Varshney,
\newblock ``A primer on zeroth-order optimization in signal processing and
  machine learning: Principals, recent advances, and applications,''
\newblock {\em IEEE Signal Processing Magazine}, vol. 37, no. 5, pp. 43--54,
  2020.

\bibitem{zhu2017prune}
Michael Zhu and Suyog Gupta,
\newblock ``To prune, or not to prune: exploring the efficacy of pruning for
  model compression,''
\newblock {\em arXiv preprint arXiv:1710.01878}, 2017.

\bibitem{janowsky1989pruning}
Steven~A Janowsky,
\newblock ``Pruning versus clipping in neural networks,''
\newblock {\em Physical Review A}, vol. 39, no. 12, pp. 6600, 1989.

\bibitem{peste2021ac}
Alexandra Peste, Eugenia Iofinova, Adrian Vladu, and Dan Alistarh,
\newblock ``Ac/dc: Alternating compressed/decompressed training of deep neural
  networks,''
\newblock {\em Advances in Neural Information Processing Systems}, vol. 34, pp.
  8557--8570, 2021.

\bibitem{mozer1989skeletonization}
Michael~C Mozer and Paul Smolensky,
\newblock ``Skeletonization: A technique for trimming the fat from a network
  via relevance assessment,''
\newblock in {\em Advances in neural information processing systems}, 1989, pp.
  107--115.

\bibitem{evci2020rigging}
Utku Evci, Trevor Gale, Jacob Menick, Pablo~Samuel Castro, and Erich Elsen,
\newblock ``Rigging the lottery: Making all tickets winners,''
\newblock in {\em International Conference on Machine Learning}. PMLR, 2020,
  pp. 2943--2952.

\bibitem{molchanov2019importance}
Pavlo Molchanov, Arun Mallya, Stephen Tyree, Iuri Frosio, and Jan Kautz,
\newblock ``Importance estimation for neural network pruning,''
\newblock in {\em Proceedings of the IEEE Conference on Computer Vision and
  Pattern Recognition}, 2019, pp. 11264--11272.

\bibitem{yao2020pyhessian}
Zhewei Yao, Amir Gholami, Kurt Keutzer, and Michael~W Mahoney,
\newblock ``Pyhessian: Neural networks through the lens of the hessian,''
\newblock in {\em 2020 IEEE International Conference on Big Data (Big Data)}.
  IEEE, 2020, pp. 581--590.

\bibitem{lecun1990optimal}
Yann LeCun, John~S Denker, and Sara~A Solla,
\newblock ``Optimal brain damage,''
\newblock in {\em Advances in neural information processing systems}, 1990, pp.
  598--605.

\bibitem{hassibi1993second}
Babak Hassibi and David~G Stork,
\newblock {\em Second order derivatives for network pruning: Optimal brain
  surgeon},
\newblock Morgan Kaufmann, 1993.

\bibitem{molchanov2016pruning}
Pavlo Molchanov, Stephen Tyree, Tero Karras, Timo Aila, and Jan Kautz,
\newblock ``Pruning convolutional neural networks for resource efficient
  inference,''
\newblock {\em arXiv preprint arXiv:1611.06440}, 2016.

\bibitem{singh2020woodfisher}
Sidak~Pal Singh and Dan Alistarh,
\newblock ``Woodfisher: Efficient second-order approximation for neural network
  compression,''
\newblock {\em Advances in Neural Information Processing Systems}, vol. 33, pp.
  18098--18109, 2020.

\bibitem{liu2017learning}
Zhuang Liu, Jianguo Li, Zhiqiang Shen, Gao Huang, Shoumeng Yan, and Changshui
  Zhang,
\newblock ``Learning efficient convolutional networks through network
  slimming,''
\newblock in {\em Proceedings of the IEEE International Conference on Computer
  Vision}, 2017, pp. 2736--2744.

\bibitem{he2017channel}
Yihui He, Xiangyu Zhang, and Jian Sun,
\newblock ``Channel pruning for accelerating very deep neural networks,''
\newblock in {\em Proceedings of the IEEE international conference on computer
  vision}, 2017, pp. 1389--1397.

\bibitem{zhou2016less}
Hao Zhou, Jose~M Alvarez, and Fatih Porikli,
\newblock ``Less is more: Towards compact cnns,''
\newblock in {\em European Conference on Computer Vision}. Springer, 2016, pp.
  662--677.

\bibitem{louizos2017learning}
Christos Louizos, Max Welling, and Diederik~P Kingma,
\newblock ``Learning sparse neural networks through $ l\_0 $ regularization,''
\newblock {\em arXiv preprint arXiv:1712.01312}, 2017.

\bibitem{guo2021gdp}
Yi~Guo, Huan Yuan, Jianchao Tan, Zhangyang Wang, Sen Yang, and Ji~Liu,
\newblock ``Gdp: Stabilized neural network pruning via gates with
  differentiable polarization,''
\newblock in {\em Proceedings of the IEEE/CVF International Conference on
  Computer Vision}, 2021, pp. 5239--5250.

\bibitem{liberis2021differentiable}
Edgar Liberis and Nicholas~D Lane,
\newblock ``Differentiable network pruning for microcontrollers,''
\newblock {\em arXiv preprint arXiv:2110.08350}, 2021.

\bibitem{kusupati2020soft}
Aditya Kusupati, Vivek Ramanujan, Raghav Somani, Mitchell Wortsman, Prateek
  Jain, Sham Kakade, and Ali Farhadi,
\newblock ``Soft threshold weight reparameterization for learnable sparsity,''
\newblock in {\em International Conference on Machine Learning}. PMLR, 2020,
  pp. 5544--5555.

\bibitem{xue2021rethinking}
Chao Xue, Xiaoxing Wang, Junchi Yan, Yonggang Hu, Xiaokang Yang, and Kewei Sun,
\newblock ``Rethinking bi-level optimization in neural architecture search: A
  gibbs sampling perspective,''
\newblock in {\em Proceedings of the AAAI Conference on Artificial
  Intelligence}, 2021, vol.~35, pp. 10551--10559.

\bibitem{zhou2021effective}
Xiao Zhou, Weizhong Zhang, Hang Xu, and Tong Zhang,
\newblock ``Effective sparsification of neural networks with global sparsity
  constraint,''
\newblock in {\em Proceedings of the IEEE/CVF Conference on Computer Vision and
  Pattern Recognition}, 2021, pp. 3599--3608.

\bibitem{gale2019state}
Trevor Gale, Erich Elsen, and Sara Hooker,
\newblock ``The state of sparsity in deep neural networks,''
\newblock {\em arXiv}, vol. abs/1902.09574, 2019.

\bibitem{pmlr-v139-zhang21c}
Zhenyu Zhang, Xuxi Chen, Tianlong Chen, and Zhangyang Wang,
\newblock ``Efficient lottery ticket finding: Less data is more,''
\newblock in {\em Proceedings of the 38th International Conference on Machine
  Learning}, Marina Meila and Tong Zhang, Eds. 18--24 Jul 2021, vol. 139 of
  {\em Proceedings of Machine Learning Research}, pp. 12380--12390, PMLR.

\bibitem{chen2020lottery2}
Tianlong Chen, Jonathan Frankle, Shiyu Chang, Sijia Liu, Yang Zhang, Michael
  Carbin, and Zhangyang Wang,
\newblock ``The lottery tickets hypothesis for supervised and self-supervised
  pre-training in computer vision models,''
\newblock {\em arXiv preprint arXiv:2012.06908}, 2020.

\bibitem{yu2019playing}
Haonan Yu, Sergey Edunov, Yuandong Tian, and Ari~S. Morcos,
\newblock ``Playing the lottery with rewards and multiple languages: lottery
  tickets in rl and nlp,''
\newblock in {\em 8th International Conference on Learning Representations},
  2020.

\bibitem{chen2021gans}
Xuxi Chen, Zhenyu Zhang, Yongduo Sui, and Tianlong Chen,
\newblock ``{\{}GAN{\}}s can play lottery tickets too,''
\newblock in {\em International Conference on Learning Representations}, 2021.

\bibitem{ma2021good}
Haoyu Ma, Tianlong Chen, Ting-Kuei Hu, Chenyu You, Xiaohui Xie, and Zhangyang
  Wang,
\newblock ``Good students play big lottery better,''
\newblock {\em arXiv preprint arXiv:2101.03255}, 2021.

\bibitem{gan2021playing}
Zhe Gan, Yen-Chun Chen, Linjie Li, Tianlong Chen, Yu~Cheng, Shuohang Wang, and
  Jingjing Liu,
\newblock ``Playing lottery tickets with vision and language,''
\newblock {\em arXiv preprint arXiv:2104.11832}, 2021.

\bibitem{chen2021unified}
Tianlong Chen, Yongduo Sui, Xuxi Chen, Aston Zhang, and Zhangyang Wang,
\newblock ``A unified lottery ticket hypothesis for graph neural networks,''
\newblock {\em arXiv preprint arXiv:2102.06790}, 2021.

\bibitem{kalibhat2021winning}
Neha~Mukund Kalibhat, Yogesh Balaji, and Soheil Feizi,
\newblock ``Winning lottery tickets in deep generative models,'' 2021.

\bibitem{chen2021ultra}
Tianlong Chen, Yu~Cheng, Zhe Gan, Jingjing Liu, and Zhangyang Wang,
\newblock ``Ultra-data-efficient gan training: Drawing a lottery ticket first,
  then training it toughly,''
\newblock {\em arXiv preprint arXiv:2103.00397}, 2021.

\bibitem{chen2020long}
Tianlong Chen, Zhenyu Zhang, Sijia Liu, Shiyu Chang, and Zhangyang Wang,
\newblock ``Long live the lottery: The existence of winning tickets in lifelong
  learning,''
\newblock in {\em International Conference on Learning Representations}, 2020.

\bibitem{li2016pruning}
Hao Li, Asim Kadav, Igor Durdanovic, Hanan Samet, and Hans~Peter Graf,
\newblock ``Pruning filters for efficient convnets,''
\newblock {\em arXiv preprint arXiv:1608.08710}, 2016.

\bibitem{niu2020patdnn}
Wei Niu, Xiaolong Ma, Sheng Lin, Shihao Wang, Xuehai Qian, Xue Lin, Yanzhi
  Wang, and Bin Ren,
\newblock ``Patdnn: Achieving real-time dnn execution on mobile devices with
  pattern-based weight pruning,''
\newblock in {\em Proceedings of the Twenty-Fifth International Conference on
  Architectural Support for Programming Languages and Operating Systems}, 2020,
  pp. 907--922.

\bibitem{ma2020image}
Xiaolong Ma, Wei Niu, Tianyun Zhang, Sijia Liu, Sheng Lin, Hongjia Li, Wujie
  Wen, Xiang Chen, Jian Tang, Kaisheng Ma, et~al.,
\newblock ``An image enhancing pattern-based sparsity for real-time inference
  on mobile devices,''
\newblock in {\em European Conference on Computer Vision}. Springer, 2020, pp.
  629--645.

\bibitem{wang2022paca}
Jingyu Wang, Songming Yu, Zhuqing Yuan, Jinshan Yue, Zhe Yuan, Ruoyang Liu,
  Yanzhi Wang, Huazhong Yang, Xueqing Li, and Yongpan Liu,
\newblock ``Paca: A pattern pruning algorithm and channel-fused high pe
  utilization accelerator for cnns,''
\newblock {\em IEEE Transactions on Computer-Aided Design of Integrated
  Circuits and Systems}, 2022.

\bibitem{van2020single}
Joost van Amersfoort, Milad Alizadeh, Sebastian Farquhar, Nicholas Lane, and
  Yarin Gal,
\newblock ``Single shot structured pruning before training,''
\newblock {\em arXiv preprint arXiv:2007.00389}, 2020.

\bibitem{falk1995bilevel}
James~E Falk and Jiming Liu,
\newblock ``On bilevel programming, part i: general nonlinear cases,''
\newblock {\em Mathematical Programming}, vol. 70, no. 1, pp. 47--72, 1995.

\bibitem{vicente1994descent}
Luis Vicente, Gilles Savard, and Joaquim J{\'u}dice,
\newblock ``Descent approaches for quadratic bilevel programming,''
\newblock {\em Journal of Optimization Theory and Applications}, vol. 81, no.
  2, pp. 379--399, 1994.

\bibitem{chen2022gradient}
Can Chen, Xi~Chen, Chen Ma, Zixuan Liu, and Xue Liu,
\newblock ``Gradient-based bi-level optimization for deep learning: A survey,''
\newblock {\em arXiv preprint arXiv:2207.11719}, 2022.

\bibitem{white1993penalty}
Douglas~J White and G~Anandalingam,
\newblock ``A penalty function approach for solving bi-level linear programs,''
\newblock {\em Journal of Global Optimization}, vol. 3, no. 4, pp. 397--419,
  1993.

\bibitem{gould2016differentiating}
Stephen Gould, Basura Fernando, Anoop Cherian, Peter Anderson, Rodrigo~Santa
  Cruz, and Edison Guo,
\newblock ``On differentiating parameterized argmin and argmax problems with
  application to bi-level optimization,''
\newblock {\em arXiv preprint arXiv:1607.05447}, 2016.

\bibitem{sabach2017first}
Shoham Sabach and Shimrit Shtern,
\newblock ``A first order method for solving convex bilevel optimization
  problems,''
\newblock {\em SIAM Journal on Optimization}, vol. 27, no. 2, pp. 640--660,
  2017.

\bibitem{liu2020generic}
Risheng Liu, Pan Mu, Xiaoming Yuan, Shangzhi Zeng, and Jin Zhang,
\newblock ``A generic first-order algorithmic framework for bi-level
  programming beyond lower-level singleton,''
\newblock in {\em International Conference on Machine Learning}. PMLR, 2020,
  pp. 6305--6315.

\bibitem{li2020improved}
Junyi Li, Bin Gu, and Heng Huang,
\newblock ``Improved bilevel model: Fast and optimal algorithm with theoretical
  guarantee,''
\newblock {\em arXiv preprint arXiv:2009.00690}, 2020.

\bibitem{ghadimi2018approximation}
Saeed Ghadimi and Mengdi Wang,
\newblock ``Approximation methods for bilevel programming,''
\newblock {\em arXiv preprint arXiv:1802.02246}, 2018.

\bibitem{ji2020bilevel}
Kaiyi Ji, Junjie Yang, and Yingbin Liang,
\newblock ``Bilevel optimization: Nonasymptotic analysis and faster
  algorithms,''
\newblock {\em arXiv preprint arXiv:2010.07962}, 2020.

\bibitem{hong2020two}
Mingyi Hong, Hoi-To Wai, Zhaoran Wang, and Zhuoran Yang,
\newblock ``A two-timescale framework for bilevel optimization: Complexity
  analysis and application to actor-critic,''
\newblock {\em arXiv preprint arXiv:2007.05170}, 2020.

\bibitem{franceschi2017forward}
Luca Franceschi, Michele Donini, Paolo Frasconi, and Massimiliano Pontil,
\newblock ``Forward and reverse gradient-based hyperparameter optimization,''
\newblock in {\em International Conference on Machine Learning}. PMLR, 2017,
  pp. 1165--1173.

\bibitem{grazzi2020iteration}
Riccardo Grazzi, Luca Franceschi, Massimiliano Pontil, and Saverio Salzo,
\newblock ``On the iteration complexity of hypergradient computation,''
\newblock in {\em International Conference on Machine Learning}. PMLR, 2020,
  pp. 3748--3758.

\bibitem{shaban2019truncated}
Amirreza Shaban, Ching-An Cheng, Nathan Hatch, and Byron Boots,
\newblock ``Truncated back-propagation for bilevel optimization,''
\newblock in {\em The 22nd International Conference on Artificial Intelligence
  and Statistics}. PMLR, 2019, pp. 1723--1732.

\bibitem{zhang2021bat}
Yihua Zhang, Guanhuan Zhang, Prashant Khanduri, Mingyi Hong, Shiyu Chang, and
  Sijia Liu,
\newblock ``Revisiting and advancing fast adversarial training through the lens
  of bi-level optimization,''
\newblock {\em arXiv preprint arXiv:2112.12376}, 2021.

\bibitem{rajeswaran2019meta}
Aravind Rajeswaran, Chelsea Finn, Sham~M Kakade, and Sergey Levine,
\newblock ``Meta-learning with implicit gradients,''
\newblock in {\em Advances in Neural Information Processing Systems}, 2019, pp.
  113--124.

\bibitem{huang2020metapoison}
W~Ronny Huang, Jonas Geiping, Liam Fowl, Gavin Taylor, and Tom Goldstein,
\newblock ``Metapoison: Practical general-purpose clean-label data poisoning,''
\newblock {\em arXiv preprint arXiv:2004.00225}, 2020.

\bibitem{liu2018darts}
Hanxiao Liu, Karen Simonyan, and Yiming Yang,
\newblock ``Darts: Differentiable architecture search,''
\newblock {\em arXiv preprint arXiv:1806.09055}, 2018.

\bibitem{chen2019research}
Zhangyu Chen, Dong Liu, Xiaofei Wu, and Xiaochun Xu,
\newblock ``Research on distributed renewable energy transaction
  decision-making based on multi-agent bilevel cooperative reinforcement
  learning,''
\newblock 2019.

\bibitem{ning2020dsa}
Xuefei Ning, Tianchen Zhao, Wenshuo Li, Peng Lei, Yu~Wang, and Huazhong Yang,
\newblock ``Dsa: More efficient budgeted pruning via differentiable sparsity
  allocation,''
\newblock in {\em European Conference on Computer Vision}. Springer, 2020, pp.
  592--607.

\bibitem{wen2016learning}
Wei Wen, Chunpeng Wu, Yandan Wang, Yiran Chen, and Hai Li,
\newblock ``Learning structured sparsity in deep neural networks,''
\newblock {\em Advances in neural information processing systems}, vol. 29,
  2016.

\bibitem{zhang2018systematic}
Tianyun Zhang, Shaokai Ye, Kaiqi Zhang, Jian Tang, Wujie Wen, Makan Fardad, and
  Yanzhi Wang,
\newblock ``A systematic dnn weight pruning framework using alternating
  direction method of multipliers,''
\newblock in {\em Proceedings of the European Conference on Computer Vision
  (ECCV)}, 2018, pp. 184--199.

\bibitem{hoefler2021sparsity}
Torsten Hoefler, Dan Alistarh, Tal Ben-Nun, Nikoli Dryden, and Alexandra Peste,
\newblock ``Sparsity in deep learning: Pruning and growth for efficient
  inference and training in neural networks,''
\newblock {\em Journal of Machine Learning Research}, vol. 22, no. 241, pp.
  1--124, 2021.

\bibitem{wang2020proximal}
Hao Wang, Xiangyu Yang, Yuanming Shi, and Jun Lin,
\newblock ``A proximal iteratively reweighted approach for efficient network
  sparsification,''
\newblock {\em IEEE Transactions on Computers}, vol. 71, no. 1, pp. 185--196,
  2020.

\bibitem{bach2012optimization}
Francis Bach, Rodolphe Jenatton, Julien Mairal, Guillaume Obozinski, et~al.,
\newblock ``Optimization with sparsity-inducing penalties,''
\newblock {\em Foundations and Trends{\textregistered} in Machine Learning},
  vol. 4, no. 1, pp. 1--106, 2012.

\bibitem{shaham2015understanding}
Uri Shaham, Yutaro Yamada, and Sahand Negahban,
\newblock ``Understanding adversarial training: Increasing local stability of
  neural nets through robust optimization,''
\newblock {\em arXiv preprint arXiv:1511.05432}, 2015.

\bibitem{deng2021meta}
Xiang Deng and Zhongfei~Mark Zhang,
\newblock ``Is the meta-learning idea able to improve the generalization of
  deep neural networks on the standard supervised learning?,''
\newblock in {\em 2020 25th International Conference on Pattern Recognition
  (ICPR)}. IEEE, 2021, pp. 150--157.

\bibitem{csiszar1984information}
Imre Csisz{\'a}r,
\newblock ``Information geonetry and alternating minimization procedures,''
\newblock {\em Statistics and decisions}, vol. 1, pp. 205--237, 1984.

\bibitem{alfarra2020decision}
Motasem Alfarra, Adel Bibi, Hasan Hammoud, Mohamed Gaafar, and Bernard Ghanem,
\newblock ``On the decision boundaries of neural networks: A tropical geometry
  perspective,''
\newblock {\em arXiv preprint arXiv:2002.08838}, 2020.

\bibitem{finn2017model}
Chelsea Finn, Pieter Abbeel, and Sergey Levine,
\newblock ``Model-agnostic meta-learning for fast adaptation of deep
  networks,''
\newblock {\em arXiv preprint arXiv:1703.03400}, 2017.

\bibitem{krizhevsky2009learning}
A.~Krizhevsky and G.~Hinton,
\newblock ``Learning multiple layers of features from tiny images,''
\newblock {\em Master's thesis, Department of Computer Science, University of
  Toronto}, 2009.

\bibitem{le2015tiny}
Ya~Le and Xuan Yang,
\newblock ``Tiny imagenet visual recognition challenge,''
\newblock {\em CS 231N}, vol. 7, no. 7, pp. 3, 2015.

\bibitem{deng2009imagenet}
Jia Deng, Wei Dong, Richard Socher, Li-Jia Li, Kai Li, and Li~Fei-Fei,
\newblock ``Imagenet: A large-scale hierarchical image database,''
\newblock in {\em Computer Vision and Pattern Recognition, 2009. CVPR 2009.
  IEEE Conference on}. IEEE, 2009, pp. 248--255.

\bibitem{he2016deep}
Kaiming He, Xiangyu Zhang, Shaoqing Ren, and Jian Sun,
\newblock ``Deep residual learning for image recognition,''
\newblock in {\em Proceedings of the IEEE conference on computer vision and
  pattern recognition}, 2016, pp. 770--778.

\bibitem{simonyan2014very}
Karen Simonyan and Andrew Zisserman,
\newblock ``Very deep convolutional networks for large-scale image
  recognition,''
\newblock {\em arXiv preprint arXiv:1409.1556}, 2014.

\end{thebibliography}
}}

\appendix

\onecolumn
\setcounter{section}{0}

\section*{Appendix}

\setcounter{section}{0}
\setcounter{figure}{0}
\makeatletter 
\renewcommand{\thefigure}{A\arabic{figure}}
\renewcommand{\theHfigure}{A\arabic{figure}}
\renewcommand{\thetable}{A\arabic{table}}
\renewcommand{\theHtable}{A\arabic{table}}

\makeatother
\setcounter{table}{0}

\setcounter{mylemma}{0}
\renewcommand{\themylemma}{A\arabic{mylemma}}
\setcounter{algorithm}{0}
\renewcommand{\thealgorithm}{A\arabic{algorithm}}
\setcounter{equation}{0}
\renewcommand{\theequation}{A\arabic{equation}}



\section{Derivation of \eqref{eq: exact_IG}}
\label{app: proof}
\label{eq: app_IG}
Based on the fact that the   $\btheta^*(\mask)$ is satisfied with the stationary condition of the lower-level objective function in \eqref{eq: GD_bi_level},
we   obtain 

\vspace*{-6mm}
{
\small
\begin{align}
\displaystyle
  \nabla_{\btheta} g(\mathbf m, \btheta^* )
    = \nabla_{\btheta} \ell(\mask \odot \btheta^* ) + \gamma \btheta^*  = \mathbf 0,
    \label{eq: stationary}
\end{align}
}%
where for ease of notation, we omit the dependence of $\btheta^*(\mask)$ w.r.t. $\mathbf m$.
We   then take    derivative of the second equality of \eqref{eq: stationary} w.r.t. $\mask$ by using the implicit 
function theory. This leads to

\vspace*{-5mm}
{\small
\begin{align}
& \displaystyle
    \nabla^2_{\mask \btheta} \ell (\mask \odot \btheta^* ) + \frac{d\btheta^*(\mask)}{d\mask} 
    \nabla^2_{\btheta} \ell(\mask \odot \btheta^*)
   + \gamma \frac{d\btheta^*(\mask)}{d\mask} 
   = \mathbf 0; \nonumber \\
 \Longrightarrow  &
 \frac{d\btheta^*(\mask)}{d\mask} 
 =  -   \nabla^2_{\mask \btheta} \ell (\mask \odot \btheta^* ) [ \nabla^2_{\btheta } \ell(\mask \odot \btheta^*) + \gamma \mathbf I ]^{-1}.
 \label{eq: exact_IG2}
\end{align}
}%


\section{{\biprune} Algorithm Details}
\label{app: algorithm}

\begin{figure}[htb]
\centerline{
\includegraphics[width=1.0\textwidth]{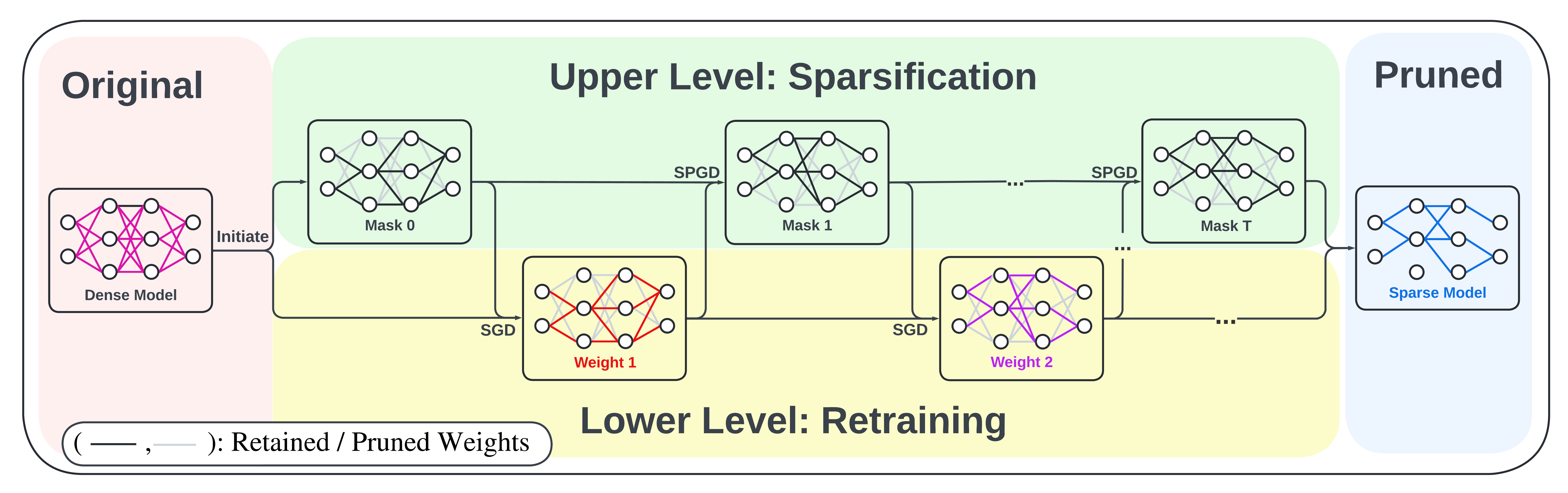}
}
\caption{\footnotesize{
Overview of the {\biprune} pruning algorithm. The {\biprune} algorithm   iteratively   carry out model retraining in the lower level and pruning in the upper level. In the plots, SGD refers to the lower-level stochastic gradient descent update and SPGD refers to the upper-level stochastic projected gradient descent. Masks may vary between each iteration, and the pruned weights are indicated using the light gray color. 
Different colors of the edges in the neural networks refer to the weight update. The arcs in this figure represent the data flow of weights/connections.
}}
\label{fig: algorithm overview}
\end{figure}

At iteration $t$ of {\biprune}, there are two main steps:

\ding{72} \textit{Lower-level SGD for model retraining}: 
Given $\mask^{(t-1)}$, $\btheta^{(t-1)}$, and 
$
\mathbf z^{(t-1)} 
\Def \mask^{(t-1)} \odot \btheta^{(t-1)}$, we update $\btheta^{(t)}$ by applying SGD (stochastic gradient descent)  to the lower-level problem of \eqref{eq: prob_bi_level},%

\vspace*{-5mm}
{\small 
\begin{align}
    \btheta^{(t)} & = \btheta^{(t-1)} - \alpha \nabla_{\btheta} g(\mask^{(t-1)}, \btheta^{(t-1)})  
   \overset{
   \eqref{eq: chainrule_theta}}{=}
    \btheta^{(t-1)} - \alpha [ \mask^{(t-1)} \odot 
    \nabla_{\mathbf z} \ell(\mathbf z )\left | \right._{\mathbf z = \mathbf z^{(t-1)}} + \gamma \btheta^{(t-1)} ],
    \tag{$\btheta$-step}
    \label{appeq: theta_step}
\end{align}}%
where $\alpha > 0$ is the lower-level   learning rate.

\ding{72} \textit{Upper-level SPGD for pruning}: Given $\mask^{(t-1)}$, $ \btheta^{(t)}$, and $\mathbf z^{(t+1/2)} \Def \mask^{(t-1)} \odot \btheta^{(t)}$, we update   $\mask$ using PGD (projected gradient descent)  along the IG-enhanced descent direction \eqref{eq: GD_bi_level},%


\vspace*{-5mm}
{\small
\begin{align}
    \mask^{(t)} = & \mathcal P_{\mathcal S} \left [ 
    \mask^{(t-1)}  - \beta  \frac{d \ell(\mask \odot \btheta^{(t)})}{d \mask} \left | \right._{\mask =     \mask^{(t-1)} }
    \right ] \nonumber \\
\overset{\eqref{eq: upper_update_final}}{=}    
    &\mathcal P_{\mathcal S} \left [ 
    \mask^{(t-1)}  - \beta  \left ( \btheta^{(t)} - 
    \frac{1}{\gamma} \mask^{(t-1)} \odot \gradzsimple \left | \right._{\mathbf z = \mathbf z^{(t+1/2)}} \right ) \odot \gradzsimple  \left | \right._{\mathbf z = \mathbf z^{(t+1/2)}}
    \right ],
    \tag{$\mask$-step}
    \label{eq:m-step-c2}
\end{align}}%
where $\beta > 0$ is the upper-level learning rate, and $\mathcal P_{\mathcal S}(\cdot)$ denotes the Euclidean projection   onto the constraint set $\mathcal S$ given by
$\mathcal S = \{ \mathbf m\,|\, \mathbf m \in \{ 0, 1\}^n, \mathbf  1^T \mathbf m \leq k \}$ in \eqref{eq: prob_bi_level} and is achieved by the top-$k$ hard-thresholding operation as will be detailed below.  

\paragraph{Implementation  of discrete optimization.}
In the actual implementation, we use $\widetilde \mask^{(t)} \in [0, 1]^d$ and obtain $\mask^{(t)} \in \{0, 1\}^d$ where $\mask^{(t)} \gets \mathcal P_{\mathcal S} \left[ \widetilde \mask^{(t)} \right]$. The \eqref{eq:m-step-c2} is then implemented as the following:

\begin{align}
\widetilde \mask^{(t)} = 
    &
    \widetilde \mask^{(t-1)}  - \beta  \left ( \btheta^{(t)} - 
    \frac{1}{\gamma} \widetilde \mask^{(t-1)} \odot \gradzsimple \left | \right._{\mathbf z = \mathbf z^{(t+1/2)}} \right ) \odot \gradzsimple  \left | \right._{\mathbf z = \mathbf z^{(t+1/2)}}
    \tag{$\widetilde \mask$-step}
    \label{eq:soft-m-step}
\end{align}

and then $\mask^{(t)} \gets \mathcal P_{\mathcal S} \left[ \widetilde \mask^{(t)} \right]$, with $\mathbf z^{(t+1/2)} := \mask^{(t-1)} \odot \btheta^{(t)}$ as defined above. 


\begin{algorithm}[H]
        \caption{{\biprune}}
        \begin{algorithmic}[1]
          \State \textbf{Initialize:} Model $\boldsymbol \theta_0$, pruning mask score $\mask_0$, binary mask $\mathbf{z}^*$, sparse ratio $p\%$,  regularization parameter $\lambda$, upper- and lower-level learning rate $\alpha$ and $\beta$.
          \For{Iteration $t=0,1,\ldots,$}
          \State Pick \emph{different} random data batches $\mathcal B_\alpha$ and $\mathcal B_\beta$ for different levels of tasks.
        \State \textbf{Lower-level}: Update model parameters using data batch $\mathcal{B}_\beta$ via SGD calling:
        
        \vspace*{-1mm}
        {\small
        \begin{align}
        \displaystyle
             \btheta_{t+1} = \btheta_t - \beta \frac{d \ell_{\mathrm{tr}}(\mask \odot \btheta)}{d \btheta } \left. \Bigr|_{\mask = \mathbf{z}^*, \btheta=\btheta_t} \right .
         \label{eq: SGD}
        \end{align}
        }%
        
     \State \textbf{Upper-level}: Update pruning mask score using data batch $\mathcal{B}_\alpha$ via SGD calling:
     {
     \small
     \begin{align}
     \displaystyle
     \raisetag{6mm}
         \hspace*{-3mm} \mask_{t+1} = \mask_t - \alpha \left (\nabla_{\mask}\ltr({\mask} \odot {\btheta}) - 
    {\frac{1}{\gamma}\nabla_{\mathbf z} \ltr (\mathbf z)\left |_{\mathbf z = {\mask} \odot {\btheta}} \right.  \odot \nabla_{{\btheta}} \ltr ({\mask}\odot {\btheta})
    } \right) \Bigr |_{\mask=\mask_t, \btheta=\btheta_{t+1}}
     \end{align}
     }%
    \State \textbf{ Update the binary mask $\mathbf{z}^*$}: Hard-threshold the mask score $\mask$ with the give sparse ratio $p$:
    {\small
    \begin{align}
    \displaystyle
        \mathbf{z}^* = \mathcal T_{
        \{0,1\}^d} (\mask_{t + 1}, s).
        \label{eq: threshold}
    \end{align}
    }%
      \EndFor
        \end{algorithmic}
        \label{alg: BiP}
      \end{algorithm}

\section{Additional Experimental Details and Results}

\subsection{Datasets and Models}
Our dataset and model choices follow the pruning benchmark in\,\citep{ma2021sanity}. We summarize the datasets and model configurations in Tab.\,\ref{table: impl_details}. 
In particular, we would like to stress that we adopt the ResNet-18 with convolutional kernels of $3\times3$ in the first layer for Tiny-ImageNet, aligned with CIFAR-10 and CIFAR-100, compared to ImageNet ($7 \times 7$). See \url{https://github.com/kuangliu/pytorch-cifar/blob/master/models/resnet.py} for more details.

\subsection{Detailed Training Settings}
\label{app: training}

\paragraph{Baselines.}
For both unstructured and structured pruning settings, we consider four baseline methods across various pruning categories, including IMP\,\cite{frankle2018lottery}, OMP\,\cite{frankle2018lottery}, {\hydra}\,\cite{sehwag2020hydra} and {\grasp}\,\cite{wang2020pick}. For {\hydra} and {\grasp}, we adopt the original setting as well as hyper-parameter choices on their official code repositories. For IMP and OMP, we adopt the settings from the current SOTA implementations\,\citep{ma2021sanity}. 
Details on the pruning schedules 
can be found in Tab.\,\ref{tab: efficiency}.
In particular, {\hydra} prunes the dense model to the desired sparsity with 100 epoch for pruning and 100 epoch for retraining. {\grasp}   conducts one-shot pruning to the target sparsity, followed by the 200-epoch retraining. In each pruning iteration, IMP prunes 20\% of the remaining parameters before 160-epoch retraining. 
{\hydra} adopts the cosine learning rate scheduler for both pruning and retraining stage.
The learning rate scheduler for IMP, OMP, and {\grasp} is the step learning rate with a learning rate decay rate of 0.1 at 50\% and 75\% epochs. The initial learning rate for all the methods are   0.1. 

\begin{table}[!t]
\centering
\caption{\footnotesize{Dataset and model setups. The following parameters are shared across all the methods.
}}
\label{table: impl_details}
\begin{adjustbox}{width=0.90\textwidth}
\begin{tabular}{l|cccccccc}
\toprule
\multirow{2}{*}{Settings} & \multicolumn{4}{c}{CIFAR-10} & \multicolumn{2}{c}{CIFAR-100} & \multicolumn{1}{c}{Tiny-ImageNet} & \multicolumn{1}{c}{ImageNet}\\ \cmidrule(lr){2-5} \cmidrule(lr){6-7} \cmidrule(lr){8-8} \cmidrule(lr){9-9}    
& RN-18 & RN-20 & RN-56 & VGG-16 & RN-18 & RN-20  & RN-18 & RN-50\\ \midrule
Batch Size & 64 & 64 & 64 & 64 & 64 & 64 & 32 & 1024 \\ \midrule
Model Size & $11.22$ M & $0.27$ M & $0.85$ M & $14.72$ M & $11.22$ M & $0.27$ M & $11.22$ M & $25.56$ M \\
\bottomrule
\end{tabular}
\end{adjustbox}
\end{table}


\paragraph{Hyper-parameters for {\biprune}.}
In both structured and unstructured pruning settings, cosine learning rate schedulers are adopted, and {\biprune} takes an initial learning rate of 0.1 for the upper-level problem (pruning) and 0.01 for the lower-level problem (retraining). 
The lower-level regularization coefficient   $\lambda$ is set to $1.0$ throughout the experiments. By default, we only take one SGD step for lower-level optimization in all settings. \RV{Ablation studies on different SGD steps for lower-level optimization can be found in Fig.~\ref{fig: ablation_study}(b) and Fig.~\ref{fig: step_convergence_rebuttal}. We use 100 training epochs for {\biprune}, and ablation studies on different training epochs for larger pruning ratios can be found in Fig.~\ref{fig: epoch_extreme_sparsity_rebuttal}}.

\begin{table}[!t]
\vspace*{0mm}
\centering
\caption{\footnotesize{Computation complexities of different pruning methods on (CIFAR-10, ResNet-18) in unstructured pruning setting. The training epoch numbers of pruning/retraining  baselines are consistent with their official settings or the latest benchmark implementations. All the evaluations are based on a single Tesla-V100 GPU. 
}}
\label{tab: efficiency}
\resizebox{.7\columnwidth}{!}{%
\begin{tabular}{c|cccc|c}
\toprule[1pt]
\midrule
\multicolumn{5}{c|}{{Runtime v.s. targeted sparsity}} & \multirow{2}{*}{\begin{tabular}[c]{@{}c@{}} Training epoch \# \end{tabular}} \\
\multicolumn{1}{c|}{Method $\backslash$ Sparsity} &
  \multicolumn{1}{c|}{$20\%$} &
  \multicolumn{1}{c|}{$59\%$} &
  \multicolumn{1}{c|}{$83.2\%$} &
  $95.6\%$ &
   \\ \midrule
\multicolumn{1}{c|}{IMP} &
  \multicolumn{1}{c|}{69 min} &
  \multicolumn{1}{c|}{276 min} &
  \multicolumn{1}{c|}{621 min} &
  966 min &
  160 epoch retrain \\ \midrule
\grasp &
  \multicolumn{4}{c|}{89 min} &
  200 epoch retrain \\ \midrule
OMP &
  \multicolumn{4}{c|}{69 min} &
  160 epoch retrain \\ \midrule
\hydra &
  \multicolumn{4}{c|}{115 min} &
  \begin{tabular}[c]{@{}c@{}}100 epoch prune\\ 100 epoch retrain\end{tabular} \\ \midrule
\biprune &
  \multicolumn{4}{c|}{86 min} &
  100 epoch         \\ \midrule
\bottomrule[1pt]
\end{tabular}%
}
\end{table}

\begin{table}[]
\centering
\caption{\footnotesize{Detailed training details for each method. All the baselines adopt the recommended settings either from the official or their latest benchmark (\emph{e.g.}, LTH\cite{ma2021sanity}) for a fair comparison. Note by default setting, only our method {\biprune} do not require additional epochs for retraining.}}
\label{tab: training_details}
\resizebox{\columnwidth}{!}{%
\begin{tabular}{c|ccccccccc}
\toprule[1pt]
\midrule
Method &
  Epoch Number &
  \begin{tabular}[c]{@{}c@{}}Initial \\ Learning Rate\end{tabular} &
  \begin{tabular}[c]{@{}c@{}}Learning Rate\\  Scheduler\end{tabular} &
  \begin{tabular}[c]{@{}c@{}}Learning Rate \\ Decay Factor\end{tabular} &
  \begin{tabular}[c]{@{}c@{}}Learning Rate \\ Decay Epoch\end{tabular} &
  Mementum &
  \begin{tabular}[c]{@{}c@{}}Weight \\ Decay\end{tabular} &
  \begin{tabular}[c]{@{}c@{}}Rewind \\ Epoch\end{tabular} &
  Warm-up \\ \midrule
IMP &
  160 for Retrain &
  0.1 &
  Step LR &
  10 &
  80/120 &
  0.9 &
  5.00E-04 &
  8 &
  75 for VGG16 \\ \midrule
OMP &
  160 for Retrain &
  0.1 &
  Step LR &
  10 &
  80/120 &
  0.9 &
  5.00E-04 &
  8 &
  75 for VGG16 \\\midrule
{\hydra} &
  \begin{tabular}[c]{@{}c@{}}100 for Prune\\ 100 for Retrain\end{tabular} &
  0.1 &
  Cosine LR &
  N/A &
  N/A &
  0.9 &
  5.00E-04 &
  N/A &
  75 for VGG16 \\\midrule
{\grasp} &
  200 for Retrain &
  0.1 &
  Step LR &
  10 &
  100/150 &
  0.9 &
  5.00E-04 &
  N/A &
  75 for VGG16 \\\midrule
{\biprune} &
  100 &
  0.1 for $\mask$; 0.01 for $\btheta$ &
  Cosine LR &
  N/A &
  N/A &
  0.9 &
  5.00E-04 &
  N/A &
  75 for VGG16 \\ \midrule
\bottomrule[1pt]
\end{tabular}%
}
\end{table}


\begin{figure}[htb]
\centerline{
\begin{tabular}{cc}
\hspace*{0mm}\includegraphics[width=.39\textwidth,height=!]{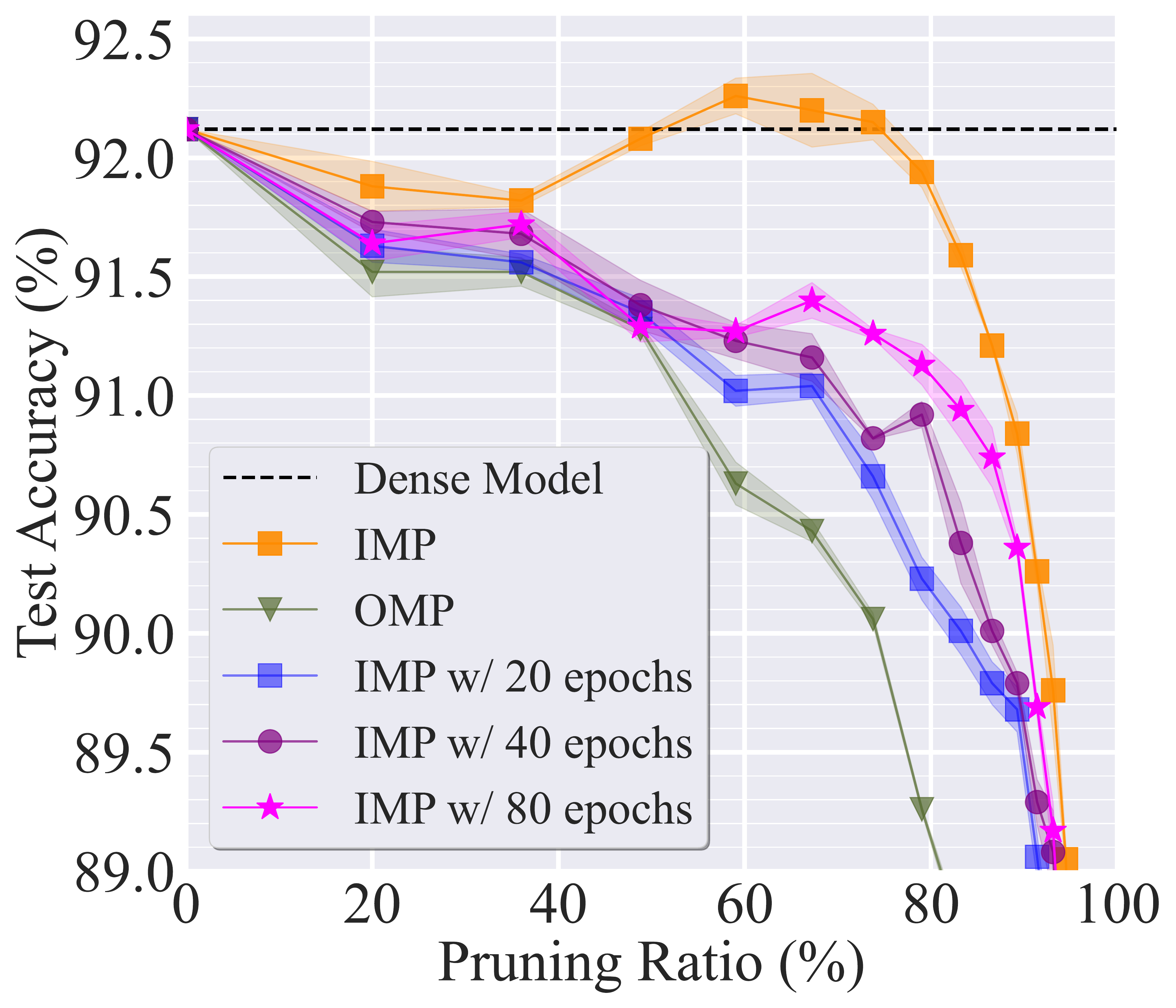}  
&\hspace*{-4mm}\includegraphics[width=.39\textwidth,height=!]{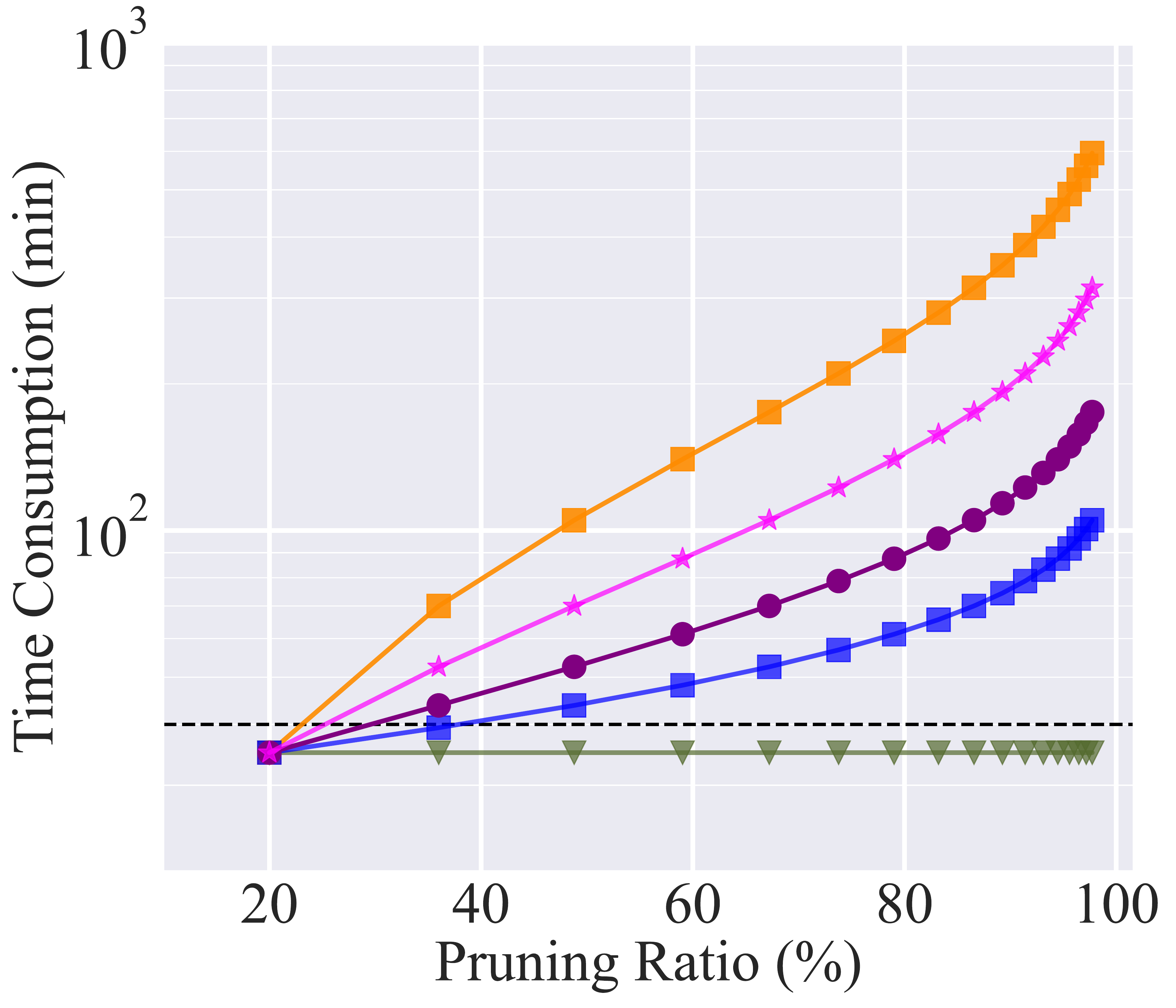}
\end{tabular}}
\vspace*{-3mm}
\caption{\footnotesize{
Performance comparisons among OMP, IMP, and IMP with less retraining epochs on CIFAR-10 with ResNet-20. 
}}
  \label{fig: compare_IMP_OMP_GMP}
\end{figure}

\begin{figure}[t]
\centerline{
\begin{tabular}{cc}
\hspace*{0mm}\includegraphics[width=.45\textwidth,height=!]{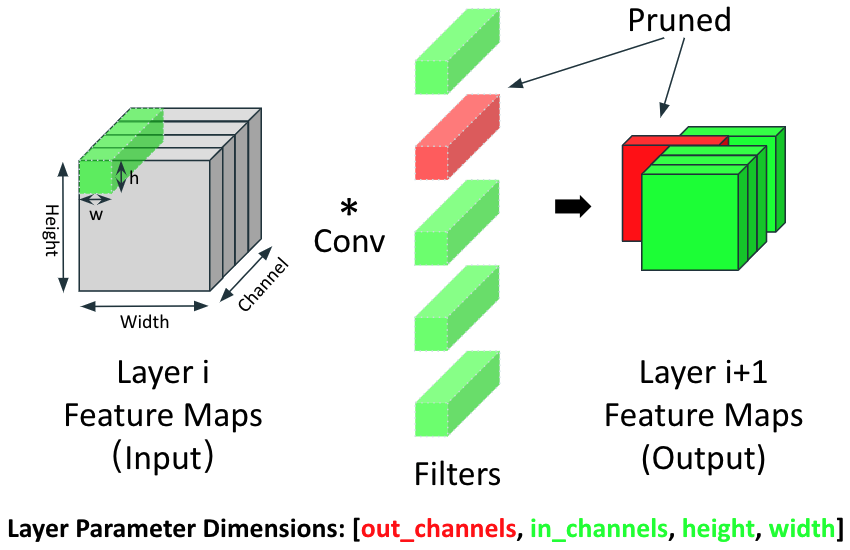}  
&\hspace*{0mm}\includegraphics[width=.45\textwidth,height=!]{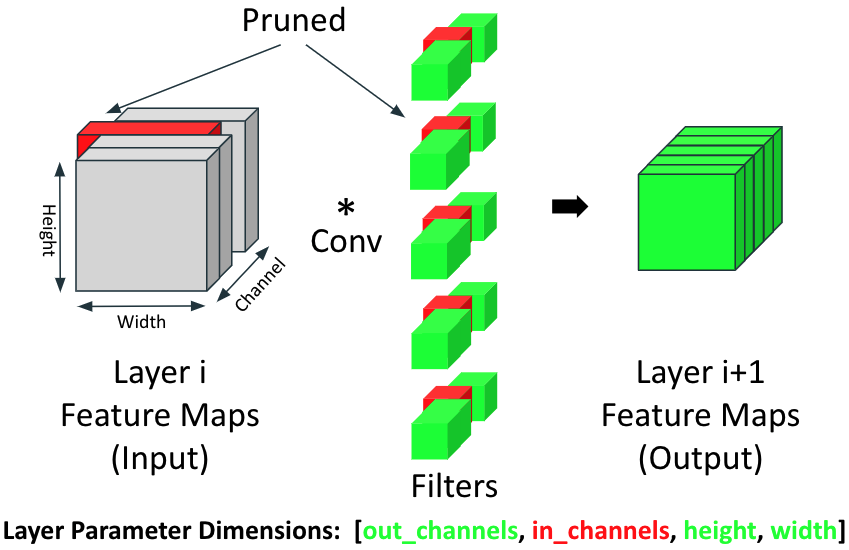}
\\
\footnotesize{(a) Filter-wise Pruning} &   \footnotesize{(b) Channel-wise Pruning}
\end{tabular}}
\caption{\footnotesize{Illustration of filter-wise pruning and channel-wise pruning. The blocks in the middle column in (a) and (b) represent the parameters (filters) of the $i$th convolutional layer, where the red ones represent the pruning unit in each setting. The left blocks in gray denote the input feature maps and the right columns denote the output feature maps generated by the corresponding filters marked in the same color.}}
  \label{fig: illustration}
\end{figure}

\paragraph{Structured pruning.} To differentiate of the filter-wise and channel-wise structured pruning setting, we illustrate the details of these settings in Fig.\,\ref{fig: illustration}. Note, the filter-wise pruning setting prunes the output dimension (output channel) of the parameters in one layer, while the channel-wise prunes the input dimension (input channel).

\subsection{Additional Experiment Results}
\label{app: add_results}

\paragraph{Comparison with   IMP using reduced retraining epochs.}

As IMP is significantly more time-consuming than one-shot pruning methods, a natural way to improve the efficiency is to decrease the retraining epoch numbers at each pruning cycle. In Fig.\,\ref{fig: compare_IMP_OMP_GMP}, the performance and time consumption of IMP using 20, 40, and 80 epochs at   each retraining cycle are presented. The results and conclusions are in general aligned with Fig.\,\ref{fig: one_shot_results_merged}. \underline{First}, with fewer epoch numbers, the time consumption decreases at the cost of evident performance degradation. \underline{Second}, IMP with fewer epoch numbers are unable to obtain winning tickets. 
Thus, the direct simplification of IMP would hamper the pruning accuracy.
This experiment   shows the difficulty of    achieving efficient and effective pruning  under the scope of heuristics-based pruning, and thus   justifies the necessity in 
developing a more powerful optimization-based pruning method.

\paragraph{Comparison with more prune-at-initialization baselines.}

\begin{figure}[thb]
\centerline{
\includegraphics[width=.39\textwidth,height=!]{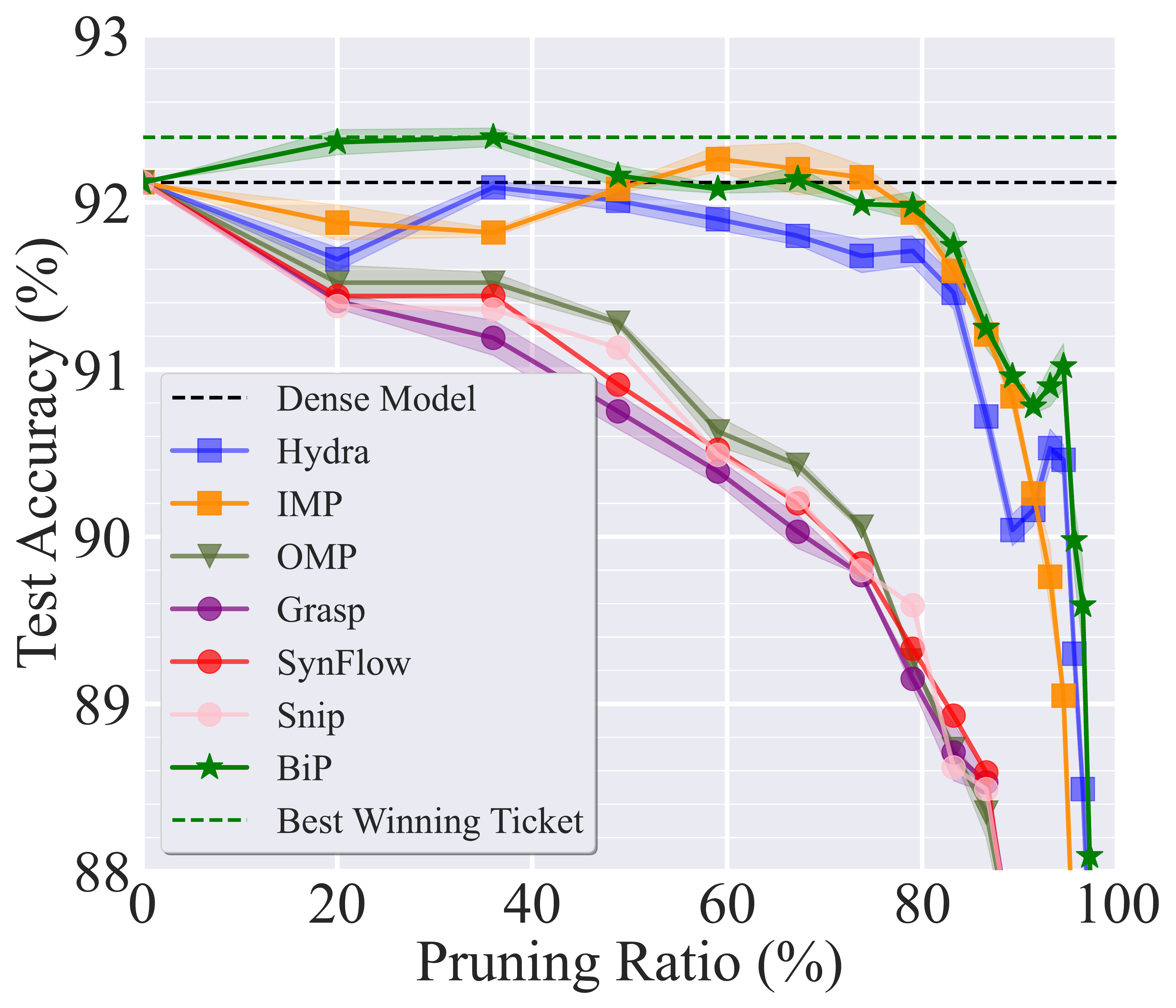}
}
\caption{\footnotesize{Unstructured pruning performance   of {\biprune} vs. prune-at-initialization baselines on (CIFAR-10, ResNet-20). 
}}
\label{fig: compare_init}
\end{figure}

In Fig.\,\ref{fig: compare_init}, we include more heuristics-based one-shot pruning baselines (\synflow\,\citep{tanaka2020pruning}, \snip\,\cite{lee2018snip})
for comparison. Together with {\grasp}, these methods belong to the category of pruning at initialization, which  determines the sparse sub-networks prior to training. As we can see, the advantage of our method over the newly added methods are clear, and the benefit becomes more significant as the sparsity increases. 
This further demonstrates the superiority of the optimization-basis of {\biprune} over the heuristics-based one-shot methods.

\paragraph{{Experiments on unstructured pruning with more baselines.}}

\begin{figure}[htb]
\centerline{
\begin{tabular}{ccc}
    \hspace*{-2mm} \includegraphics[width=.3\textwidth,height=!]{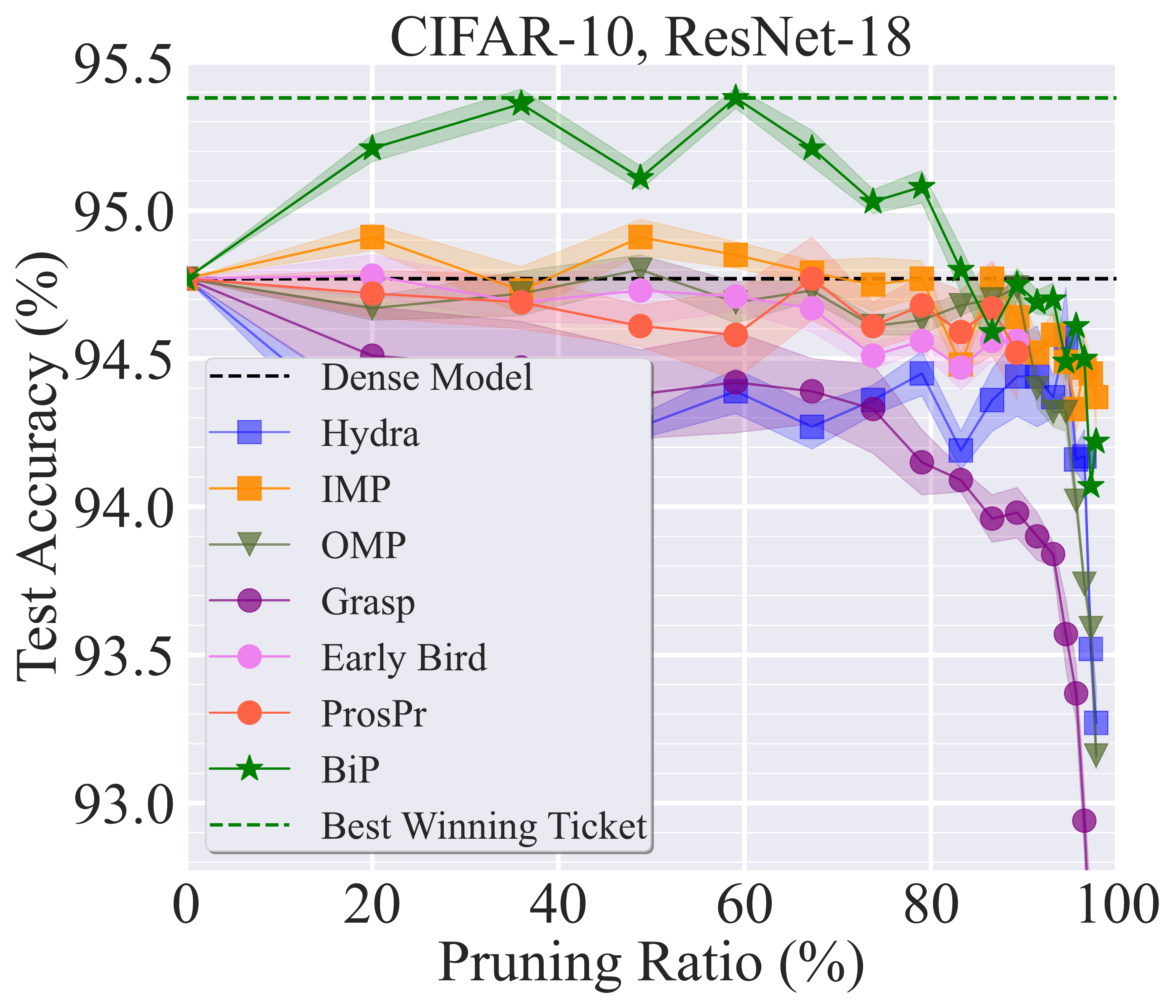} &
    \hspace*{-5mm}  \includegraphics[width=.3\textwidth,height=!]{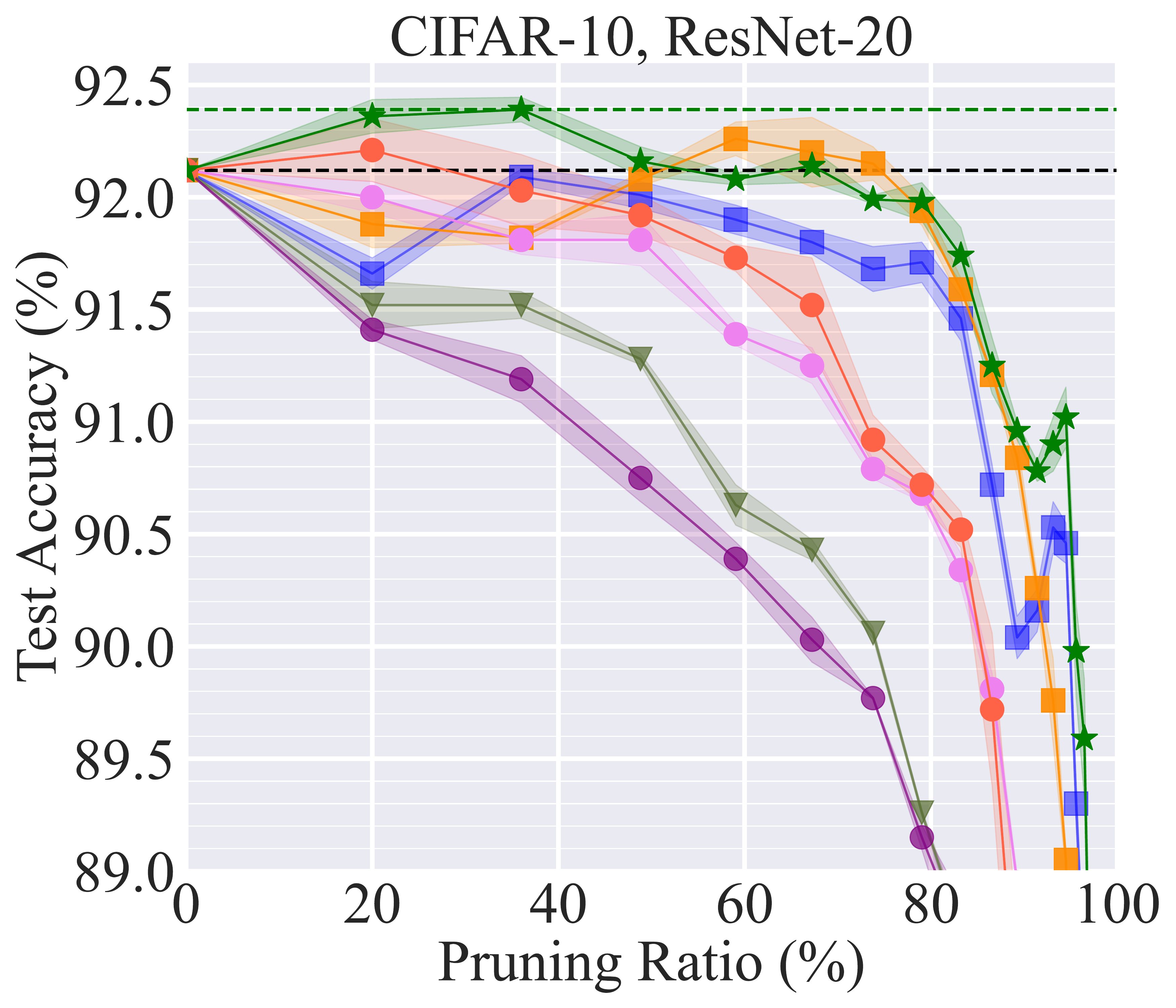} &
    \hspace*{-5mm}  \includegraphics[width=.3\textwidth,height=!]{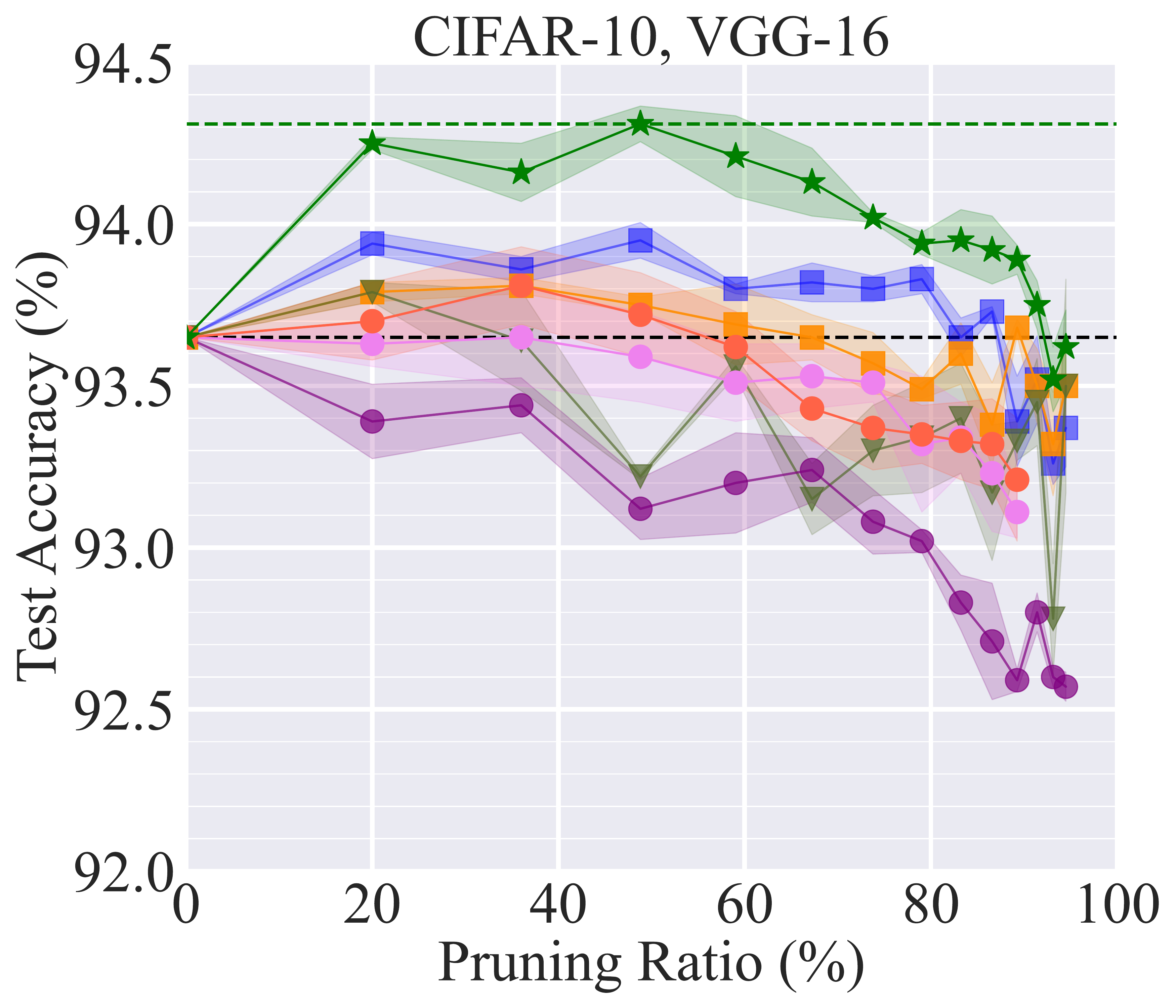} \\
    
    \hspace*{-2mm} \includegraphics[width=.3\textwidth,height=!]{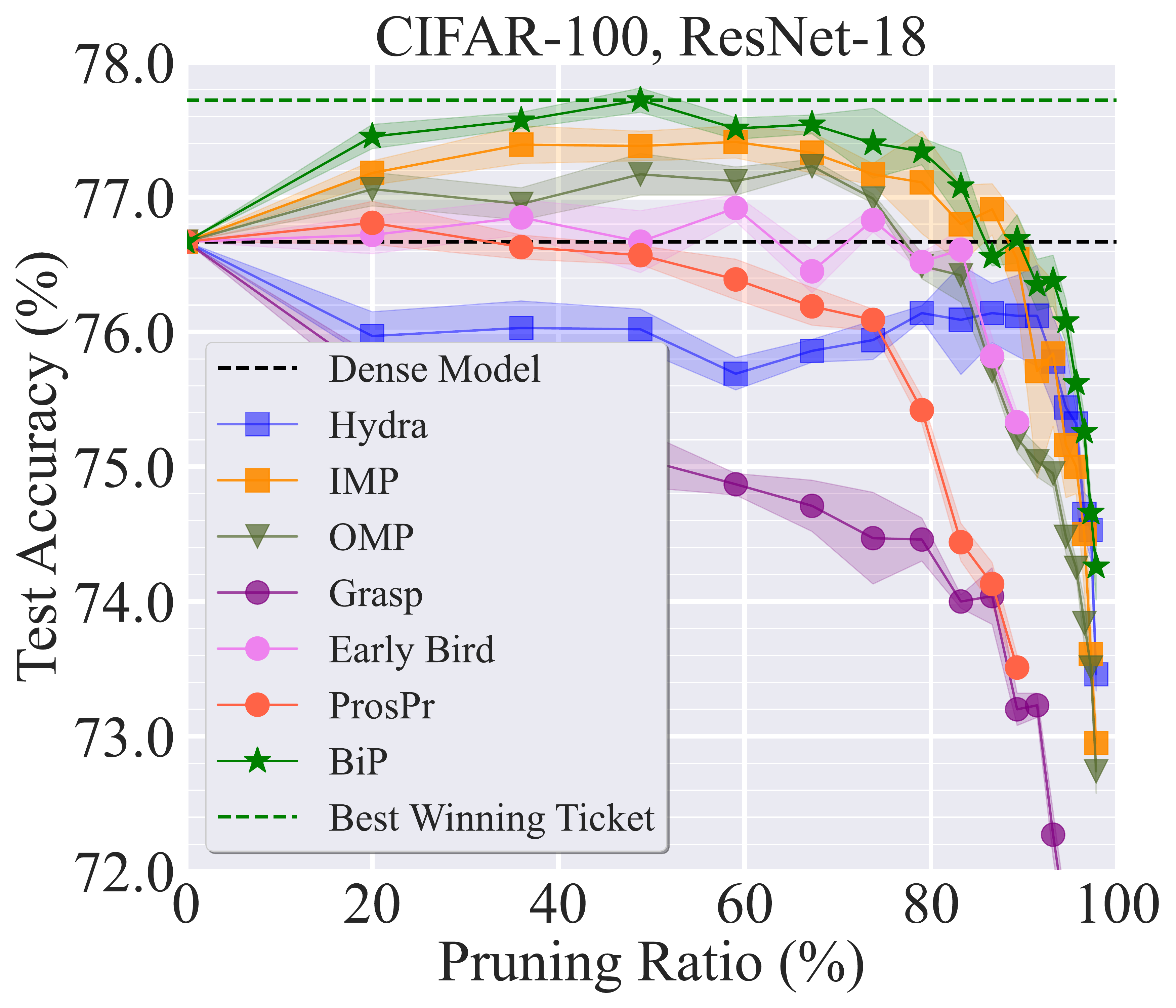} &
    \hspace*{-5mm}  \includegraphics[width=.3\textwidth,height=!]{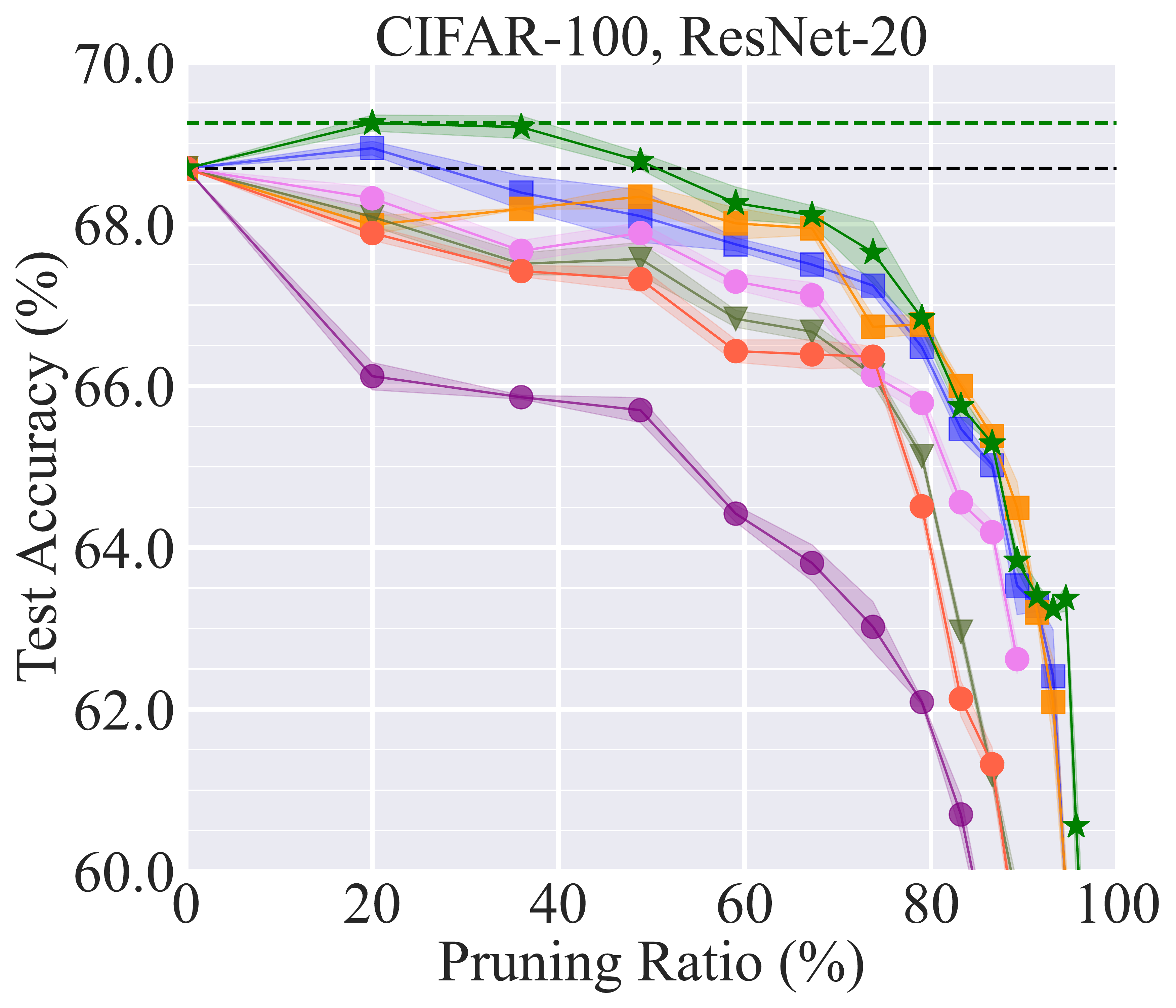} &
    \hspace*{-5mm}  \includegraphics[width=.3\textwidth,height=!]{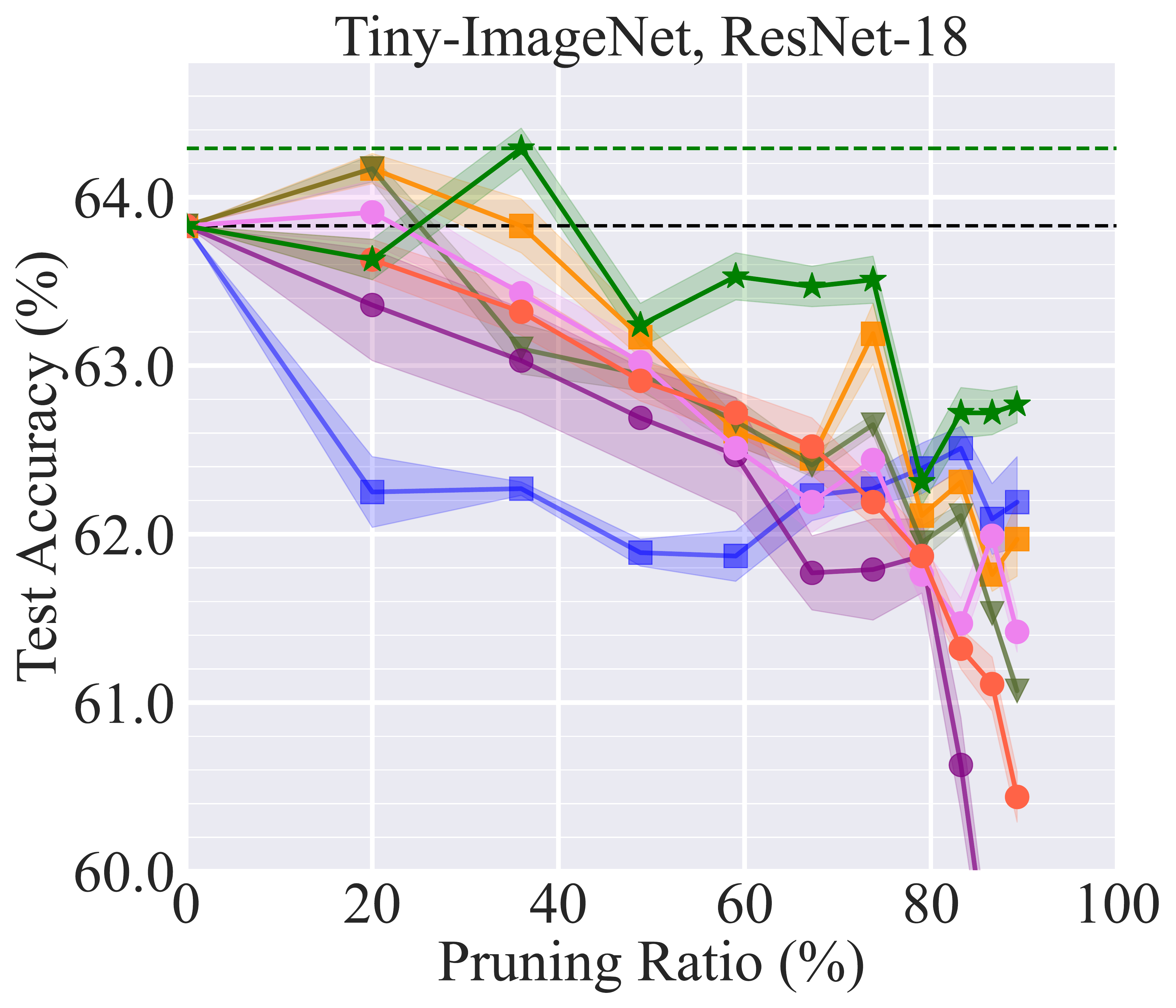} 
\end{tabular}}
\caption{\footnotesize{Unstructured pruning trajectory given by test accuracy (\%) vs. pruning ratio (\%). The visual presentation setting is consistent with Fig.\,\ref{fig: unstructured_performance_cifar}. We consider two more baseline methods: \textsc{EarlyBird} \cite{you2019drawing} and \textsc{ProsPr} \cite{alizadeh2022prospect}.}}
\label{fig: unstructured_rebuttal}
\end{figure}

{We compare our proposed method {\biprune} to more baselines on different datasets and architectures in Fig.\,\ref{fig: unstructured_rebuttal}. We add two more baselines, including \textsc{EarlyBird} \cite{you2019drawing} and \textsc{ProsPr} \cite{alizadeh2022prospect}.} The results show that \textsc{ProsPr} is indeed better than {\grasp} but is still not as good as IMP and our method {\biprune} in different architecture-dataset combinations. Meanwhile, except for the unstructured pruning settings of ResNet18 pruning over CIFAR10 and CIFAR100, \textsc{ProsPr}, as a pruning before training, can achieve comparable performance to the state-of-the-art implementation of OMP. However, the gap between this SOTA pruning-at-initialization method and our method still exists. Besides, the result shows that \textsc{EarlyBird} can effectively achieve comparable or even better testing performance than OMP in most different architecture-dataset combinations, which is also the main contribution of \citep{you2019drawing}. However, \textsc{EarlyBird} is still not as strong as IMP in testing performance. 

\paragraph{{Experiments on structured pruning with more baselines.}}
\begin{figure}[htb]
\centerline{
\begin{tabular}{cccc}
    \hspace*{-2mm} \includegraphics[width=.25\textwidth,height=!]{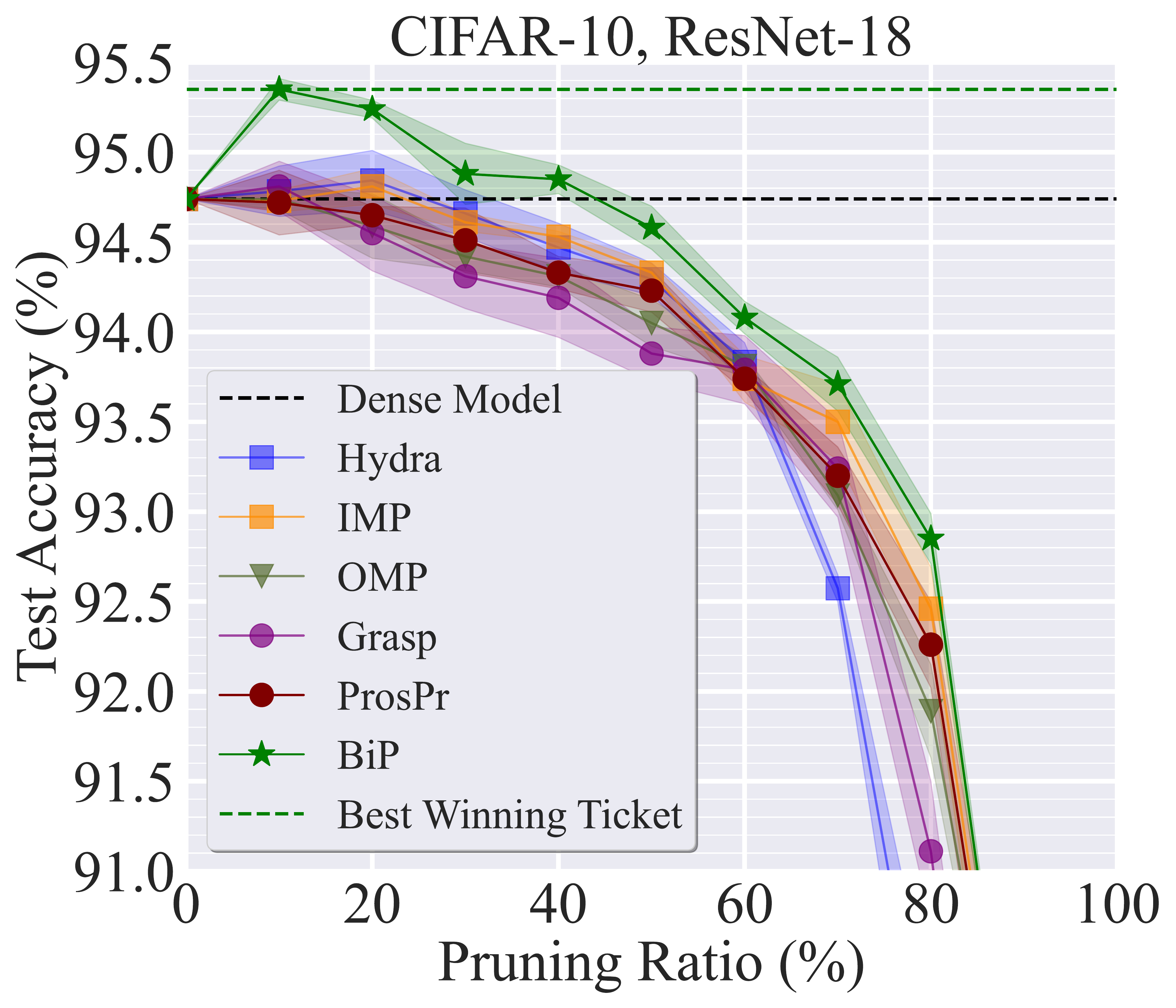} &
    \hspace*{-5mm}  \includegraphics[width=.25\textwidth,height=!]{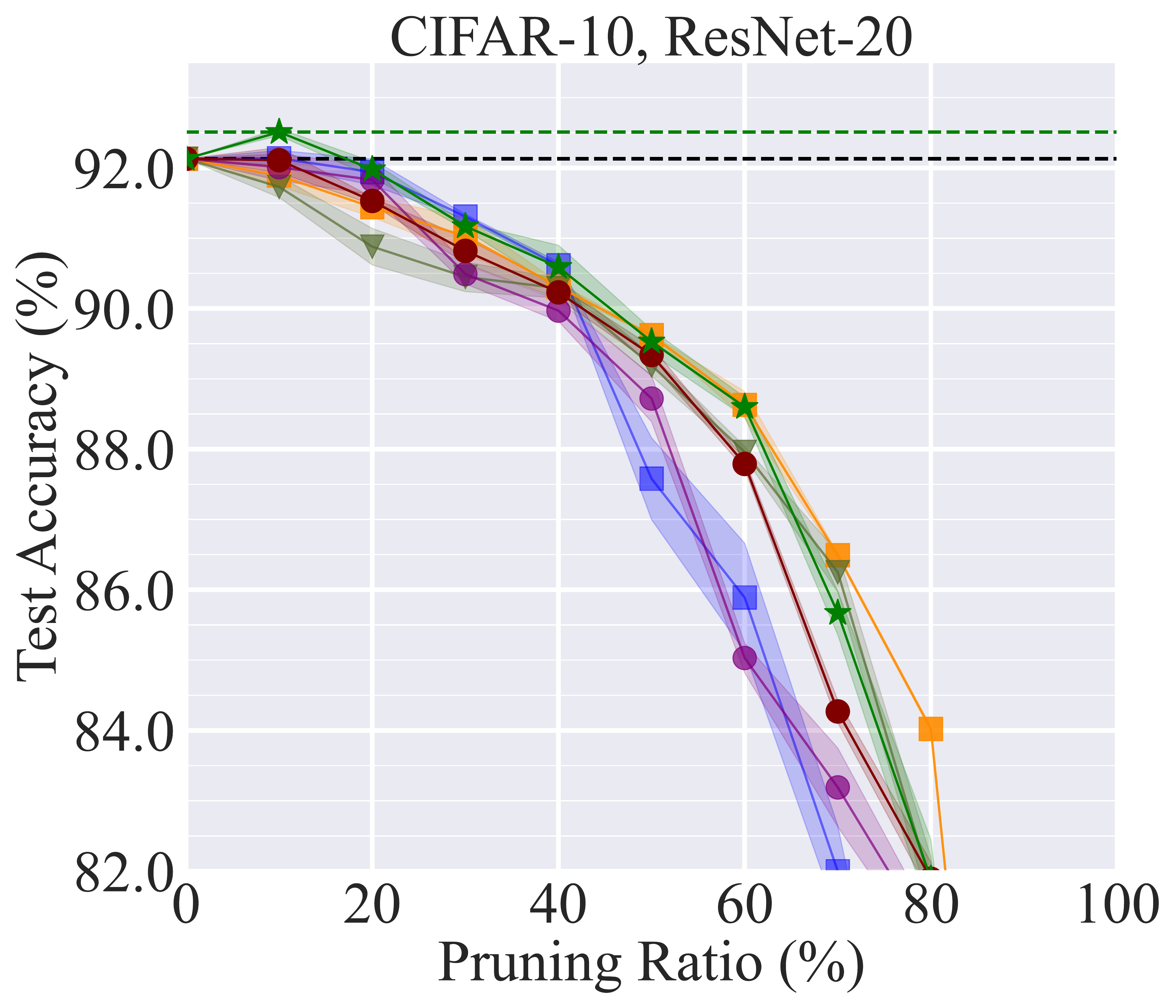} &
    \hspace*{-5mm}  \includegraphics[width=.25\textwidth,height=!]{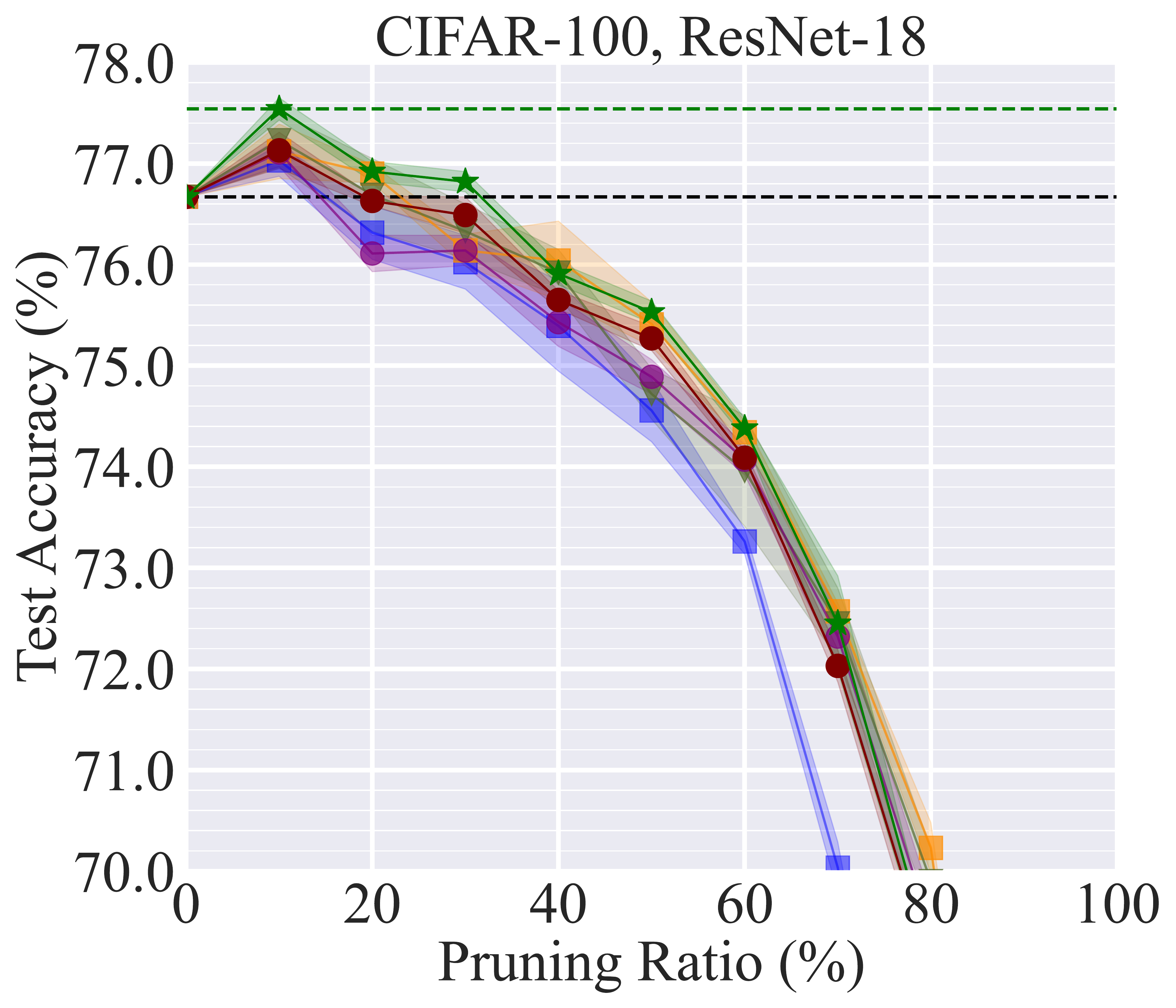} &
    \hspace*{-5mm} \includegraphics[width=.25\textwidth,height=!]{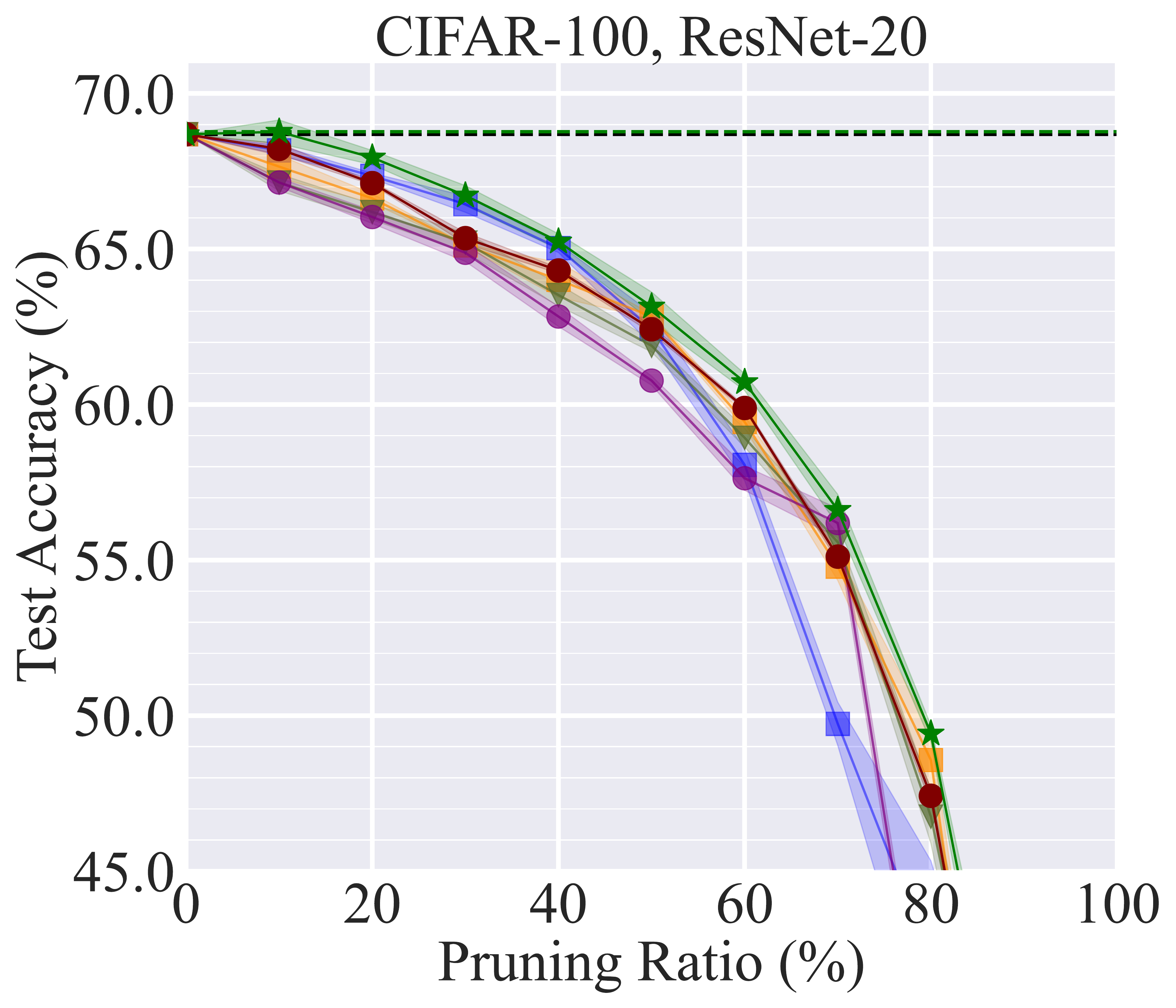} 
\end{tabular}}
\caption{\footnotesize{Structured pruning trajectory given by test accuracy (\%) vs. pruning ratio (\%).  The visual presentation setting is consistent with Fig.\,\ref{fig: structured_performance_cifar}. We add \textsc{ProsPr} \cite{alizadeh2022prospect} as our new baseline. }}
\label{fig: structured_rebuttal}
\end{figure}

We compare our proposed method {\biprune} to the new baseline \textsc{ProsPr} on different datasets and architectures in Fig.\,\ref{fig: structured_rebuttal}. on the structured pruning setting. As we can see, {\biprune} consistently outperforms \textsc{ProsPr} and still stands top among all the baselines.

\begin{figure}[htb]
\centerline{
\includegraphics[width=.35\textwidth,height=!]{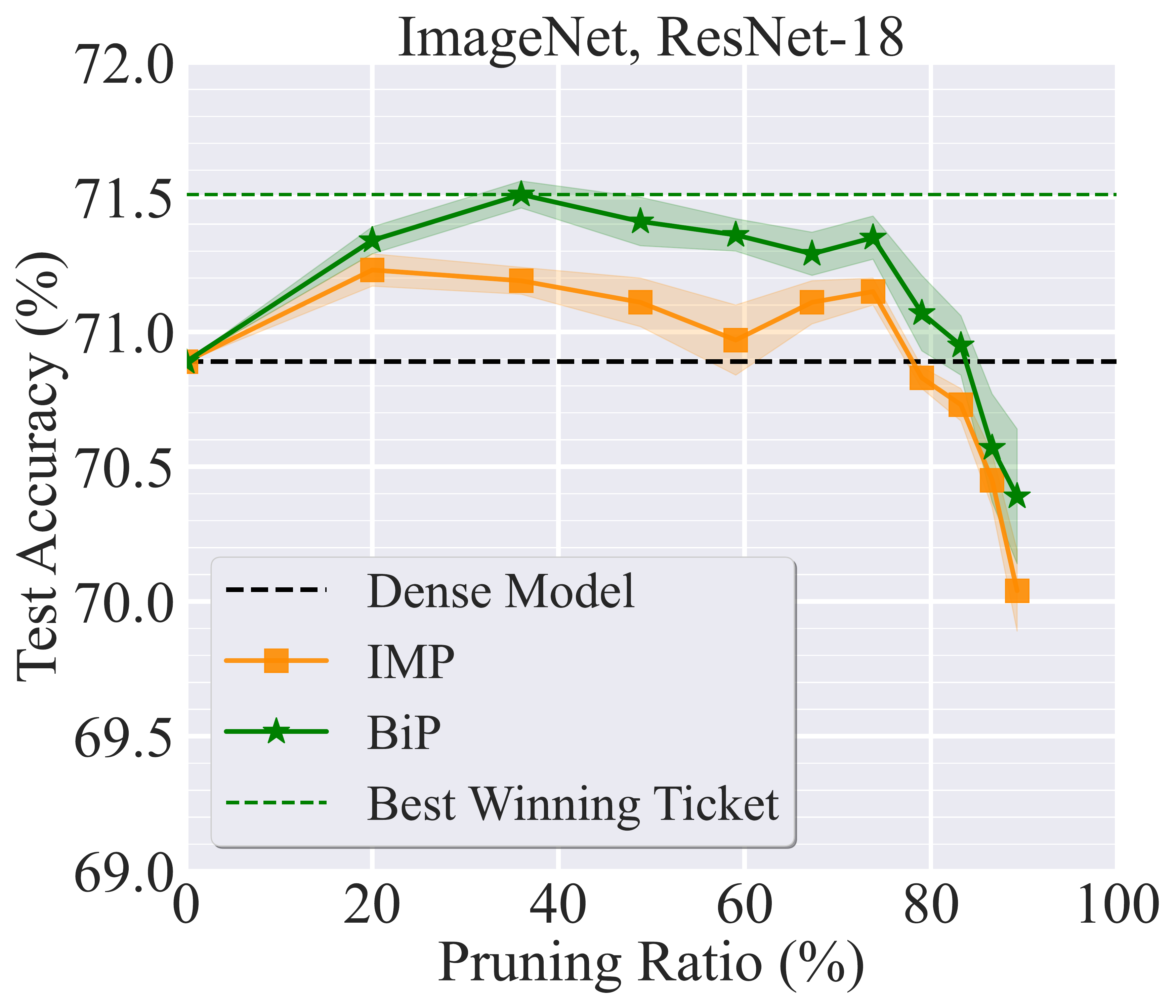}
}
\caption{\footnotesize{Unstructured pruning trajectory on ImageNet with ResNet-18. The experiment setting is consistent with Fig.\,\ref{fig: unstructured_performance_cifar}}. We only compare {\biprune} with our strongest baseline IMP due to limited computational resource.}
\label{fig: resnet18_imagenet_rebuttal}
\end{figure}

\paragraph{More results on ImageNet.} In Fig.\,\ref{fig: resnet18_imagenet_rebuttal}, we provide additional results on the dataset ImageNet with ResNet-18 in the unstructured pruning setting, in addition to the results of (ImageNet, ResNet-50) shown in Fig.\,\ref{fig: unstructured_performance_cifar}. As we can see, the performance of {\biprune} still outperforms the strongest baseline IMP and the same conclusion can be drawn as Fig.\,\ref{fig: unstructured_performance_cifar}.

\begin{figure}[t]
\centerline{
\begin{tabular}{cccc}
    \hspace*{-2mm} \includegraphics[width=.25\textwidth,height=!]{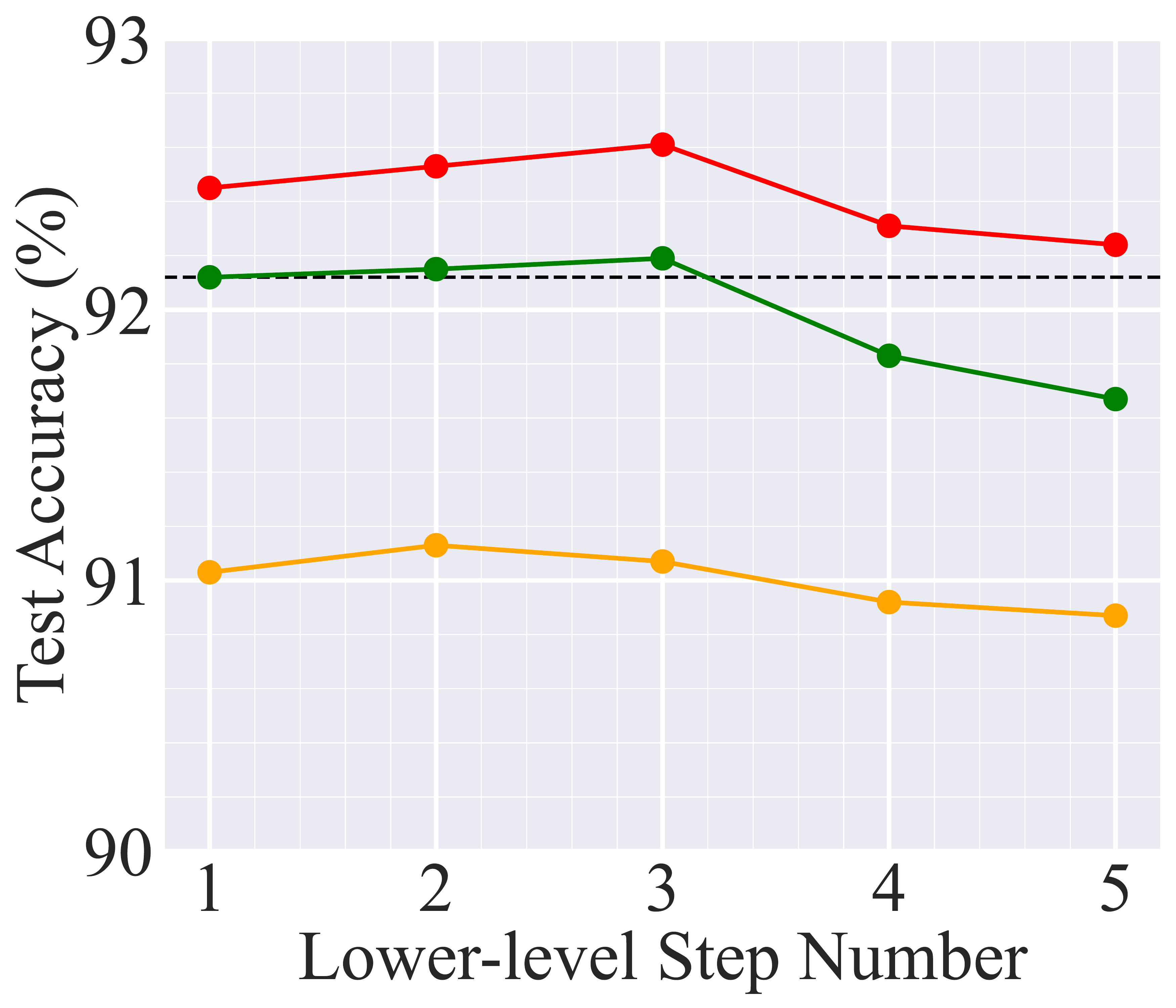} \vspace*{-0.5mm}
    &
    \hspace*{-5mm}  \includegraphics[width=.25\textwidth,height=!]{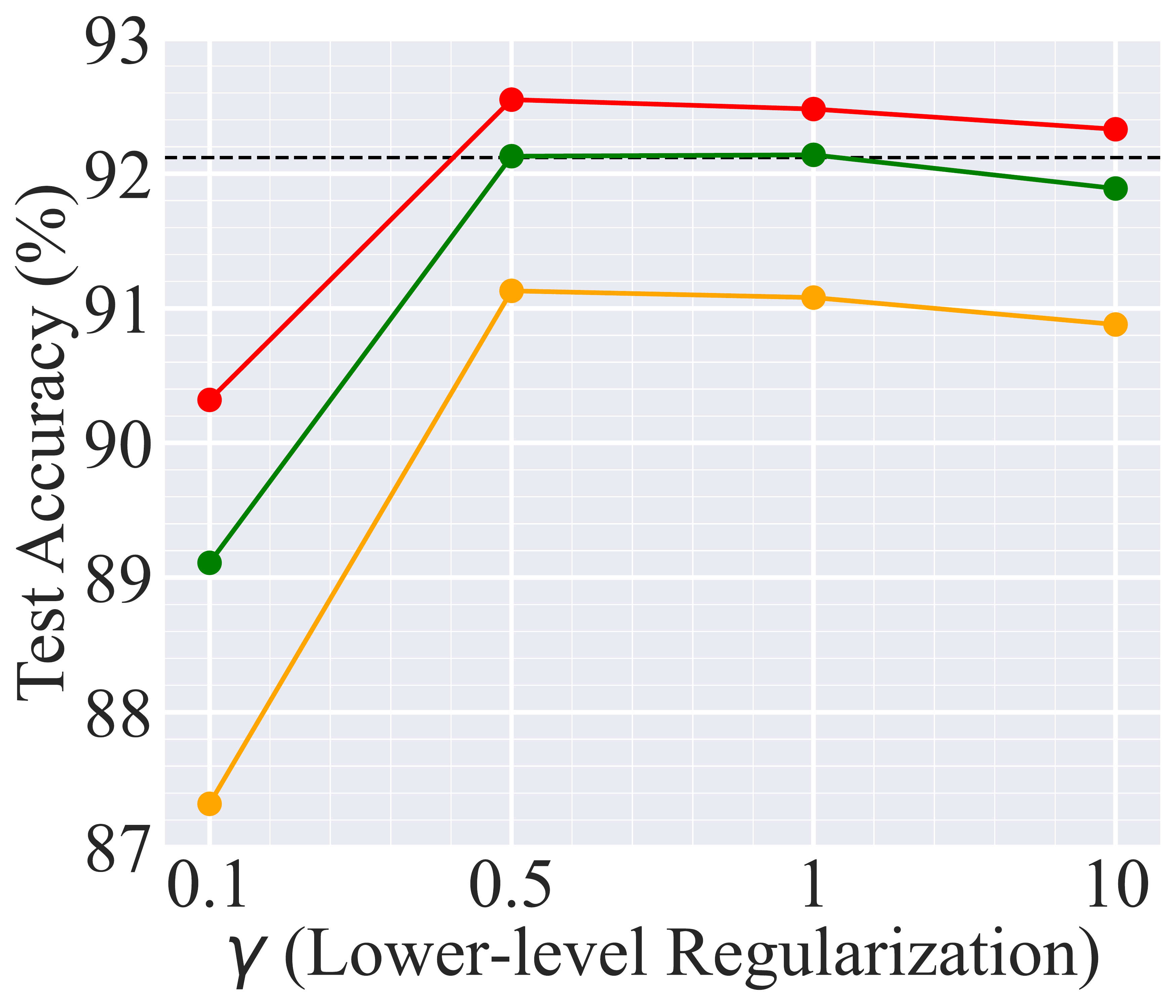}\vspace*{-0.5mm}
    &
    \hspace{-5mm}   \includegraphics[width=.25\textwidth, height=!]{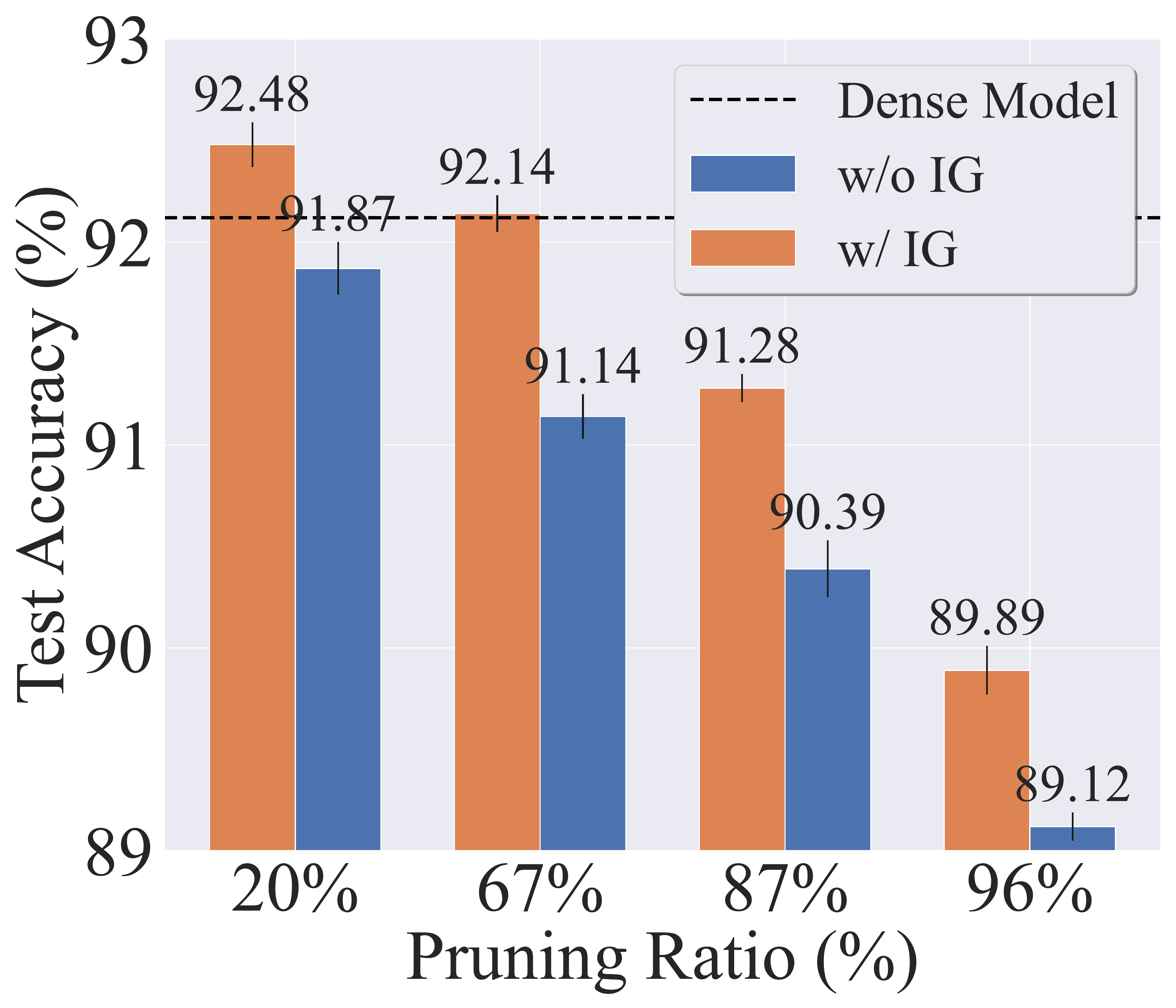}\vspace*{-0.5mm}
    \\
    \footnotesize{(a)} 
     & 
    \footnotesize{(b)} 
     & 
    \footnotesize{(c)} 
\end{tabular}}
\caption{\footnotesize{The sensitivity of {\biprune} to (a) the lower-level step number $N$, (b) the lower-level regularizer $\gamma$, and (c) the contribution of the IG-term at various pruning ratios on (CIFAR-10, ResNet-20). Each curve or column represents a certain sparsity. In (c), we fix the $\gamma$ to 1.0 and compare the performance of the IG-involved/excluded\,\eqref{eq: GD_bi_level} {\biprune}.
}}
\label{fig: ablation_study}
\end{figure}

\begin{figure}[thb]
\centerline{
\begin{tabular}{cccc}
    \includegraphics[width=.35\textwidth,height=!]{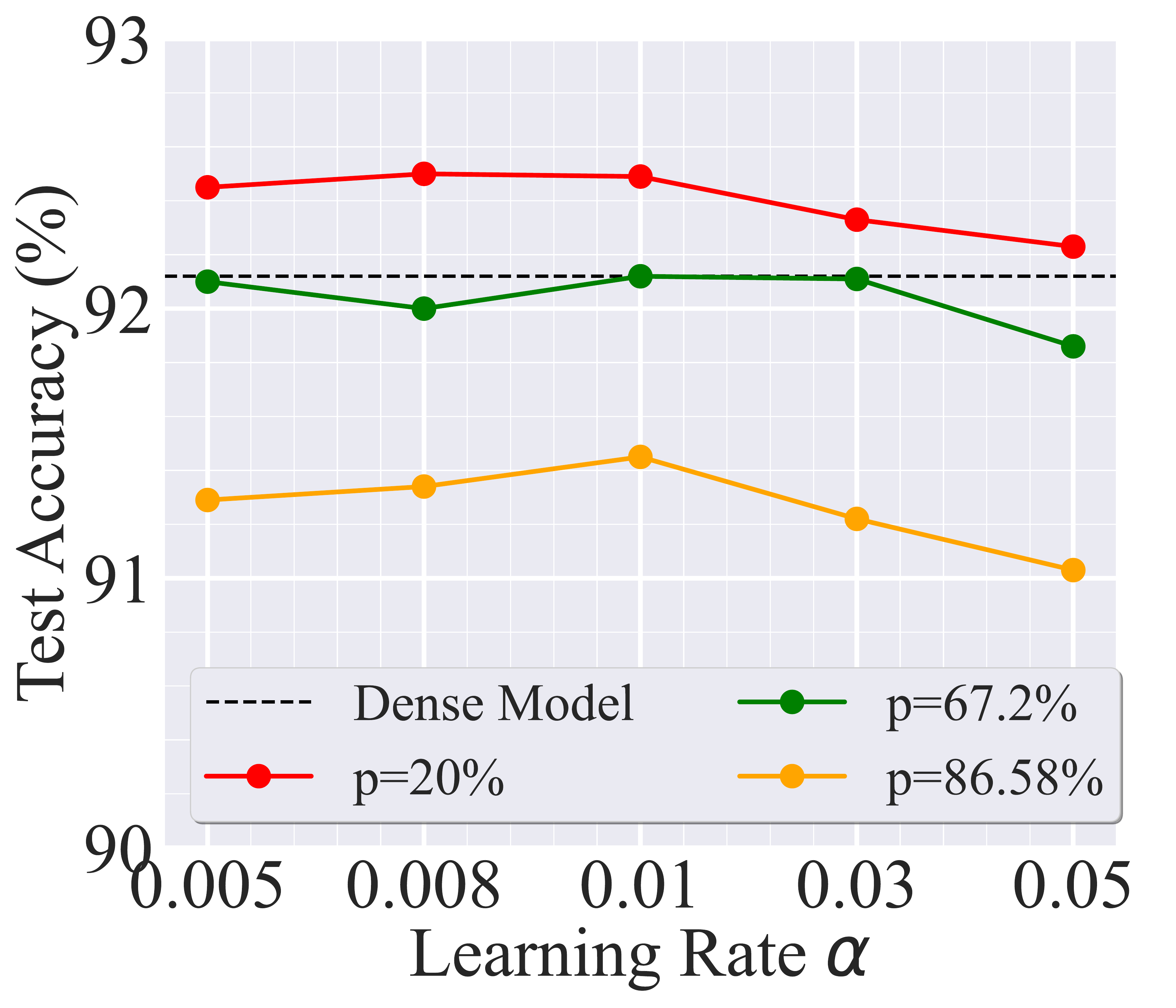} 
    &
    \includegraphics[width=.35\textwidth,height=!]{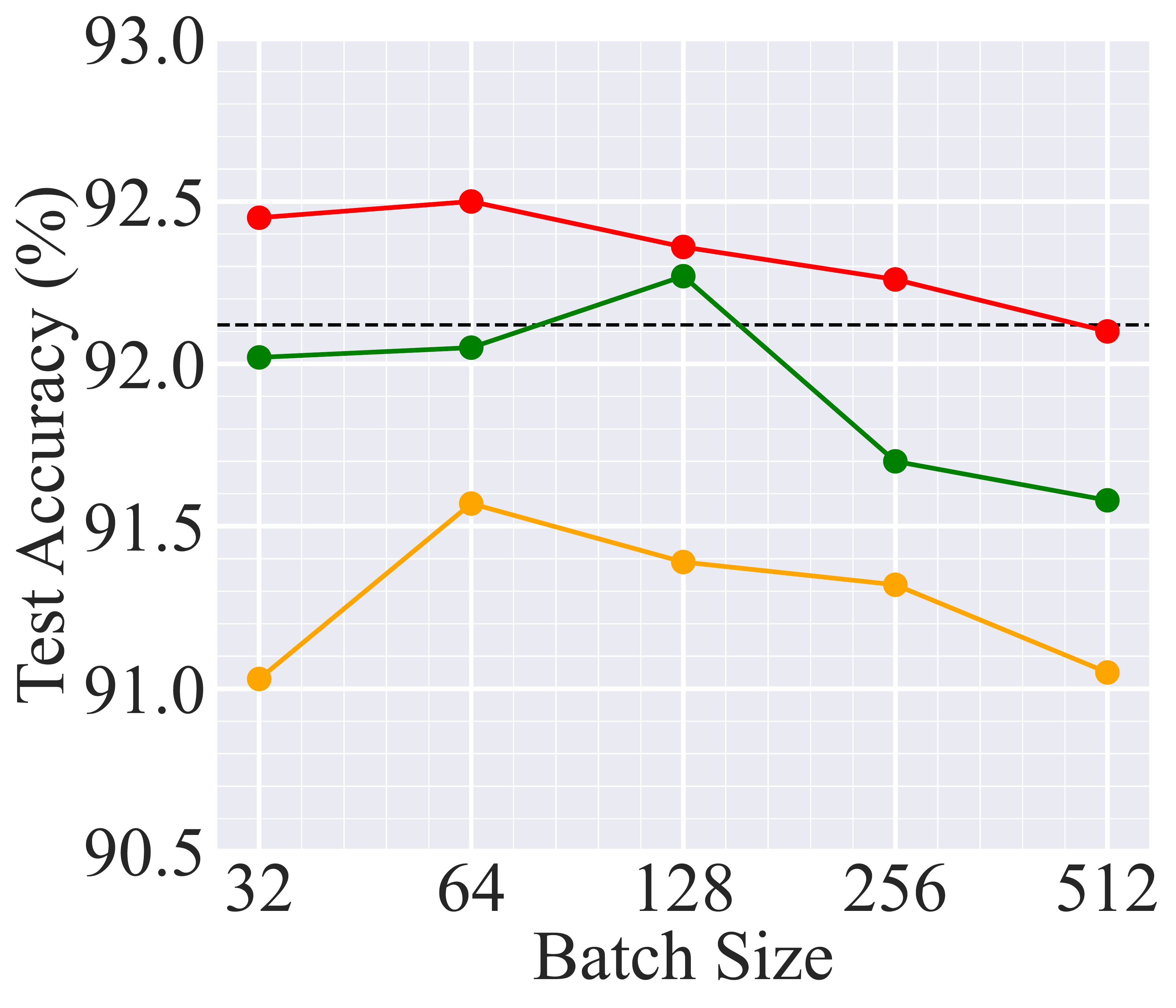} 
    \vspace*{-2mm}\\
    \footnotesize{(a)} 
    & 
    \footnotesize{(b)} 
    
\end{tabular}}
\caption{\footnotesize{ Ablation studies of {\biprune} on different hyper-parameters. All the experiments are based on CIFAR-10 with ResNet-20. We select three sparsity values of a wide range (from not sparse to extreme sparse) to make the results more general. We study the influence of different (a) lower-level learning rate $\alpha$ and (b) batch size.}}
\vspace*{-3mm}
\label{fig: ablation_study_full}
\end{figure}

\paragraph{Sanity check of {\biprune} on specialized hyperparameters.} 

In \underline{Fig.\ref{fig: ablation_study}}, we show  the sensitivity of {\biprune} to   its specialized hyperparameters at lower-level optimization, including 
the number of SGD iterations ($N$) in \eqref{eq: theta_step}, and the regularization parameter $\gamma$ in \eqref{eq: prob_bi_level}.
\underline{Fig.\,\ref{fig: ablation_study}(a)} shows the test accuracy of {\biprune}-pruned  models versus the choice of $N$. As we can see, more SGD iterations for the lower-level optimization do not improve the performance of {\biprune}. This is because in {\biprune}, the \ref{eq: theta_step} is initialized by a pre-trained model which does not ask for aggressive weight updating. The best performance of {\biprune} is achieved at $N \leq 3$. We choose $N = 1$ throughout the experiments as it is computationally lightest. 
\underline{Fig.\,\ref{fig: ablation_study}(b)} shows the performance of {\biprune} by   varying $\gamma$.
As we can see, the choice of $\gamma \in \{ 0.5, 1\}$ yields the best pruning accuracy across all pruning ratios. If $\gamma$ is too small, the lack of strong convexity of lower-level optimization would hamper the convergence. 
If $\gamma$ is too large, the lower-level optimization would depart from the true model objective and causes a performance drop. 
\underline{Fig.\,\ref{fig: ablation_study}(c)} demonstrates the necessity of the IG enhancement in {\biprune}. {We compare {\biprune} with its IG-free version by dropping the IG term in \eqref{eq: GD_bi_level}.  We observe that {\biprune} without IG (marked in blue) leads to a significant performance drop ($> 1\%$) at various sparsities.} This highlights that {\em the IG in the \eqref{eq: m_step} plays a critical role in the performance improvements obtained by {\biprune}}, justifying our novel BLO formulation for the pruning problem.
In Fig.\,\ref{fig: ablation_study_full}, we further demonstrate the influence of different choices of lower-level learning rate $\alpha$ as well as the batch size on the performance of {\biprune}. \underline{Fig.\,\ref{fig: ablation_study_full} (a)} shows that the test accuracy of {\biprune}-pruned models is not quite sensitive to  the choice of $\alpha \in \{0.01, 0.008\}$. A large $\alpha$ value (\emph{e.g.}, $\alpha > 0.05$) will slightly decrease the performance of {\biprune}. By contrast, a small $\alpha$ is preferred due to the fact that the model parameters are updated based on the   pre-trained values. \underline{Fig.\,\ref{fig: ablation_study_full} (b)} shows how the batch size   influences the performance. As we can see, a large batch size might hurt the stochasticity of the algorithm and thus degrades the performance. We list our detailed batch size choices for different datasets in Tab.\,\ref{table: impl_details}.


\paragraph{Additional structured pruning experiments.}

\begin{figure}[thb]
\centerline{
\begin{tabular}{cc}
    \includegraphics[width=.4\textwidth,height=!]{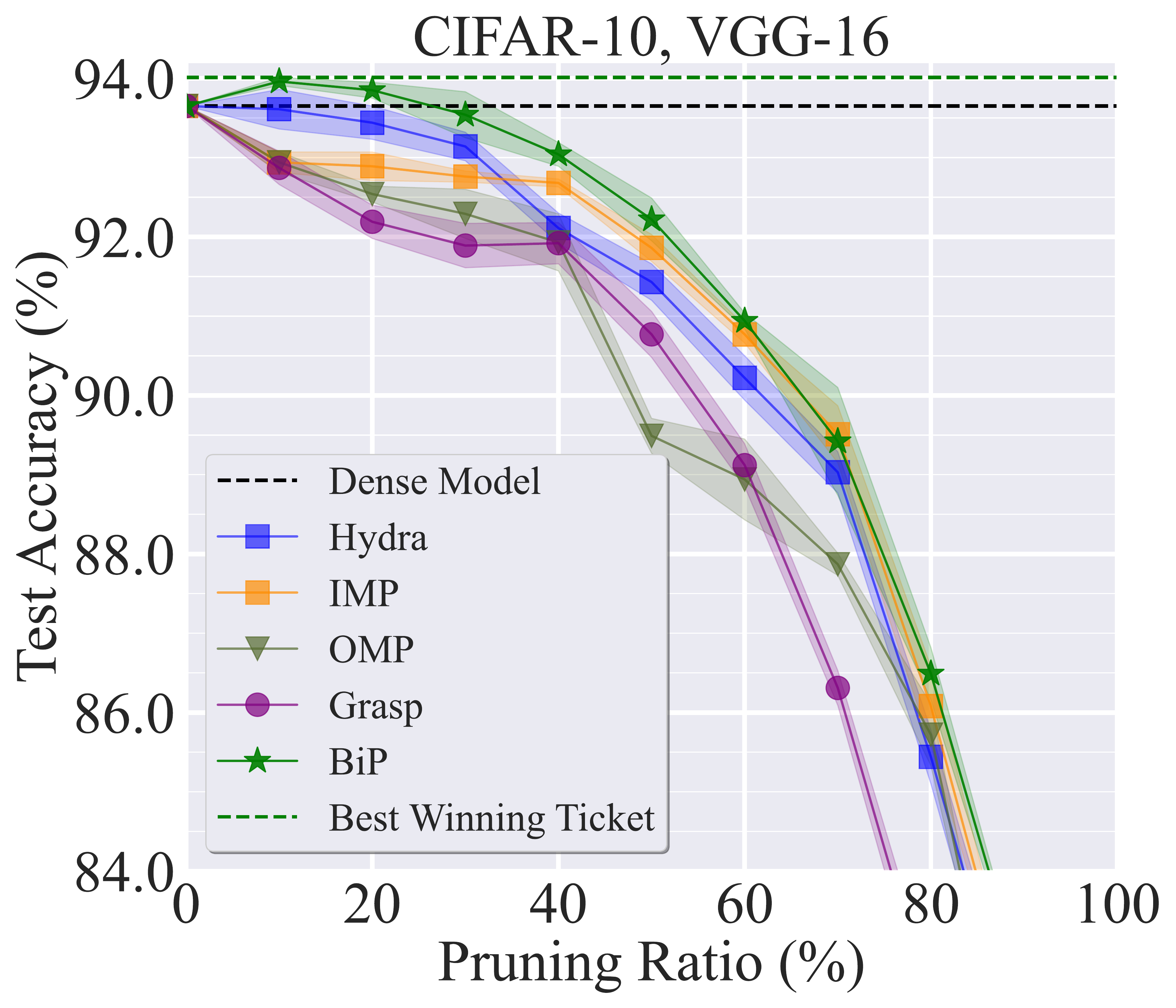} 
    
    \includegraphics[width=.4\textwidth,height=!]{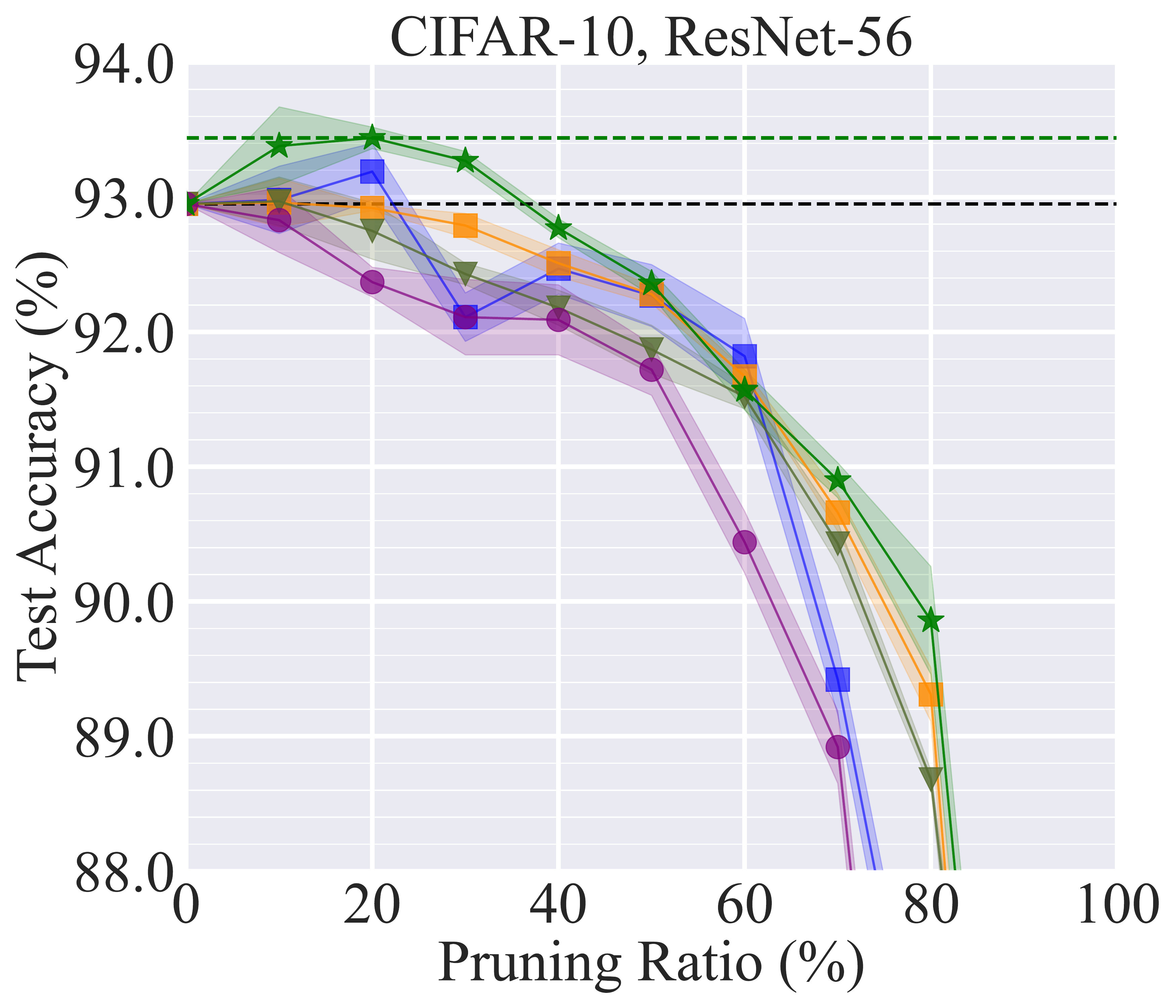} 
\end{tabular}}
\caption{\footnotesize{Filter-wise pruning test accuracy (\%) v.s. sparsity (\%) on CIFAR-10 with VGG-16 and ResNet-56.
    }}
\vspace*{-3mm}
\label{fig: structured_performance_cifar_appendix}
\end{figure}

\begin{figure}[htb]
\centerline{
\begin{tabular}{cccc}
    \hspace*{-2mm} \includegraphics[width=.3\textwidth,height=!]{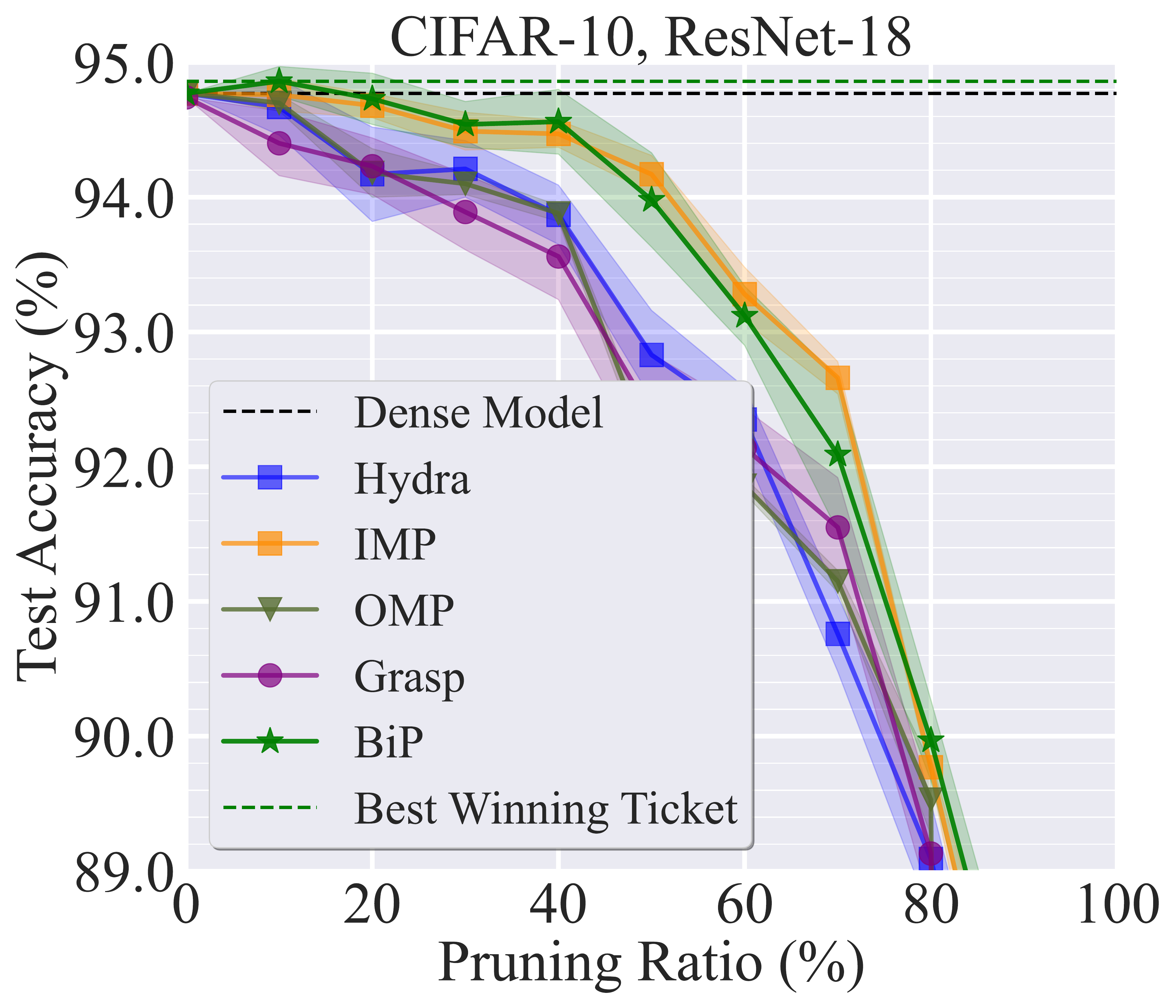} &
    \hspace*{-5mm}  \includegraphics[width=.3\textwidth,height=!]{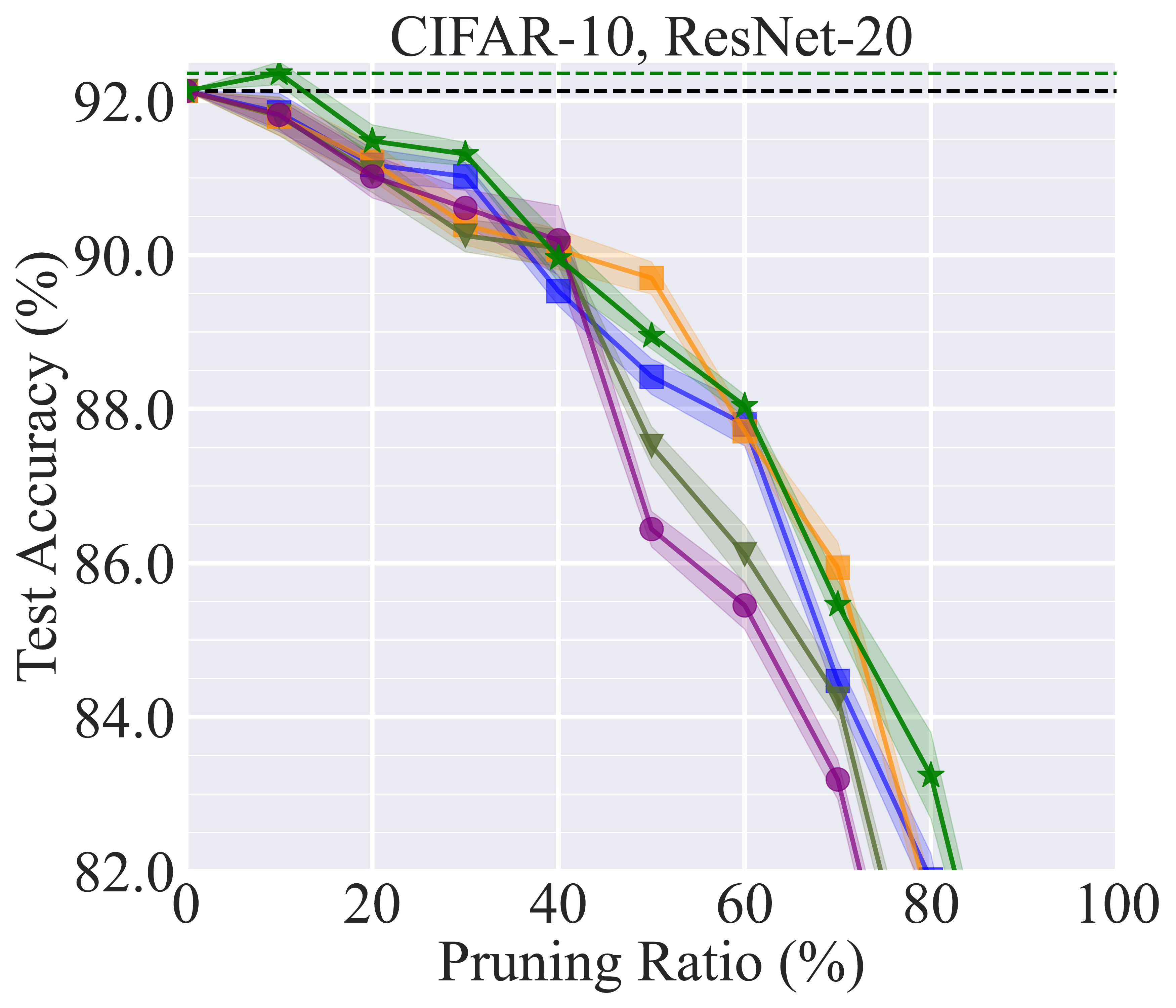} &
    \hspace*{-5mm}  \includegraphics[width=.3\textwidth,height=!]{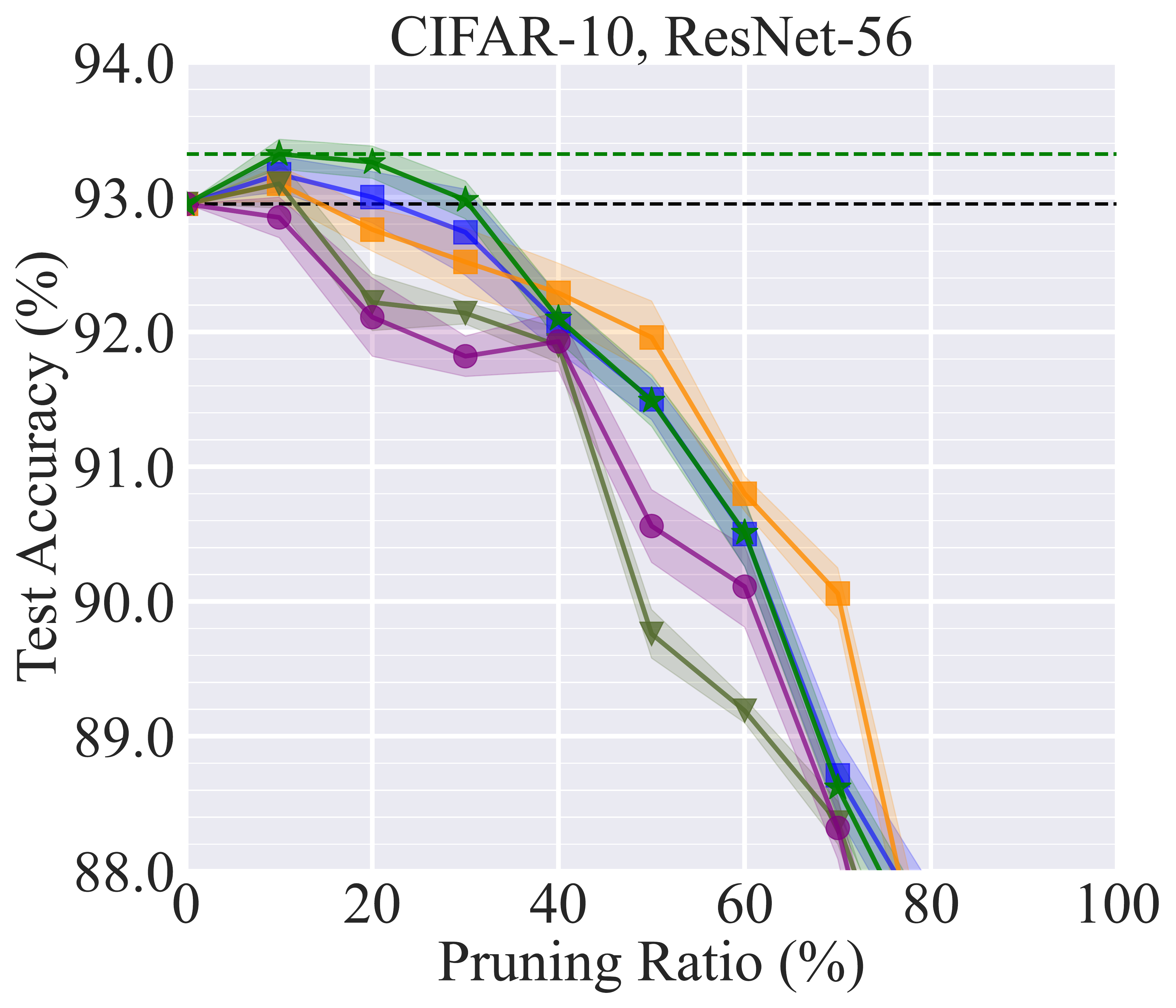} \\
    
    \hspace*{-2mm} \includegraphics[width=.3\textwidth,height=!]{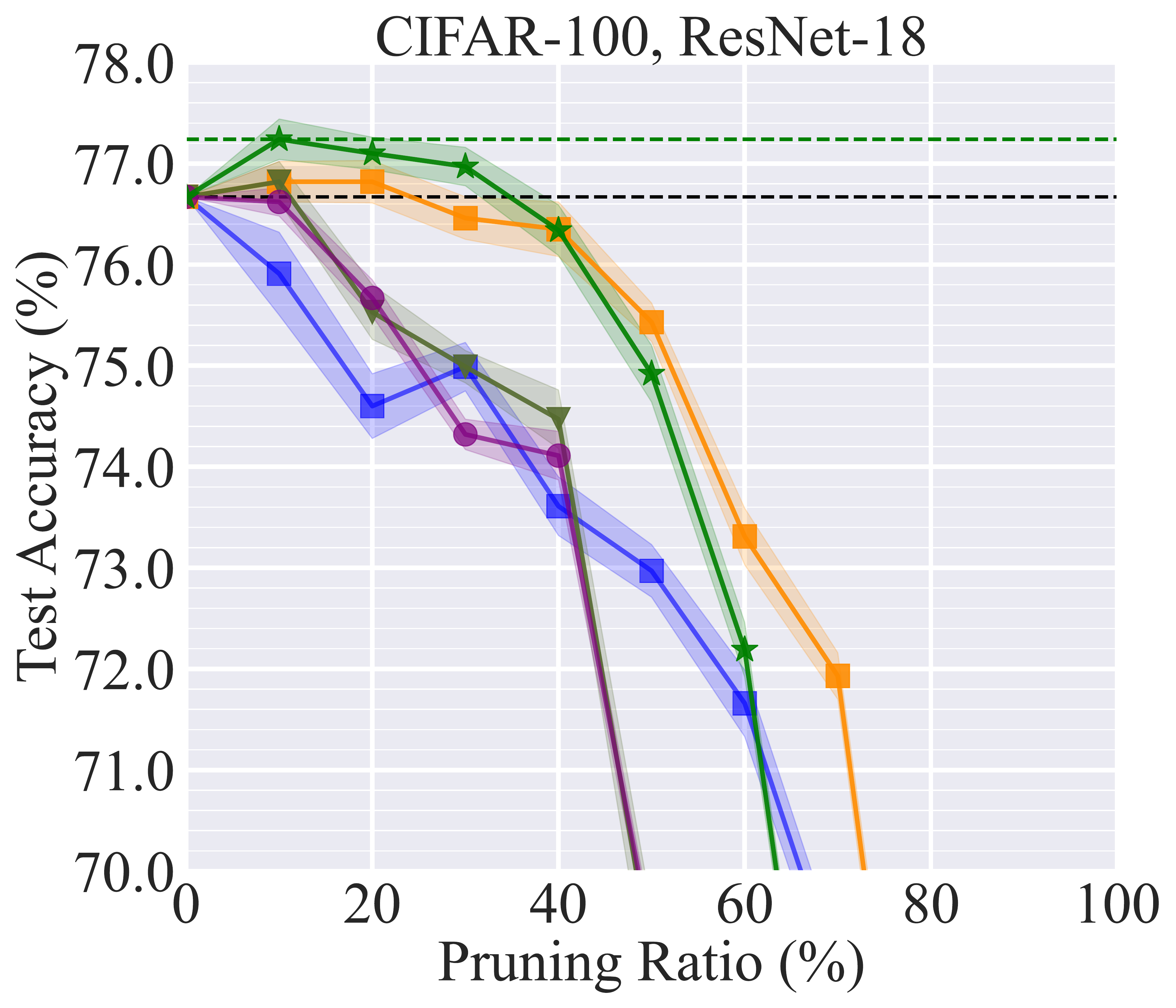} &
    \hspace*{-5mm}  \includegraphics[width=.3\textwidth,height=!]{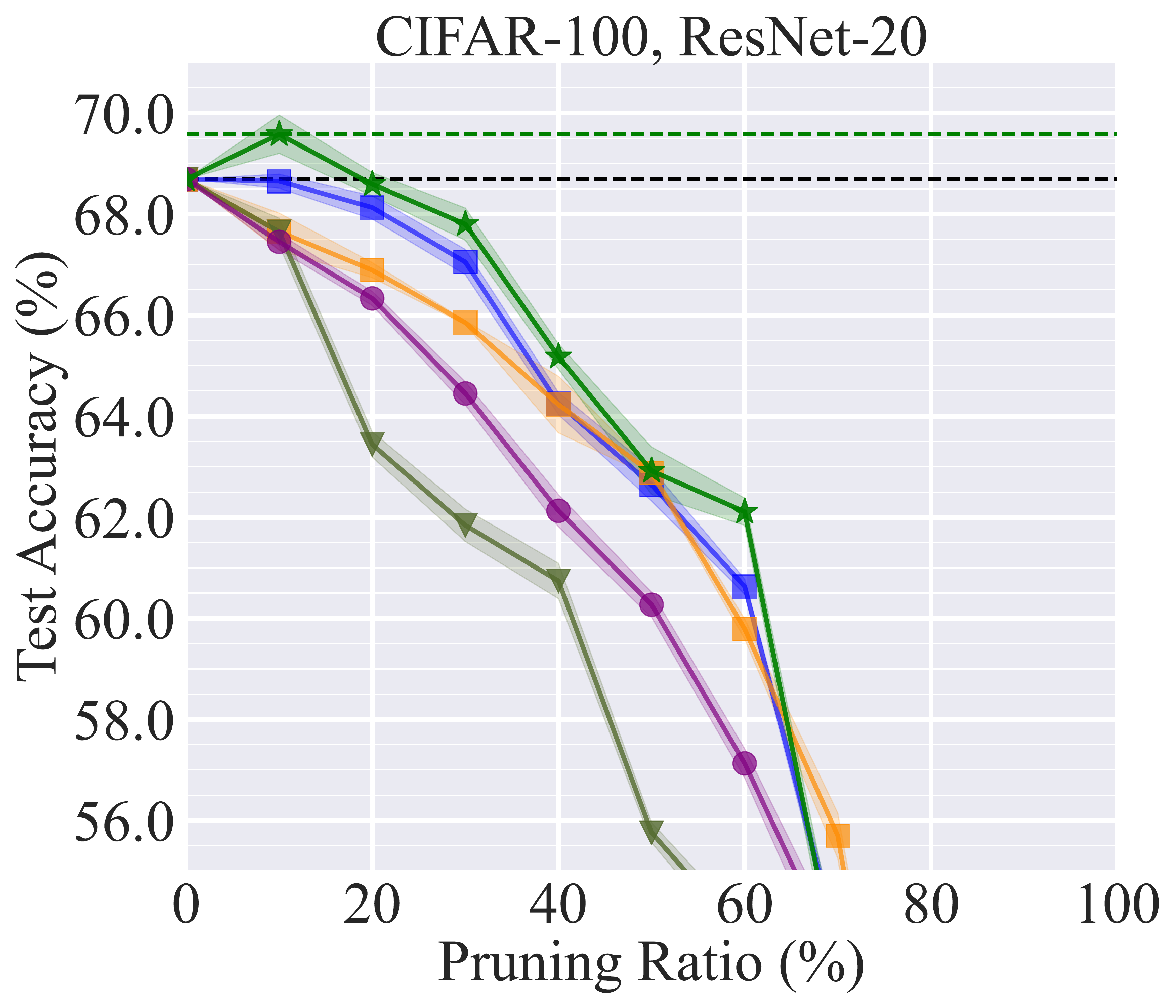} &
    \hspace*{-5mm}  \includegraphics[width=.3\textwidth,height=!]{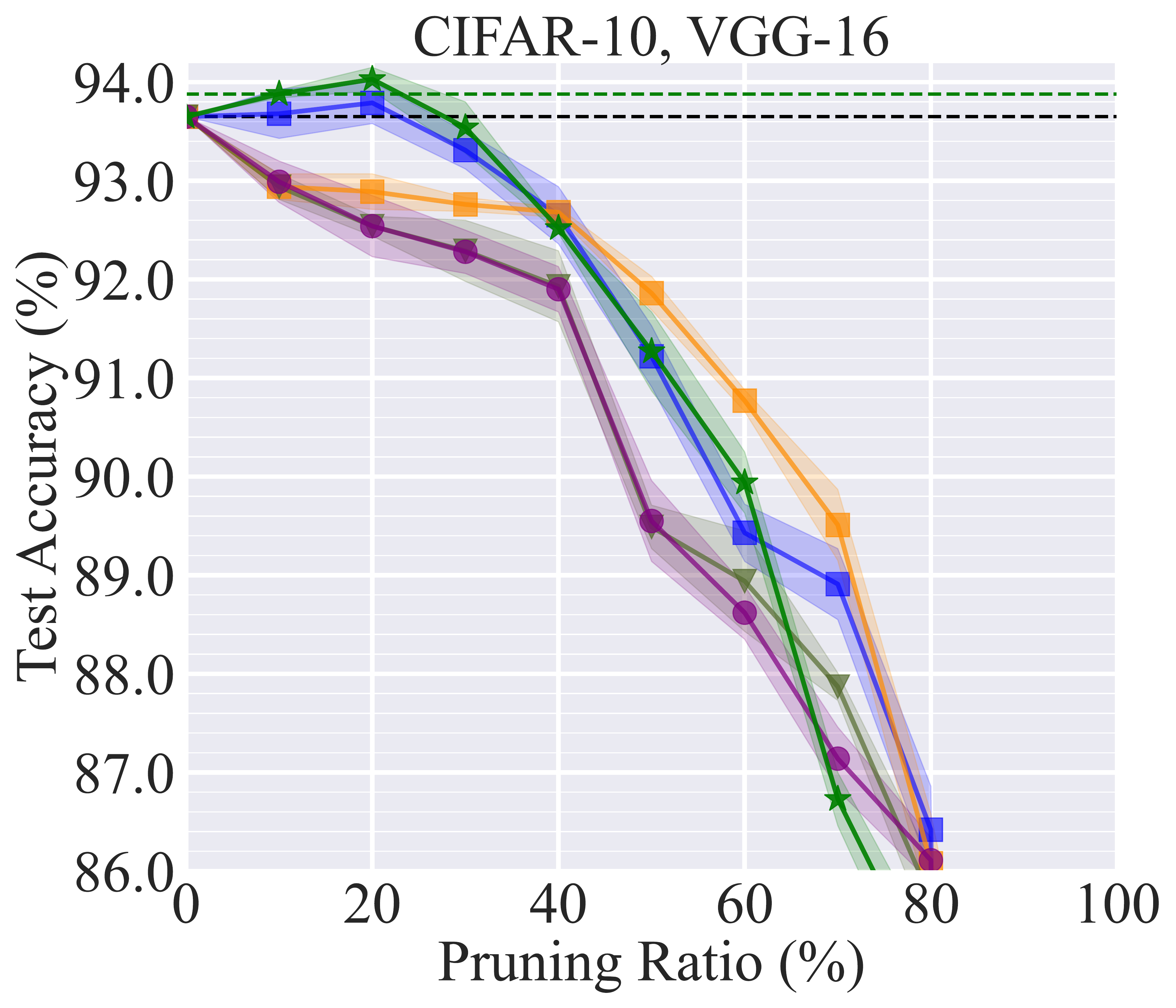} 
\end{tabular}}
\caption{\footnotesize{Channel-wise pruning test accuracy (\%) v.s. sparsity (\%).
    Settings are consistent with Fig.\,\ref{fig: structured_performance_cifar_appendix}.
    }}
\label{fig: channel_performance_cifar_appendix}
\end{figure}

In addition to   filter pruning   in Fig.\,\ref{fig: structured_performance_cifar}, we provide more results in the structured pruning setting, including both filter-wise   and channel-wise pruning (as illustrated in Fig.\,\ref{fig: illustration}). In Fig.\,\ref{fig: structured_performance_cifar_appendix}, results on CIFAR-10 with VGG-16 and ResNet-56 are added as new experiments compared to Fig.\,\ref{fig: structured_performance_cifar}. Fig.\,\ref{fig: channel_performance_cifar_appendix} shows the results of the channel-wise pruning. As we can see, consistent with Fig.\,\ref{fig: structured_performance_cifar},   {\biprune} is able to find the best winning tickets throughout the experiments while it is difficult for IMP to find winning tickets in most cases. We also notice that {\hydra}, as the optimization-based baseline, serves as a strong baseline in   filter-wise pruning. It also indicates the superiority of the optimization-based methods over the heuristics-based ones in dealing with more challenging pruning settings.

\begin{figure}
\centerline{
\includegraphics[width=.39\textwidth,height=!]{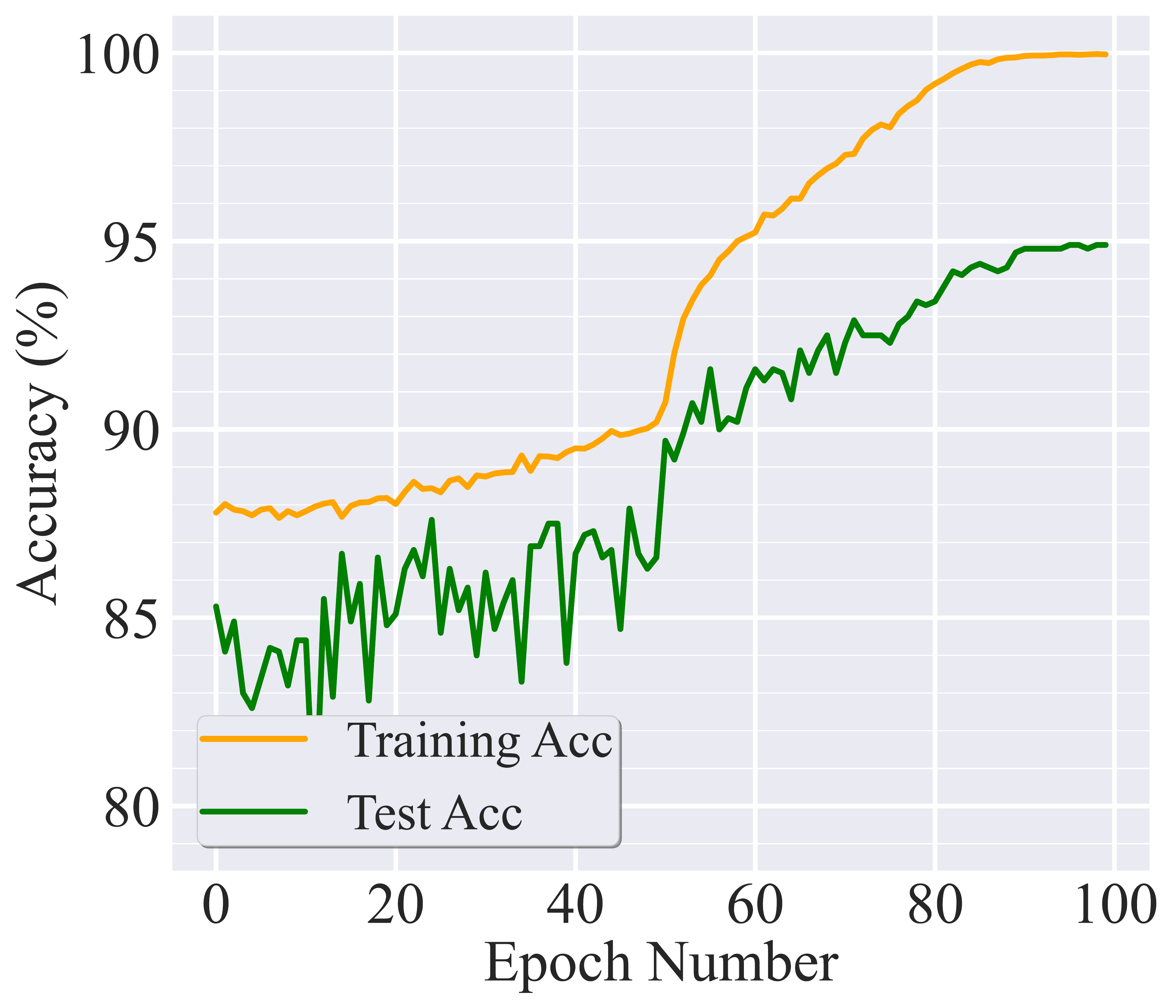}
}
\caption{\footnotesize{The training trajectory of {\biprune} for  unstructured pruning on (CIFAR-10, ResNet-18) with a pruning ratio of $p=80\%$. 
}}
\label{fig: convergence}

\end{figure}

\paragraph{Training trajectory of {\biprune}.}
We show in Fig.\,\ref{fig: convergence} that the {\biprune}   algorithm converges quite   well within 100 training epochs using a cosine learning rate scheduler.

\begin{figure}
\centerline{
\includegraphics[width=.39\textwidth,height=!]{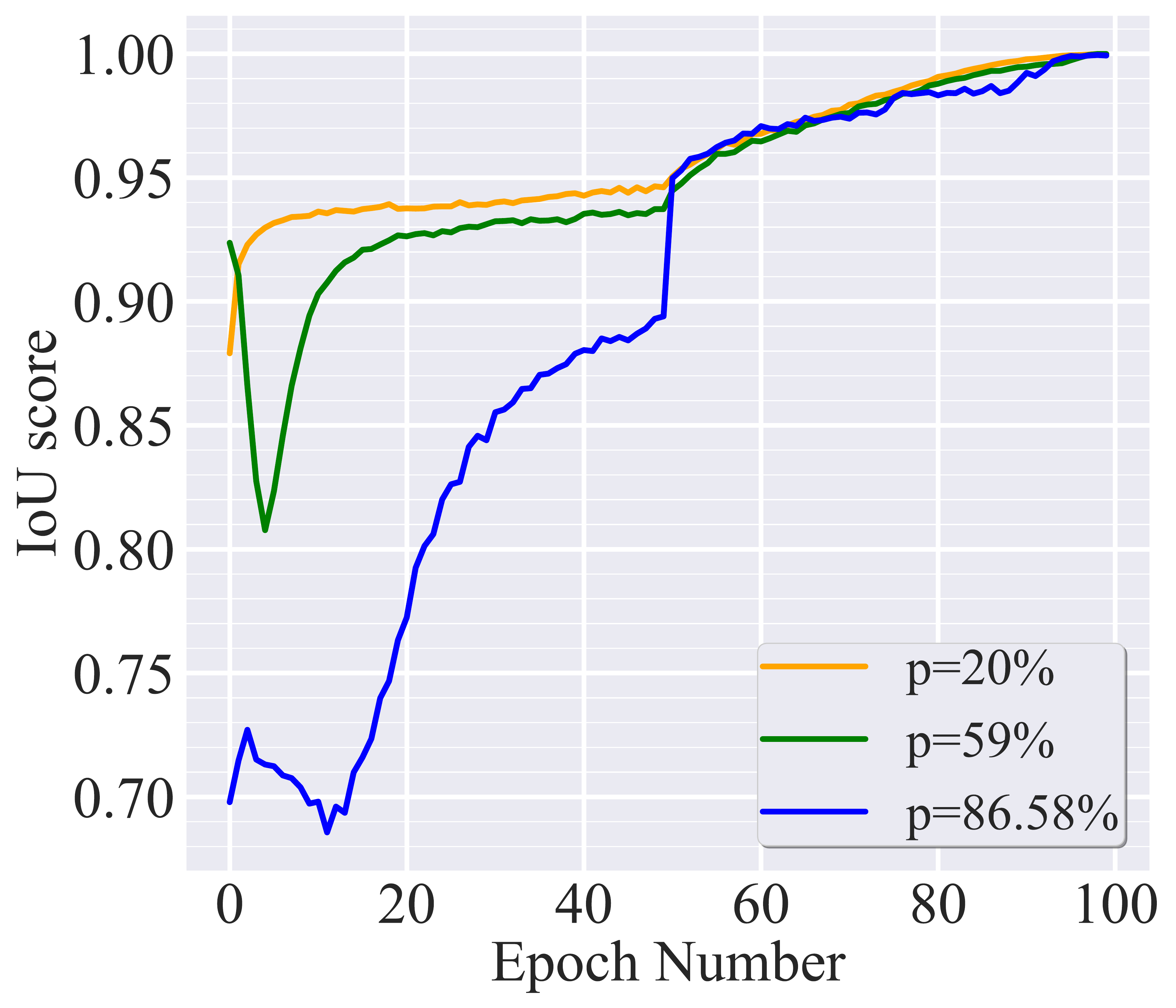}
}
\caption{\footnotesize{Training trajectory of the IoU (intersection over union) score between the masks of two adjacent epochs. We show the trajectory of different pruning ratios.}}
\label{fig: IoU_Trajectory_rebuttal}
\end{figure}

\paragraph{\RV{The training trajectory of the mask IoU score.}} To verify the argument that the mask also converges at the end of the training, we show the training trajectory of the mask similarity between two adjacent-epoch models in Fig.\,\ref{fig: IoU_Trajectory_rebuttal} at different pruning ratios. Here the mask similarity is represented through the intersection of the union (IoU) score of the two masks found by two adjacent epochs. The IoU score ranges from 0.0 to 1.0, and a higher IoU implies a larger similarity between the two masks. As we can see, the IoU score converges to 1.0 in the end, which denotes that the mask also converges at the end of the training phase. Also, with a smaller pruning ratio, the mask turns to converge more quickly.

\paragraph{\RV{The effect of different lower-level steps in {\biprune} on the training dynamics.}} 
\begin{figure}[htb]
\centerline{
\begin{tabular}{cc}
     \includegraphics[width=.4\textwidth,height=!]{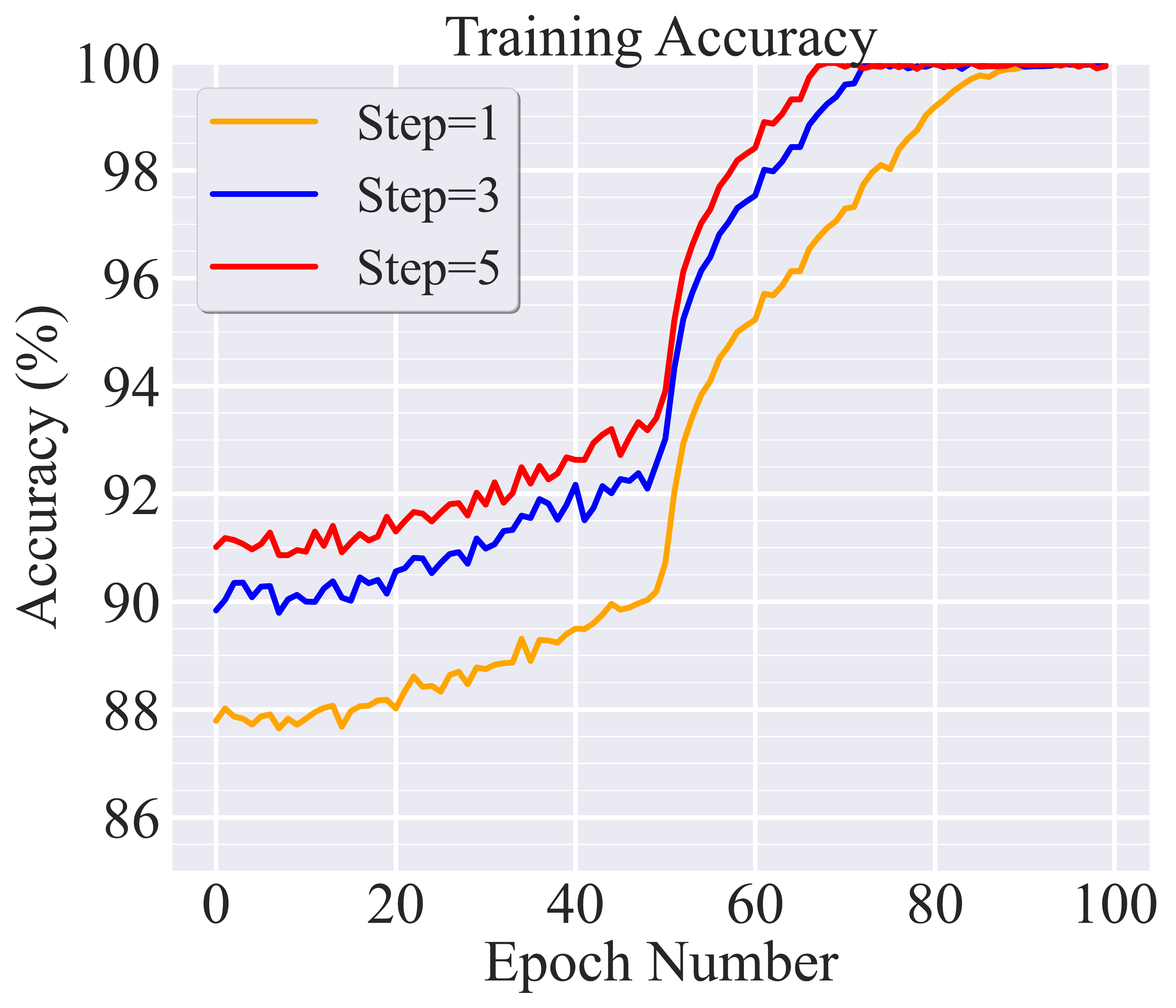} &
     \includegraphics[width=.4\textwidth,height=!]{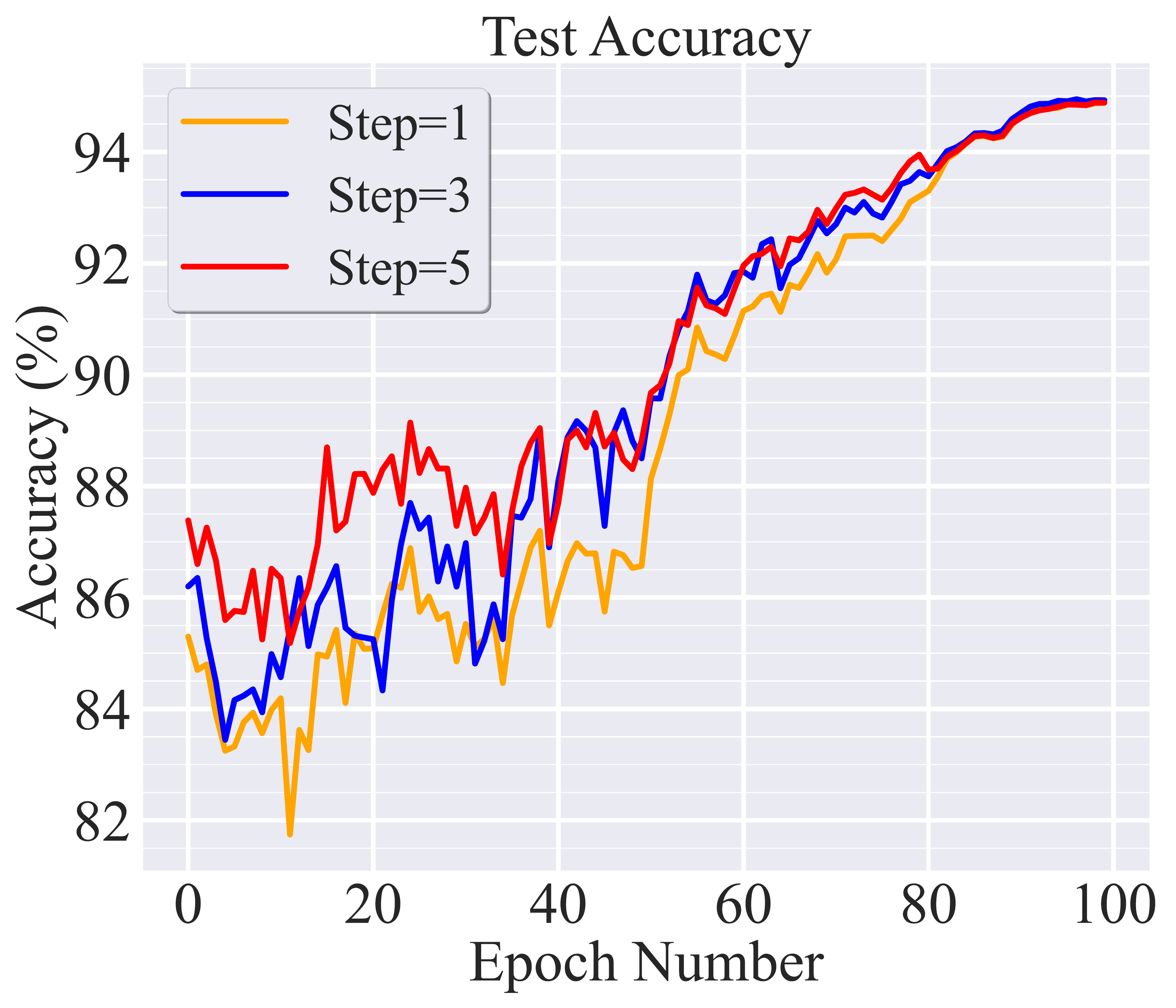}
\end{tabular}}
\caption{\footnotesize{Training dynamics of {\biprune} with different lower-level (SGD) steps on (CIFAR-10, ResNet-18) with the pruning ratio of p=80\%. }}
\label{fig: step_convergence_rebuttal}
\end{figure}
We conduct additional experiments to demonstrate the effectiveness of using one-step SGD in {\biprune}. In our new experiments, we consider the number of SGD steps, 1, 3, and 5. We report the training trajectories of {\biprune} in Fig.\,\ref{fig: step_convergence_rebuttal}. As we can see, the use of multi-step SGD accelerates model pruning convergence at its early phase. Yet, if we run {\biprune} for a sufficient number of epochs (we used 100 by default in other experiments), the final test accuracy of using different SGD settings shows little difference. Although the use of multiple SGD steps could improve the convergence speed, it introduces extra computation complexity per BLO step. Thus, from the overall computation complexity perspective, using 1 SGD step but running more epochs is advantageous in practice.

\paragraph{\RV{The effect of larger training epoch numbers with extreme sparsities.}} 
\begin{figure}[htb]
\centerline{
\begin{tabular}{ccc}
    \hspace*{-2mm} \includegraphics[width=.3\textwidth,height=!]{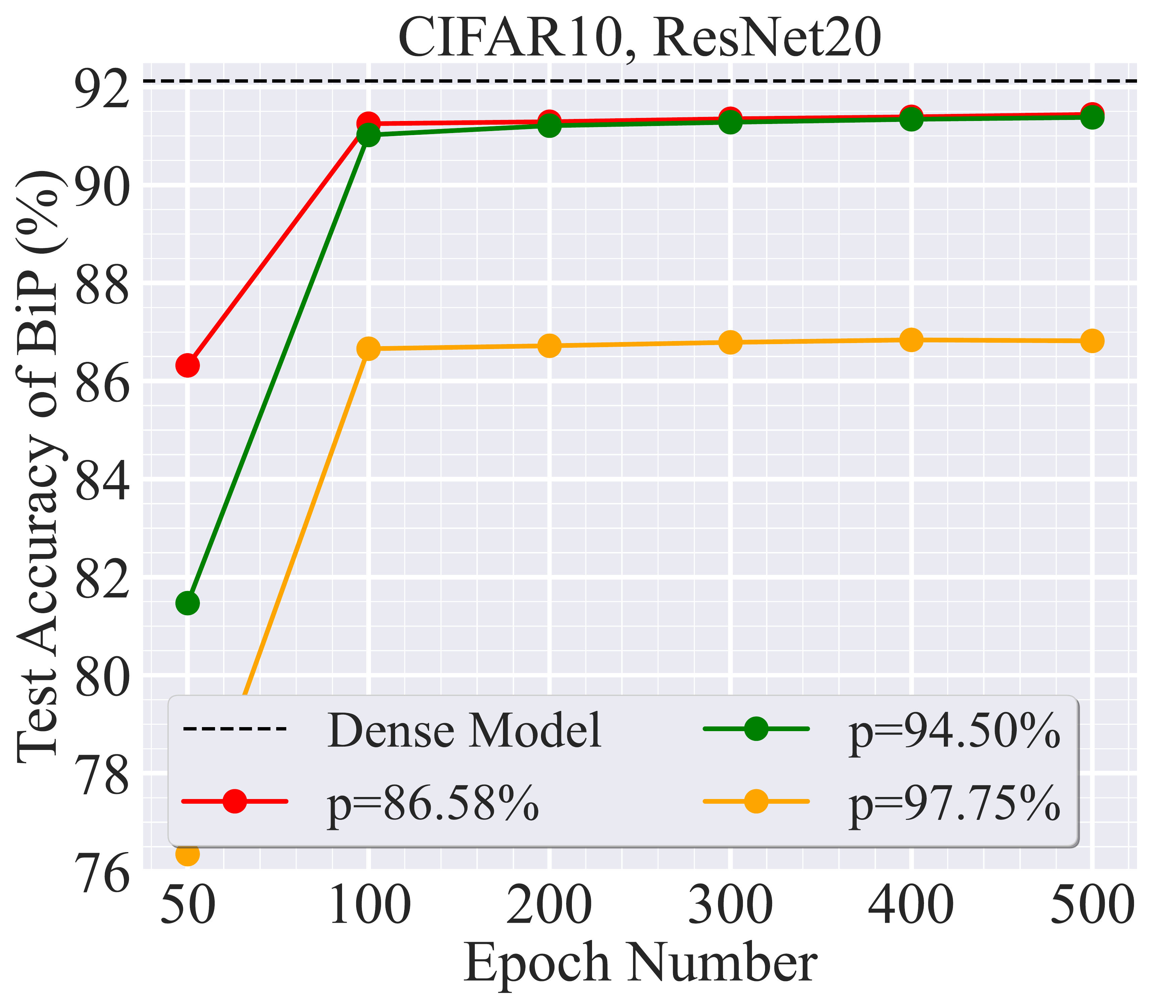} &
    \hspace*{-5mm}  \includegraphics[width=.3\textwidth,height=!]{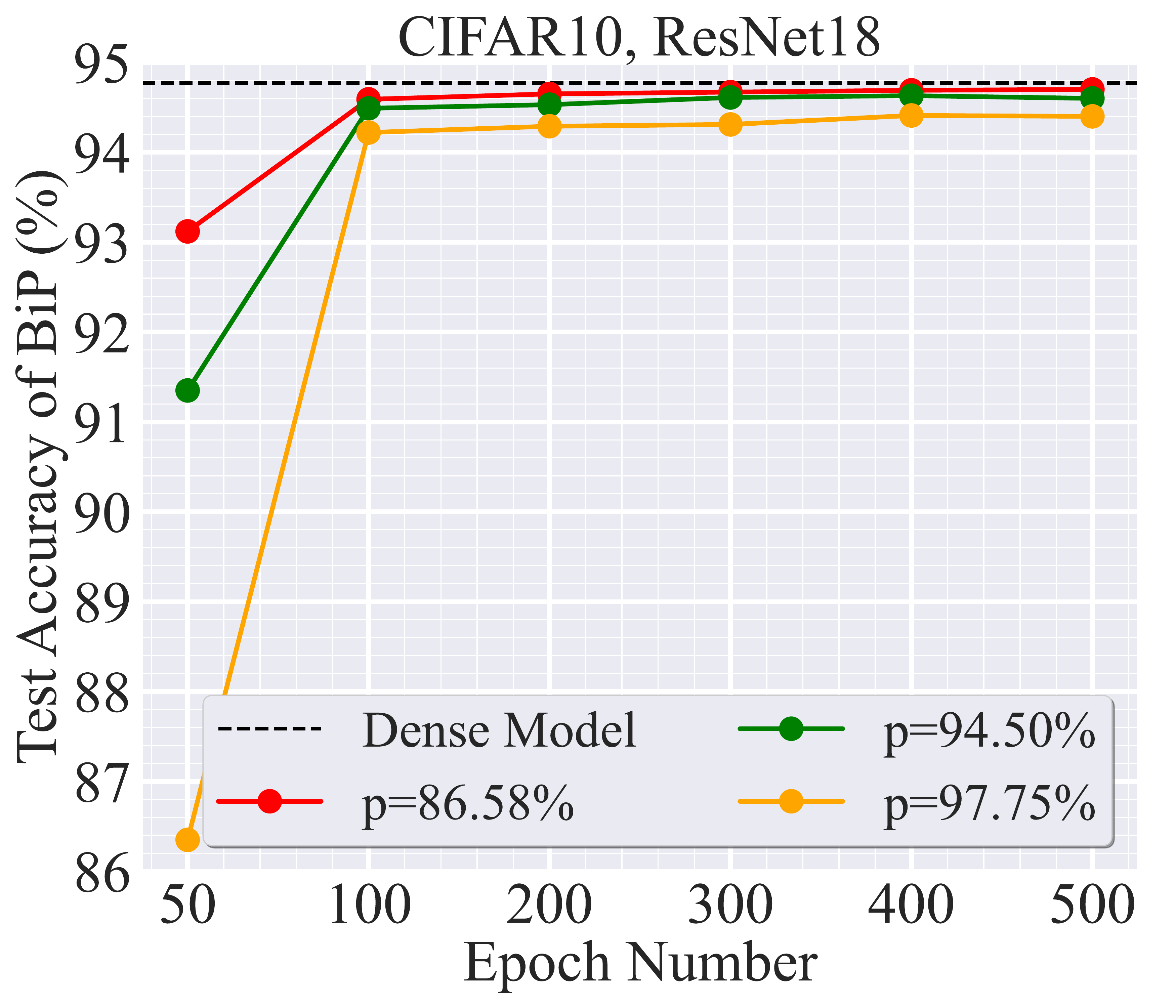} &
    \hspace*{-5mm}  \includegraphics[width=.3\textwidth,height=!]{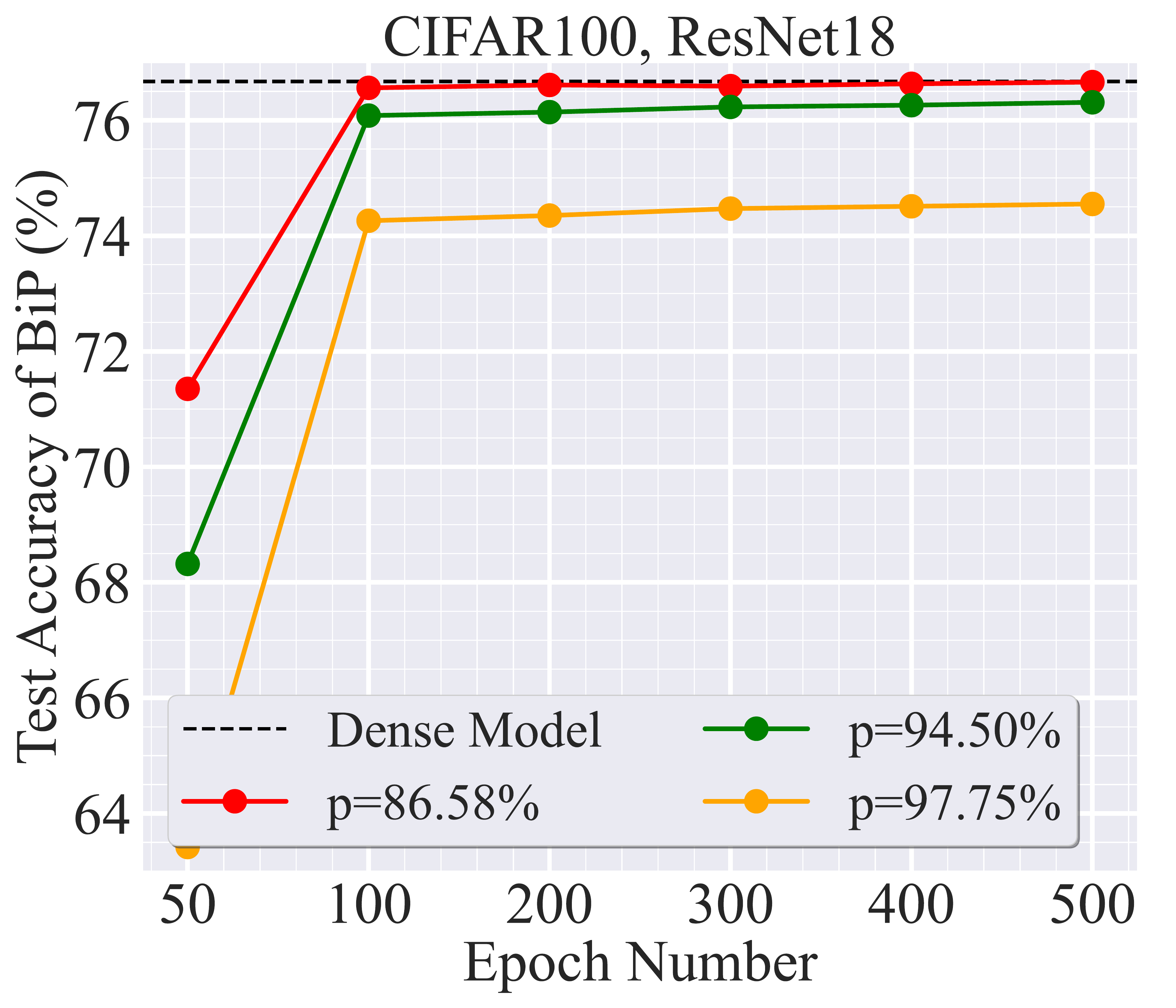}
\end{tabular}}
\caption{\footnotesize{The effect of total training epoch number on the test accuracy with large pruning ratios. The epoch number is by default set to 100 in this paper. In each sub-figure, we report the performance of {\biprune} with three different sparse ratios $p$.}}
\label{fig: epoch_extreme_sparsity_rebuttal}
\end{figure}
We allow more time (training epochs) for {\biprune} when a higher pruning rate is considered and the results are shown in Fig.\,\ref{fig: epoch_extreme_sparsity_rebuttal}. Specifically, we test three datasets and consider three pruning ratios (p=86.58\%, 94.50\%, 97.75\%). For each pruning ratio, we examine the test accuracy of {\biprune} versus the training epoch number from 50 to 500. Note that the number of training epochs in our original experiment setup was set to 100. As we can see, the performance of {\biprune} gets saturated when the epoch number is over 100. Thus, even for a higher pruning ratio, the increase of training epoch number over 100 does not gain much improvement in accuracy.

\begin{figure}[htb]
\centerline{
\begin{tabular}{cc}
     \includegraphics[width=.49\textwidth,height=!]{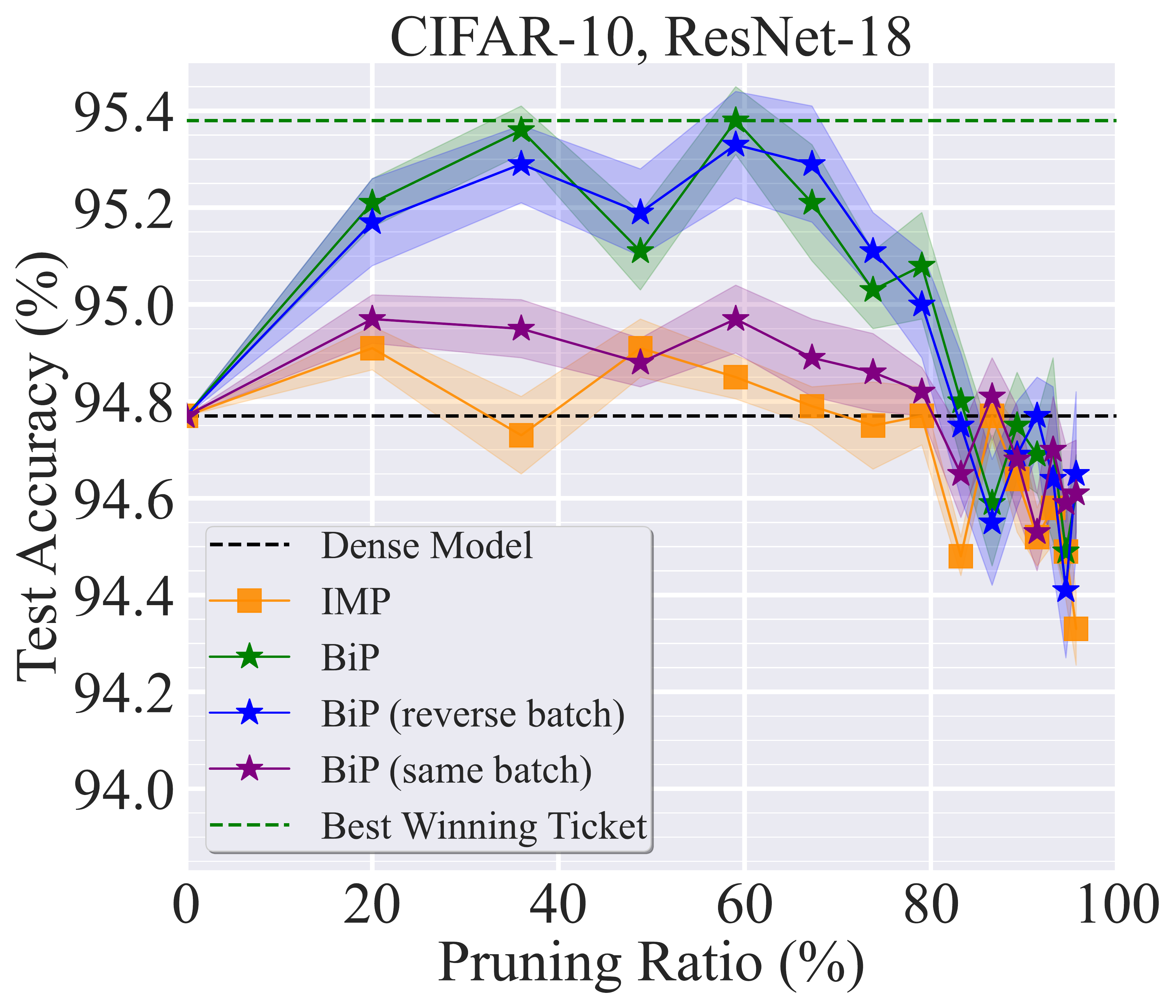} &
     \includegraphics[width=.49\textwidth,height=!]{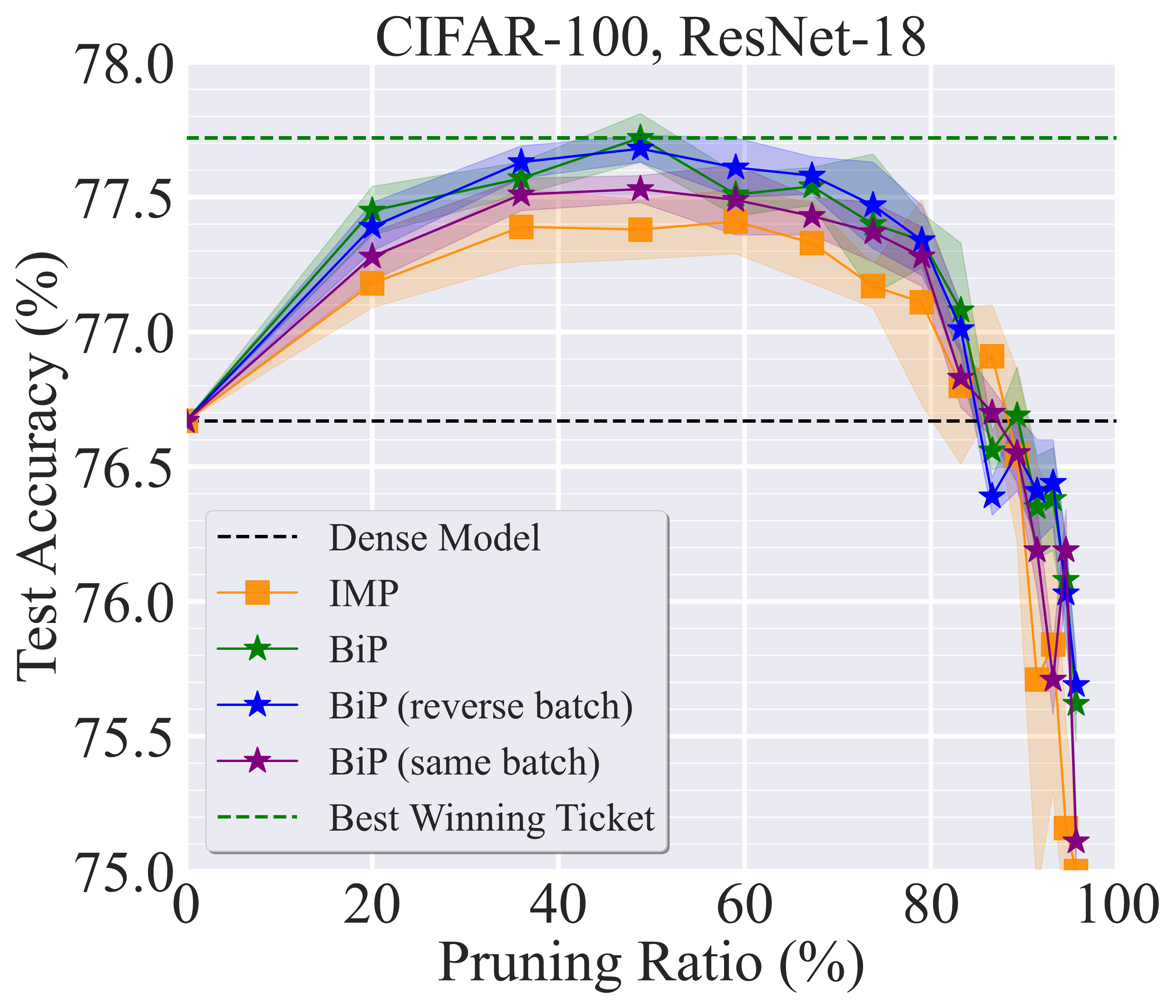}
\end{tabular}}
\caption{\footnotesize{The effect of different training batch schemes on the performance of {\biprune}. We consider two different variants of {\biprune} denoted as \textsc{BiP} (reverse batch) and \textsc{BiP} (same batch). For \textsc{BiP} (reverse batch), the data batches are fed into the upper- and lower-step in a reversed order within each epoch, while for \textsc{BiP} (same batch), the data batches for upper- and lower-level are always the same. Experiment settings are consistent with Fig.\,\ref{fig: unstructured_performance_cifar}}. For better readability, we only plot the strongest baseline IMP for comparison.}
\label{fig: batch_diverse_rebuttal}
\end{figure}

\begin{figure}[htb]
\centerline{
\begin{tabular}{cc}
     \includegraphics[width=.49\textwidth,height=!]{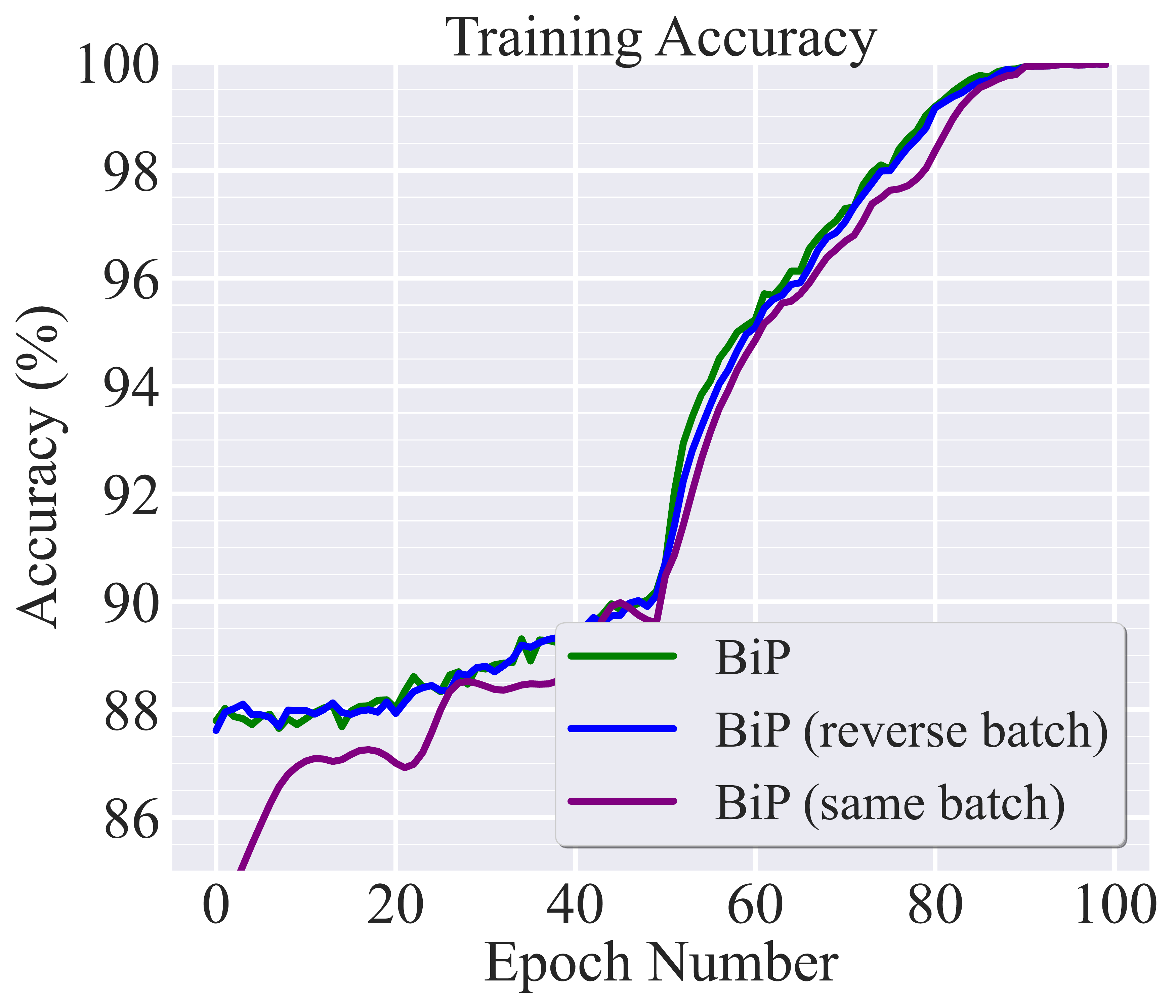} &
     \includegraphics[width=.49\textwidth,height=!]{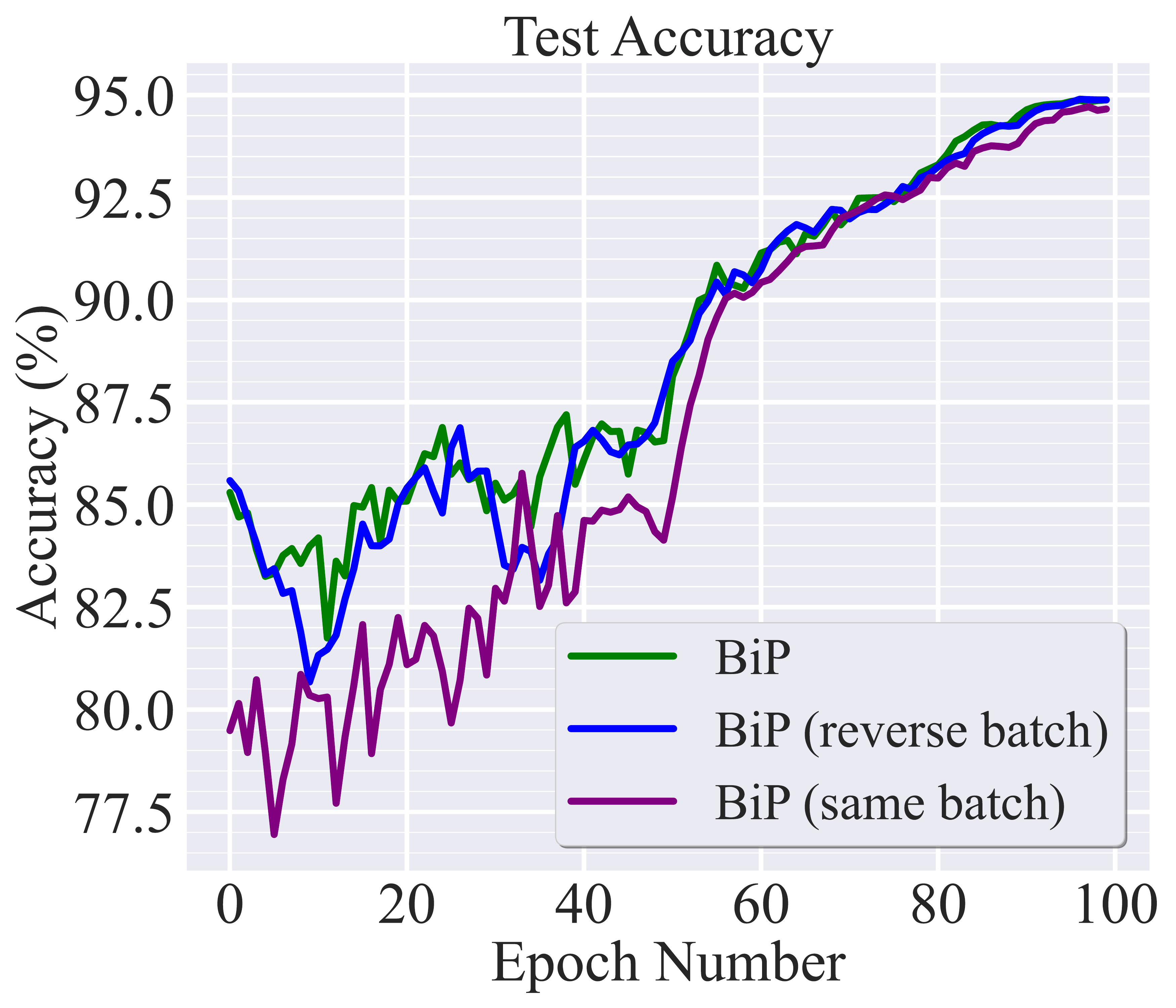}
\end{tabular}}
\caption{\footnotesize{The effect of different training batch schemes on the training dynamics of {\biprune}. We plot the training dynamics of different variants of {\biprune} on (CIFAR-10, ResNet-18) with the pruning ratio of 80\%.}}
\label{fig: batch_diverse_convergence_rebuttal}
\end{figure}

\paragraph{\RV{The effect of different training batch schemes.}} 
\RV{We conducted ablation studies on three different schemes of {\biprune}'s training batches for the upper and lower level. In addition to two different random batches for the two levels, we also consider the same batch and the reverse batch scheme. {\biprune} (same batch) always uses the same data batches for the two levels in each iteration while {\biprune} (reverse batch) uses the data batches in a reversed order for the two level. Fig.\,\ref{fig: batch_diverse_rebuttal} shows that both the random batch scheme (i.e., {\biprune}) and the reverse batch scheme can bring a better testing accuracy performance than the same batch scheme throughout different pruning ratio settings. Fig.\,\ref{fig: batch_diverse_convergence_rebuttal} shows that {\biprune} the same batch scheme converges slower compared to the other two. Both of the results indicate {\biprune} benefits from the diverse batch selection.}

\begin{table}[htb]
\centering
\caption{\footnotesize{The sparsest winning tickets found by different methods on Tiny-ImageNet and ImageNet datasets. Winning tickets refer to the sparse models with an average test accuracy no less than the dense model\,\cite{chen2020lottery}. In each cell, $p\%$ (acc$\pm$std\%) represents the sparsity as well as the test accuracy. The test accuracy of dense models can be found in the header. {\xmark} signifies that no winning ticket is found  by a   pruning method. Given the data-model setup (\textit{i.e.}, per column),
the sparsest winning ticket is highlighted in \textbf{bold}.
}}
\label{tab: sparsest_wt_rebuttal}
\resizebox{0.6\columnwidth}{!}{%
\begin{tabular}{c|c|c|c}
\toprule[1pt]
\midrule
\multirow{3}{*}{Method} & \multicolumn{1}{c}{Tiny-ImageNet}                               & \multicolumn{2}{|c}{ImageNet} \\
                        & ResNet-18                        & ResNet-18                 & ResNet-50      \\ 
                        & (63.83\%)                     & (70.89\%)                 &   (75.85\%)      \\\midrule
IMP                     &  20\% (64.17$\pm$0.11\%)       & 74\% (71.15$\pm$0.19\%)   & \textbf{80}\% (76.05$\pm$0.13\%)       \\ 
OMP                     & 20\% (64.17$\pm$0.11\%)           & \xmark   & \xmark         \\
\grasp                  & \xmark                        & \xmark                    & \xmark         \\
{\hydra}                & \xmark                       & \xmark                    & \xmark  \\
{\biprune}              & \textbf{36}\% (64.29$\pm$0.13\%)   &\textbf{83}\% (70.95$\pm$0.12\%)  & 74\% (76.09$\pm$0.11\%)\\ \midrule
\bottomrule[1pt]
\end{tabular}%
}
\end{table}

\subsection{Broader Impact}
\label{sec: broader}
We do not recognize any potential negative social impacts of our work. Instead, we believe our work can inspire many techniques for model compression. The finding of structure-aware winning ticket also benefits the design of embedded solutions to deploying large-scale models on resource-limited edge devices (\textit{e.g.}, FPGAs), providing broader impact on both scientific research and practical applications (\textit{e.g.}, autonomous vehicles).

\end{document}